**Multi-Cohort Intelligence Algorithm: An Intra- and Inter-group Learning Behavior based Socio-inspired Optimization Methodology**

Apoorva S Shastri[2], Anand J Kulkarni*[1,2]

[1]Odette School of Business, University of Windsor, 401 Sunset Avenue, Windsor
ON N9B3P4 Canada,
Email: kulk0003@uwindsor.ca, Ph: 1 519 253 3000 x4939

[2]Symbiosis Institute of Technology, Symbiosis International (Deemed University), Pune MH 412 115 India
Email: anand.kulkarni@sitpune.edu.in; apoorva.shastri@sitpune.edu.in
Ph: 91 20 39116468

# Abstract

A Multi-Cohort Intelligence (Multi-CI) metaheuristic algorithm in emerging socio-inspired optimization domain is proposed. The algorithm implements intra-group and inter-group learning mechanisms. It focusses on the interaction amongst different cohorts. The performance of the algorithm is validated by solving 75 unconstrained test problems with dimensions up to 30. The solutions were comparing with several recent algorithms such as Particle Swarm Optimization, Covariance Matrix Adaptation Evolution Strategy, Artificial Bee Colony, Self-adaptive differential evolution algorithm, Comprehensive Learning Particle Swarm Optimization, Backtracking Search Optimization Algorithm and Ideology Algorithm. The Wilcoxon signed rank test was carried out for the statistical analysis and verification of the performance. The proposed Multi-CI outperformed these algorithms in terms of the solution quality including objective function value and computational cost, i.e. computational time and functional evaluations. The prominent feature of the Multi-CI algorithm along with the limitations are discussed as well. In addition, an illustrative example is also solved and every detail is provided.

Keywords: Multi-Cohort Intelligence Algorithm, Socio-inspired optimization, Intra- and Inter-group Learning, Unconstrained Optimization, Metaheuristic

# 1. Introduction

Several nature-inspired optimization algorithms have been developed so far. The notable algorithms are Evolutionary Algorithms (EAs), Genetic Algorithms (GAs), Swarm Optimization (SO) techniques, etc. These methods have proven their superiority in terms of solution quality and computational time over the traditional (exact) methods for solving a wide variety of problem classes. In agreement with the no-free-lunch theorem, certain modifications and supportive techniques are required to be incorporated into these methods when applying for solving a variety of class of problems. This motivated the researchers to resort to development of new optimization methods. An Artificial Intelligence (AI) based socio-inspired optimization methodology referred to as Cohort Intelligence (CI) was proposed by Kulkarni et al. in 2013. It is inspired from the interactive and competitive social behaviour of individual candidates in a cohort. Every candidate exhibits self-interested behaviour and tries to improve it by learning from the other candidates in the cohort. The learning refers to following/adopting the qualities associated with the behaviour of the other candidates. The candidates iteratively follow one another based on certain probability and the cohort is considered saturated/converged when no further improvement in the behaviour of any of the candidates is possible for considerable number of attempts.

The CI methodology was validated by solving several unconstrained test problems (Kulkarni et al. 2013). The algorithm performed better as compared to several versions of the Particle Swarm Optimization (PSO) such as Chaos-PSO (CPSO) and Linearly Decreasing Weight PSO (LDWPSO) (Liu et al. 2010) as well as Robust Hybrid PSO (RHPSO) (Xu et al. 2013). Then was applied for solving a combinatorial problem such as Knapsack problem (Kulkarni and Shabir2016). The algorithm yielded comparable solutions as compared to the Integer programming (IP), Harmony Search (HS) (Zou et al. 2011;Geem et al. 2001), Improved HS (IHS) (Zou et al. 2011;Mahdavi et al. 2007), Novel Global HS (NGHS) (Zou et al. 2011;Layeb 2011, 2013), Quantum Inspired HS Algorithm (QIHSA) (Layeb 2013) and Quantum Inspired Cuckoo Search Algorithm (QICSA) (Layeb 2011).The combinatorial problems from healthcare domain as well as complex large sized Supply Chain problems such as Sea-Cargo problem and Selection of Cross-Border Shippers were also solved (Kulkarni et al. 2016). Furthermore, CI contributed in design of fractional PID controller (Shah and Kulkarni, 2017). CI was applied for solving mechanical engineering problems such as discrete and mixed variable engineering problems (Kale and Kulkarni, 2017) and cup forming design problems (Kulkarni, Kulkarni, Kulkarni, Kakandikar 2016). Recently several variations of CI were proposed by Patankar and Kulkarni (2018). In addition, CI with Cognitive Computing (CICC) was applied for solving steganography problems by (Sarmah and Kulkarni, 2017, 2018). The CI performance was better as compared to the IP solutions as well as specially developed Multi Random Start Local Search (MRSLS) method. In these problems, constraints were handled using a specially developed probability based constraint handling approach. In addition, Traveling Salesman Problem (TSP) was also solved (Kulkarni et al. 2017). The approach was further adopted for solving continuous constrained test problems (Shastri et al. 2016, Kulkarni et al 2016). In addition, complex problem of heat exchanger was also solved using CI method (Dhavale et al. 2016). The solutions were comparable to the techniques such as Differential Evolution (DE) (Price et al. 2005) and GA (Deb et al. 2000).Furthermore, a modified version of CI referred to as MCI as well as its hybridized version with K-means performed better as compared to K-means, K-means++ as well as Genetic Algorithm (GA) (Maulik and Bandyopadhyay 2000), Simulated Annealing (SA) (Niknam and Amiri 2010; Selim and Alsultan 1991), Tabu Search (TS) (Niknam and Amiri 2010), Ant Colony Optimization (ACO) (Shelokar et al. 2004), Honeybee Mating Optimization (HBMO) (Fathian and Amiri 2008) and Particle Swarm Optimization (PSO) (Kao et al. 2008).

It is important to mention here that in the current version of CI (including MCI in which a mutation approach was used for sampling) the candidates learn from the candidates of the same cohort. As the selection is based on roulette wheel approach it is not necessary that the candidate will follow the best candidate in every learning attempt. Even though this helps the candidates jump out of local minima, learning options are limited as only intra-group learning exists. In the society several cohorts exist which interact and compete with one another which could be referred to as inter-group learning. This makes the candidates learn from the candidates within the cohort as well as the candidates from other cohorts. In the proposed Multi-Cohort Intelligence (Multi-CI) approach intra-group learning and inter-group learning mechanisms were implemented. In the intra-group learning mechanism, every candidate based on roulette wheel approach chooses a behaviour from within its own cohort. Then it samples certain behaviours from within the close neighbourhood of the chosen behaviour. In the inter-group learning mechanism, every candidate based on roulette wheel approach chooses a behaviour from within a pool of best behaviours associated with every cohort. Then it chooses the best behaviour by sampling certain number of behaviours from within the close neighbourhood of both behaviours chosen using the intra-group learning and inter-group learning mechanisms.

This manuscript is organized as follows: Section 2 describes the Multi-Cohort Intelligence (Multi-CI) procedure. The performance analysis of the Multi-CI along with the Wilcoxon Signed Rank Test and comparison with the other algorithms is provided in Section 3. The conclusions and future directions are discussed in Section 4. A detailed illustration of the Multi-CI algorithm is provided in an Appendix at the end of the manuscript.

## 2. Multi-Cohort Intelligence (Multi-CI)

Consider a general unconstrained optimization problem (in minimization sense) as follows:

$$\text{Minimize } f(\mathbf{X}) = f(x_1, \dots x_i, \dots x_N) \tag{1}$$

$$\text{Subject to } \psi_i^{lower} \le x_i \le \psi_i^{upper} \ , \quad i = 1, \dots, N$$

In the context of Multi-CI the objective function $f(\mathbf{X})$ is considered as the behavior of an individual candidate in each cohort with associated set of qualities $\mathbf{X} = (x_1, \dots x_i, \dots, x_N)$.

The procedure begins with initialization of learning attempt counter $l = 1$, and $K$ cohorts with number of candidates $C_k$ associated with every cohort $k$, $(k = 1, \dots, K)$. Every candidate $c$ $(c = 1, \dots, C_k)$, $k = 1, \dots, K$ randomly generates qualities $\boldsymbol{X}_k^c = (x_{1,k}^c, \dots x_{i,k}^c, \dots, x_{N,k}^c)$ from within its associated sampling interval $[\psi_i^{lower}, \psi_i^{upper}]$, $i = 1, \dots, N$. The parameters such as convergence parameter ε, sampling interval reduction factor $r$, behavior variations $T$ and $T_Z$ are chosen. The algorithm steps are discussed below and the Multi-CI algorithm flowchart is presented in Figure 1.

***Step 1 (Evaluation of Behaviors)***: The pool of objective functions/behaviors of every candidate $c$ $(c = 1, \dots, C_k)$ associated with every cohort $k (k = 1, \dots, K)$ could be represented as follows:

$$\boldsymbol{F} = \begin{bmatrix} f(\boldsymbol{X}_1^1), \dots, f(\boldsymbol{X}_k^1), \dots, f(\boldsymbol{X}_K^1) \\ \vdots \quad . \quad \vdots \quad . \quad \vdots \\ f(\boldsymbol{X}_1^c), \dots, f(\boldsymbol{X}_k^c), \dots, f(\boldsymbol{X}_K^c) \\ \vdots \quad . \quad \vdots \quad . \quad \vdots \\ f(\boldsymbol{X}_1^{C_1}), \dots, f(\boldsymbol{X}_1^{C_k}), \dots, f(\boldsymbol{X}_K^{C_K}) \end{bmatrix} = [\boldsymbol{f}_1, \dots, \boldsymbol{f}_k, \dots, \boldsymbol{f}_K] \tag{2}$$

***Step 2 (Pool Z Formation)***: The best behavior (objective functions with minimum value) candidate $\hat{c}_k$, $k(k = 1, \dots, K)$ in each cohort are chosen and kept in separated pool $\boldsymbol{Z}$ and the associated set of behaviors $\boldsymbol{F}^{\boldsymbol{Z}}$ is represented as follows:

$$\boldsymbol{F}^{\boldsymbol{Z}} = [\min(\boldsymbol{f}_1), \dots, \min(\boldsymbol{f}_k), \dots, \min(\boldsymbol{f}_K)] = \left[f\left(\boldsymbol{X}_1^{\hat{c}_1}\right), \dots, f\left(\boldsymbol{X}_k^{\hat{c}_k}\right), \dots, f\left(\boldsymbol{X}_K^{\hat{c}_K}\right)\right] \tag{3}$$

***Step 3 (Probability Evaluation 1)***: The probabilities associated with each candidate except pool $\boldsymbol{Z}$ candidates $c$ $(c = 1, \dots, C_k - 1)$, in every cohort $k(k = 1, \dots, K)$ are calculated as follows:

$$p_k^c = \frac{1/f(\boldsymbol{X}_k^c)}{\sum_{c=1}^{C_k-1} 1/f(\boldsymbol{X}_k^c)} \tag{4}$$

***Step 4 (Formation of T behaviors)***: Using roulette wheel approach every candidate selects a behaviour from within its corresponding cohort (except pool $\boldsymbol{Z}$ behaviors) and forms $T$ new behaviours by sampling in close neighbourhood of the qualities associated with the selected behaviour qualities. The neighbourhood of a quality $x_{i,k}^{c}$ associated with the sampling interval $[\psi_i^{c,lower},\ \psi_i^{c,upper}]$, $i = 1, \ldots, N$ of the follower candidate $c\ (c = 1, \ldots, C_k - 1)$, $k(k = 1, \ldots, K)$ is as follows:

$$[\psi_i^{c,lower},\ \psi_i^{c,upper}] = [x_{i,k}^{\widehat{c}} - \left(\left\|\frac{\psi_i^{\widehat{c},upper} - \psi_i^{\widehat{c},lower}}{2}\right\|\right) \times r,\ \ x_{i,k}^{\widehat{c}} + \left(\left\|\frac{\psi_i^{\widehat{c},upper} - \psi_i^{\widehat{c},lower}}{2}\right\|\right) \times r] \tag{5}$$

where $\widehat{c}$ represents the candidate being followed.

The quality matrix $\boldsymbol{Z}^T$ associated with every candidate $c\ (c = 1, \ldots, C_k - 1)$ and corresponding cohort $k(k = 1, \ldots, K)$ is represented as follows:

$$\boldsymbol{Z}^T = \begin{bmatrix} \boldsymbol{Z}_1^{1,T} & \ldots & \boldsymbol{Z}_k^{1,T} & \ldots & \boldsymbol{Z}_K^{1,T} \\ \vdots & \ddots & \vdots & & \vdots \\ \boldsymbol{Z}_1^{c,T} & \ldots & \boldsymbol{Z}_k^{c,T} & \ldots & \boldsymbol{Z}_K^{c,T} \\ \vdots & & \vdots & \ddots & \vdots \\ \boldsymbol{Z}_1^{C_1-1,T} & \ldots & \boldsymbol{Z}_k^{C_k-1,T} & \ldots & \boldsymbol{Z}_K^{C_K-1,T} \end{bmatrix} \tag{6}$$

$$\text{where}\quad \boldsymbol{Z}_k^{c,T} = \begin{bmatrix} \boldsymbol{X}_k^{c,1} \\ \vdots \\ \boldsymbol{X}_k^{c,t} \\ \vdots \\ \boldsymbol{X}_k^{c,T} \end{bmatrix} = \begin{bmatrix} x_{1,k}^{c,1} \ldots x_{i,k}^{c,1} \ldots x_{N,k}^{c,1} \\ \vdots \quad \ddots \quad \vdots \qquad \vdots \\ x_{1,k}^{c,t} \ldots x_{i,k}^{c,t} \ldots x_{N,k}^{c,t} \\ \vdots \qquad \vdots \quad \ddots \quad \vdots \\ x_{1,k}^{c,T} \ldots x_{i,k}^{c,T} \ldots x_{N,k}^{c,T} \end{bmatrix}$$

The behavior matrix $\boldsymbol{F}^T$ associated with $\boldsymbol{Z}^T$ could be represented as follows:

$$\boldsymbol{F}^T = \begin{bmatrix} \boldsymbol{f}(\boldsymbol{Z}_1^{1,T}) & \ldots & \boldsymbol{f}(\boldsymbol{Z}_k^{1,T}) & \ldots & \boldsymbol{f}(\boldsymbol{Z}_K^{1,T}) \\ \vdots & \ddots & \vdots & & \vdots \\ \boldsymbol{f}(\boldsymbol{Z}_1^{c,T}) & \ldots & \boldsymbol{f}(\boldsymbol{Z}_k^{c,T}) & \ldots & \boldsymbol{f}(\boldsymbol{Z}_K^{c,T}) \\ \vdots & & \vdots & \ddots & \vdots \\ \boldsymbol{f}(\boldsymbol{Z}_1^{C_1-1,T}) & \ldots & \boldsymbol{f}(\boldsymbol{Z}_k^{C_k-1,T}) & \ldots & \boldsymbol{f}(\boldsymbol{Z}_K^{C_K-1,T}) \end{bmatrix} \tag{7}$$

$$\text{where}\quad \boldsymbol{f}(\boldsymbol{Z}_k^{c,T}) = \begin{bmatrix} f(\boldsymbol{X}_k^{c,1}) \\ \vdots \\ f(\boldsymbol{X}_k^{c,t}) \\ \vdots \\ f(\boldsymbol{X}_k^{c,T}) \end{bmatrix}$$

***Step 5 (Probability Evaluation 2)***: The probabilities associated with each pool $\boldsymbol{Z}$ candidate $\hat{c}_k$ in every cohort $k(k = 1, \ldots, K)$ are calculated as follows:

$$p^{\hat{c}_k} = \frac{1/f\left(\boldsymbol{X}_k^{\hat{c}_k}\right)}{\sum_{k=1}^{K} 1/f\left(\boldsymbol{X}_k^{\hat{c}_k}\right)} \tag{8}$$

***Step 6 (Formation of $T_Z$ behaviors)***: Also using roulette wheel approach every candidate selects a behavior from within pool $\boldsymbol{Z}$ and forms $T_Z$ new behaviors by sampling in close neighbourhood of the qualities associated with the selected behaviour. The quality matrix $\boldsymbol{Z}^{Tz}$ associated with every candidate $\hat{c}_k$ , $k(k = 1, \ldots, K)$ is represented as follows:

$$\boldsymbol{Z}^{Tz} = [\boldsymbol{Z}^{\hat{c}_1,Tz} \quad \ldots \quad \boldsymbol{Z}^{\hat{c}_k,Tz} \quad \ldots \quad \boldsymbol{Z}^{\hat{c}_K,Tz}] \tag{9}$$

$$\text{where } \boldsymbol{Z}^{\hat{c}_k,Tz} = \begin{bmatrix} \boldsymbol{X}_k^{\hat{c}_k,1} \\ \vdots \\ \boldsymbol{X}_k^{\hat{c}_k,tz} \\ \vdots \\ \boldsymbol{X}_k^{\hat{c}_k,Tz} \end{bmatrix} = \begin{bmatrix} x_{1,k}^{\hat{c}_1,1} & \ldots & x_{i,k}^{\hat{c}_k,1} & \ldots & x_{N,k}^{\hat{c}_K,1} \\ \vdots & \ddots & \vdots & & \vdots \\ x_{1,k}^{\hat{c}_1,tz} & \ldots & x_{i,k}^{\hat{c}_k,tz} & \ldots & x_{N,k}^{\hat{c}_K,tz} \\ \vdots & & \vdots & \ddots & \vdots \\ x_{1,k}^{\hat{c}_1,Tz} & \ldots & x_{i,k}^{\hat{c}_k,Tz} & \ldots & x_{N,k}^{\hat{c}_K,Tz} \end{bmatrix}$$

The behavior matrix $\boldsymbol{F}^{Tz}$ associated with $\boldsymbol{Z}^{Tz}$ could be represented as follows:

$$\boldsymbol{F}^{Tz} = [\boldsymbol{f}(\boldsymbol{Z}^{\hat{c}_1,Tz}) \ldots \boldsymbol{f}(\boldsymbol{Z}^{\hat{c}_k,Tz}) \ldots \boldsymbol{f}(\boldsymbol{Z}^{\hat{c}_K,Tz})] \tag{10}$$

$$\text{where } \boldsymbol{f}(\boldsymbol{Z}^{\hat{c}_k,Tz}) = \begin{bmatrix} f(\boldsymbol{X}_k^{\hat{c}_k,1}) \\ \vdots \\ f(\boldsymbol{X}_k^{\hat{c}_k,tz}) \\ \vdots \\ f(\boldsymbol{X}_k^{\hat{c}_k,Tz}) \end{bmatrix}$$

***Step 7 (Selection)***: Every candidate $c$ $(c = 1, \ldots, C_k - 1)$ associated with every cohort $k(k = 1, \ldots, K)$ selects the best behavior, i.e. minimum objective function value from within its behavior choices in $\boldsymbol{F}^{T}$ and $\boldsymbol{F}^{Tz}$ as follows:

$$f_{k,min}^{c} = min\left(\boldsymbol{f}(\boldsymbol{Z}_k^{c,t}), \boldsymbol{f}(\boldsymbol{Z}^{\hat{c}_k,tz})\right) \ , \ (c = 1, \ldots, C_k - 1) \ , \ k(k = 1, \ldots, K) \tag{11}$$

***Step 8 (Concatenation)*** The pool $\boldsymbol{Z}$ behaviors $\boldsymbol{F}^{\boldsymbol{Z}}$ (Equation (3)) are carry forwarded to the subsequent learning attempt. The modified pool of behaviors $\boldsymbol{F}$ is given below.

$$\boldsymbol{F} = \begin{bmatrix} f_{1,min}^{1}, & \cdots, f_{k,min}^{1}, & \cdots, f_{K,min}^{1} \\ \vdots & \ddots \quad \vdots \quad . & \vdots \\ f_{1,min}^{c}, & \cdots, f_{k,min}^{c}, & \cdots, f_{K,min}^{c} \\ \vdots & . \quad \vdots \quad \ddots & \vdots \\ f_{1,min}^{C_1}, & \cdots, f_{k,min}^{C_k}, & \cdots, f_{K,min}^{C_K} \end{bmatrix} = [\boldsymbol{f}_1, \ldots, \boldsymbol{f}_k, \ldots, \boldsymbol{f}_K] \tag{12}$$

***Step 8 (Convergence)***: The algorithm is assumed to have converged if all of the conditions listed in Equation (13) are satisfied for successive considerable number of learning attempts and accept any of the current behaviors as final solution $f^*$ from within $K$ cohorts.

$$\left.\begin{array}{c} \|max(\boldsymbol{F}^{l}) - max(\boldsymbol{F}^{l-1})\| \le \varepsilon \\ \|min(\boldsymbol{F}^{l}) - min(\boldsymbol{F}^{l-1})\| \le \varepsilon \\ \|max(\boldsymbol{F}^{l}) - min(\boldsymbol{F}^{l})\| \le \varepsilon \end{array}\right\} \tag{13}$$

where $l$ is learning attempt counter.

An illustrative example (Sphere function with 2 variables) of the above discussed Multi-CI procedure is provided in Appendix A of the manuscript. It includes every details of first learning attempt followed by evaluation of every step (1 to 8) is listed in Table A.1 till convergence.

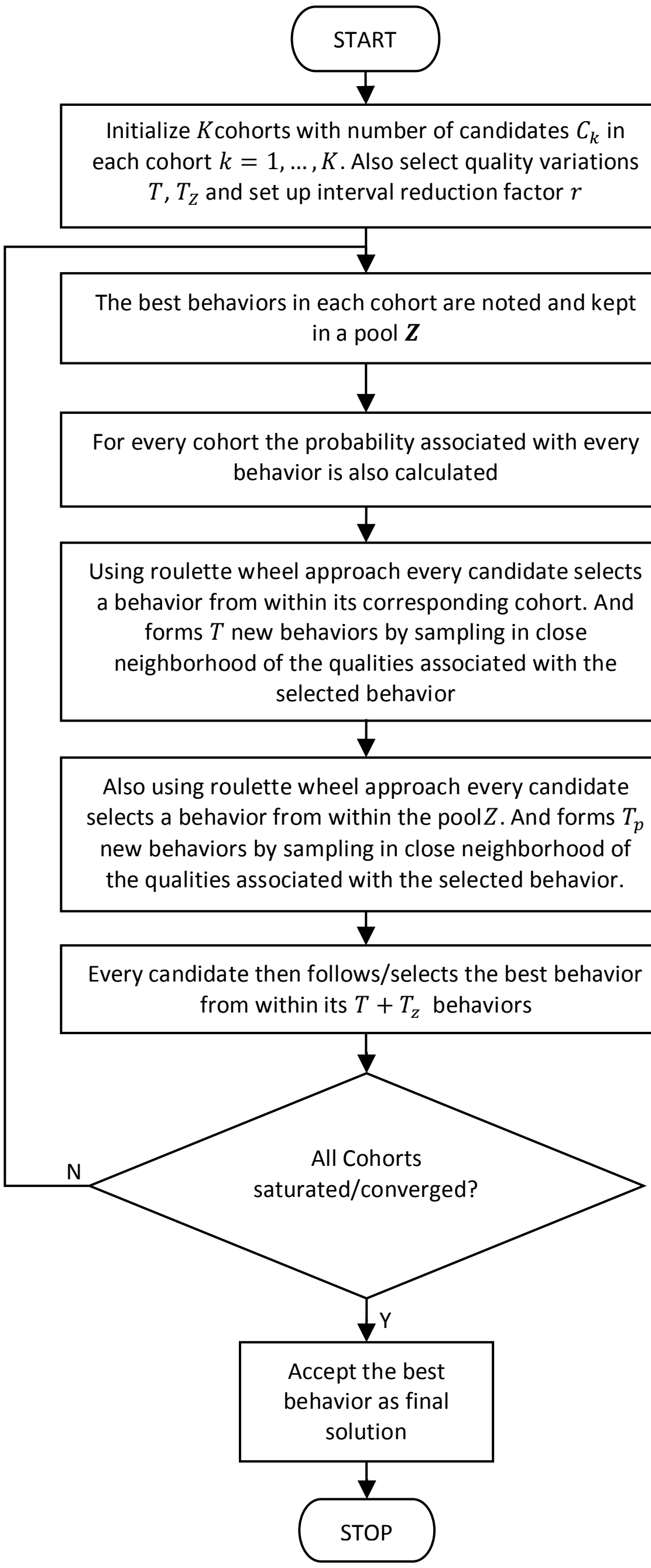

Figure 1 Multi-CI Algorithm Flowchart

# 3. Results and Discussions

The Multi-CI algorithm was coded in in MATLAB R2013a onWindows Platform with a T6400@4 GHz Intel Core 2 Duoprocessor with 4 GB RAM.The algorithm was validated by solving two well studied sets of test problems. Set 1 included 50 well studied benchmark problems (Karaboga and Akay, 2009; Karaboga and Basturk, 2007). Set 2 included 25 test problems from CEC 2005 (Suganthan et al. 2005). Set 1 test problems are listed in Table 1 and Set 2 test problems are listed in Table 2. Every problem in these test cases was solved 30 times using Multi-CI. In every run, initial behaviour of every candidate was randomly initialized. The Multi-CI parameters chosen were as follows: Number of cohorts $K = 3$, Number of candidates $C^k = 5$, Reduction factor value $r = 0.98$, Reduction factor value $r = 0.98$, Quality variation parameters $T = 5$ and $T_p = 10$.

### 3.1 Statistical Analysis

The Multi-CI algorithm presented here and the other algorithms with which the results are being compared are stochastic in nature due to which in every independent run of the algorithm the converged solution may be different than one another. A pairwise comparison of Multi-CI and every other algorithm was carried out, i.e. the converged (global minimum) values of 30 independent runs solving every problem using Multi-CI are compared with every other algorithm solving 30 independent runs of these problems. The Wilcoxon Signed-Rank test was used for such pairwise comparison. Similar to (Civicioglu, 2013), the significance value α was chosen to be 0.05 with null hypothesis H0 is: There is no difference between the median of the solutions obtained by algorithm A and the median of the solutions obtained by algorithm B for the same set of test problems, i.e. median (A) = median(B). Also, to determine whether algorithm A yielded statistically better solution than algorithm B or whether alternative hypothesis was valid, the sizes of the ranks provided by the Wilcoxon Signed-Rank test (T+ and T-) were thoroughly examined.

The mean solution, best solution and standard deviation (Std. Dev.), mean run time (in seconds) over the 30 runs of the algorithms solving Test 1 and Test 2 problems are represented in Table 3 and Table 4, respectively. The algorithms with statistically better solutions for Test 1 and Test 2 problems found using Wilcoxon Signed-Rank test are presented in Table 5 and Table 6, respectively. In these tables, '+' indicated that the null hypothesis H0 was rejected and Multi-CI performed better and '-' indicated that the null hypothesis H0 was rejected; however, Multi-CI performed worse. The '=' indicated that there is no statistical difference between the two algorithms solving the problems and none of the two algorithms being compared could be considered more successful (winner) solving that problem. The counts of statistical significant cases (+/-/=) are presented in the last row of Table 5 and 6.The multi problem based pairwise using the averages of the global solutions obtained over the 30 runs of the algorithms solving the Test 1 and Test 2 problems are presented in Table 7. The results highlighted that the Multi-CI algorithm performed significantly better than every other algorithm.

The convergence plots of few unimodal and multimodal representative functions such as Ackley function, Beale function, Fletcher function, Foxhole functions, Michalewics function, Six-hump camelback function are presented in Figure 2-7. The best solutions in every learning attempt are also plotted in Figure 2(b), 3(b), 4(b), 5(b), 6(b) and 7(b). These plots exhibited the candidates' self-supervised intra as well as inter cohort learning behaviour. The convergence also highlighted the significance of Multi-CI approach quickly reaching the optimum solution.

Table 1: The benchmark problems used in Test 1 (Dim = Dimension; Low and Up = Limitations of search space; U = Unimodal; M = Multimodal; S = Separable; N = Non-separable)

| Problem | Name | Type | Low | Up | Dimension |
|---|---|---|---|---|---|
| F1 | Foxholes | MS | -65.536 | 65.536 | 2 |
| F2 | Goldstein-Price | MN | -2 | 2 | 2 |
| F3 | Penalized | MN | -50 | 50 | 30 |
| F4 | Penalized2 | MN | -50 | 50 | 30 |
| F5 | Ackley | MN | -32 | 32 | 30 |
| F6 | Beale | UN | -4.5 | 4.5 | 5 |
| F7 | Bohachecsky1 | MS | -100 | 100 | 2 |
| F8 | Bohachecsky2 | MN | -100 | 100 | 2 |
| F9 | Bohachecsky3 | MN | -100 | 100 | 2 |
| F10 | Booth | MS | -10 | 10 | 2 |
| F11 | Branin | MS | -5 | 10 | 2 |
| F12 | Colville | UN | -10 | 10 | 4 |
| F13 | Dixon-Price | UN | -10 | 10 | 30 |
| F14 | Easom | UN | -100 | 100 | 2 |
| F15 | Fletcher | MN | -3.1416 | 3.1416 | 2 |
| F16 | Fletcher | MN | -3.1416 | 3.1416 | 5 |
| F17 | Fletcher | MN | -3.1416 | 3.1416 | 10 |
| F18 | Griewank | MN | -600 | 600 | 30 |
| F19 | Hartman3 | MN | 0 | 1 | 3 |
| F20 | Hartman6 | MN | 0 | 1 | 6 |
| F21 | Kowalik | MN | -5 | 5 | 4 |
| F22 | Langermann2 | MN | 0 | 10 | 2 |
| F23 | Langermann5 | MN | 0 | 10 | 5 |
| F24 | Langermann10 | MN | 0 | 10 | 10 |
| F25 | Matyas | UN | -10 | 10 | 2 |
| F26 | Michalewics2 | MS | 0 | 3.1416 | 2 |
| F27 | Michalewics5 | MS | 0 | 3.1416 | 5 |
| F28 | Michalewics10 | MS | 0 | 3.1416 | 10 |
| F29 | Perm | MN | -4 | 4 | 4 |
| F30 | Powell | UN | -4 | 5 | 24 |
| F31 | Powersum | MN | 0 | 4 | 4 |
| F32 | Quartic | US | -1.28 | 1.28 | 30 |
| F33 | Rastrigin | MS | -5.12 | 5.12 | 30 |
| F34 | Rosenbrock | UN | -30 | 30 | 30 |
| F35 | Schaffer | MN | -100 | 100 | 2 |
| F36 | Schwefel | MS | -500 | 500 | 30 |
| F37 | Schwefel_1_2 | UN | -100 | 100 | 30 |
| F38 | Schwefel_2_22 | UN | -10 | 10 | 30 |
| F39 | Shekel10 | MN | 0 | 10 | 4 |
| F40 | Shekel5 | MN | 0 | 10 | 4 |
| F41 | Shekel7 | MN | 0 | 10 | 4 |
| F42 | Shubert | MN | -10 | 10 | 2 |
| F43 | Six-hump camelback | MN | -5 | 5 | 2 |
| F44 | Sphere2 | US | -100 | 100 | 30 |
| F45 | Step2 | US | -100 | 100 | 30 |
| F46 | Stepint | US | -5.12 | 5.12 | 5 |
| F47 | Sumsquares | US | -10 | 10 | 30 |
| F48 | Trid6 | UN | -36 | 36 | 6 |
| F49 | Trid10 | UN | -100 | 100 | 10 |
| F50 | Zakharov | UN | -5 | 10 | 10 |

Table 2: The benchmark problems used in Test 2 (Dim = Dimension; Low and Up = Limitations of search space; U = Unimodal; M = Multimodal; E = Expanded; H = Hybrid)

| **Problem** | **Name** | **Type** | **Low** | **Up** | **Dimension** |
|---|---|---|---|---|---|
| F51 | Shifted sphere | U | -100 | 100 | 10 |
| F52 | Shifted Schwefel | U | -100 | 100 | 10 |
| F53 | Shifted rotated high conditioned elliptic function | U | -100 | 100 | 10 |
| F54 | Shifted Schwefels problem 1.2 with noise | U | -100 | 100 | 10 |
| F55 | Schwefels problem 2.6 | U | -100 | 100 | 10 |
| F56 | Shifted Rosenbrock’s | M | -100 | 100 | 10 |
| F57 | Shifted rotated Griewank’s | M | 0 | 600 | 10 |
| F58 | Shifted rotated Ackley’s | M | -32 | 32 | 10 |
| F59 | Shifted Rastrigin’s | M | -5 | 5 | 10 |
| F60 | Shifted rotated Rastrigin’s | M | -5 | 5 | 10 |
| F61 | Shifted rotated Weierstrass | M | -0.5 | 0.5 | 10 |
| F62 | Schwefels problem 2.13 | M | -100 | 100 | 10 |
| F63 | Expanded extended Griewank’s + Rosenbrock’s | E | -3 | 1 | 10 |
| F64 | Expanded rotated extended Scaffes | E | -100 | 100 | 10 |
| F65 | Hybrid composition function | HC | -5 | 5 | 10 |
| F66 | Rotated hybrid comp. Fn 1 | HC | -5 | 5 | 10 |
| F67 | Rotated hybrid comp. Fn 1 with noise | HC | -5 | 5 | 10 |
| F68 | Rotated hybrid comp. Fn 2 | HC | -5 | 5 | 10 |
| F69 | Rotated hybrid comp. Fn 2 with narrow global optimal | HC | -5 | 5 | 10 |
| F70 | Rotated hybrid comp. Fn 2 with the global optimum | HC | -5 | 5 | 10 |
| F71 | Rotated hybrid comp. Fn 3 | HC | -5 | 5 | 10 |
| F72 | Rotated hybrid comp. Fn 3 with high condition number matrix | HC | -5 | 5 | 10 |
| F73 | Non-continuous rotated hybrid comp. Fn 3 | HC | -5 | 5 | 10 |
| F74 | Rotated hybrid comp. Fn 4 | HC | -5 | 5 | 10 |
| F75 | Rotated hybrid comp. Fn 4 | HC | -2 | 5 | 10 |

Table 3: Statistical solutions to Test 1 Problems using PSO, CMAES, ABC, CLPSO, SADE, BSA, IA and Multi-CI
(Mean = Mean solution; Std. Dev. = Standard-deviation of mean solution; Best = Best solution; Runtime = Mean runtime in seconds)

| Problem | Statistics | PSO2011 | CMAES | ABC | JDE | CLPSO | SADE | BSA | IA | Multi CI |
|---|---|---|---|---|---|---|---|---|---|---|
| F1 | Mean | 1.3316029264876300 | 10.0748846367972000 | 0.9980038377944500 | 1.0641405484285200 | 1.8209961275956800 | 0.9980038377944500 | 0.9980038377944500 | 0.9980038690000000 | 0.9980038377944500 |
| | Std. Dev. | 0.9455237994690700 | 8.0277365400340800 | 0.0000000000000001 | 0.3622456829347420 | 1.6979175079427900 | 0.0000000000000000 | 0.0000000000000000 | 0.0000000000000035 | 0.0000000000000003 |
| | Best | 0.9980038377944500 | 0.9980038377944500 | 0.9980038377944500 | 0.9980038377944500 | 0.9980038377944500 | 0.9980038377944500 | 0.9980038377944500 | 0.9980038685998520 | 0.9980038377944500 |
| | Runtime | 72.527 | 44.788 | 64.976 | 51.101 | 61.650 | 66.633 | 38.125 | 43.535 | 1.092 |
| | | | | | | | | | | |
| F2 | Mean | 2.9999999999999200 | 21.8999999999995000 | 3.0000000465423000 | 2.9999999999999200 | 3.0000000000000700 | 2.9999999999999200 | 2.9999999999999200 | 3.0240147900000000 | 2.9999999999999200 |
| | Std. Dev. | 0.0000000000000013 | 32.6088098948516000 | 0.0000002350442161 | 0.0000000000000013 | 0.0000000000007941 | 0.0000000000000020 | 0.0000000000000011 | 0.0787814840000000 | 0.0000000000000005 |
| | Best | 2.9999999999999200 | 2.9999999999999200 | 2.9999999999999200 | 2.9999999999999200 | 2.9999999999999200 | 2.9999999999999200 | 2.9999999999999200 | 3.0029461118668700 | 2.9999999999999200 |
| | Runtime | 17.892 | 24.361 | 16.624 | 7.224 | 24.784 | 28.699 | 7.692 | 41.343 | 0.763 |
| | | | | | | | | | | |
| F3 | Mean | 0.1278728062391630 | 0.0241892995662904 | 0.0000000000000004 | 0.0034556340083499 | 0.0000000000000000 | 0.0034556340083499 | 0.0000000000000000 | 0.3536752140000000 | 0.0000000000000000 |
| | Std. Dev. | 0.2772792346028400 | 0.0802240262581864 | 0.0000000000000001 | 0.0189272869685522 | 0.0000000000000000 | 0.0189272869685522 | 0.0000000000000000 | 1.4205454130000000 | 0.0000000000000000 |
| | Best | 0.0000000000000000 | 0.0000000000000000 | 0.0000000000000003 | 0.0000000000000000 | 0.0000000000000000 | 0.0000000000000000 | 0.0000000000000000 | 0.0014898619035614 | 0.0000000000000000 |
| | Runtime | 139.555 | 5.851 | 84.416 | 9.492 | 38.484 | 15.992 | 18.922 | 34.494 | 90.997 |
| | | | | | | | | | | |
| F4 | Mean | 0.0043949463343535 | 0.0003662455278628 | 0.0000000000000004 | 0.0007324910557256 | 0.0000000000000000 | 0.0440448539086004 | 0.0000000000000000 | 0.0179485820000000 | 0.0000000000000000 |
| | Std. Dev. | 0.0054747064090174 | 0.0020060093719584 | 0.0000000000000001 | 0.0027875840585535 | 0.0000000000000000 | 0.2227372747439610 | 0.0000000000000000 | 0.0526650620000000 | 0.0000000000000000 |
| | Best | 0.0000000000000000 | 0.0000000000000000 | 0.0000000000000003 | 0.0000000000000000 | 0.0000000000000000 | 0.0000000000000000 | 0.0000000000000000 | 0.0000000000165491 | 0.0000000000000000 |
| | Runtime | 126.507 | 6.158 | 113.937 | 14.367 | 48.667 | 33.019 | 24.309 | 322.808 | 27.589 |
| | | | | | | | | | | |
| F5 | Mean | 1.5214322973725000 | 11.7040011684582000 | 0.0000000000000340 | 0.0811017056422860 | 0.1863456353861950 | 0.7915368220335460 | 0.0000000000000105 | 0.0000000000000009 | 0.0000000000000000 |
| | Std. Dev. | 0.6617570384662600 | 9.7201961540865200 | 0.0000000000000035 | 0.3176012689149320 | 0.4389839299322230 | 0.7561593402959740 | 0.0000000000000034 | 0.0000000000000000 | 0.0000000000000000 |
| | Best | 0.0000000000000080 | 0.0000000000000080 | 0.0000000000000293 | 0.0000000000000044 | 0.0000000000000080 | 0.0000000000000044 | 0.0000000000000080 | 0.0000000000000009 | 0.0000000000000000 |
| | Runtime | 63.039 | 3.144 | 23.293 | 11.016 | 45.734 | 40.914 | 14.396 | 49.458 | 5.243 |
| | | | | | | | | | | |
| F6 | Mean | 0.0000000041922968 | 0.2540232169641050 | 0.0000000000000028 | 0.0000000000000000 | 0.0000444354499943 | 0.0000000000000000 | 0.0000000000000000 | 0.0082236060000000 | 0.0000000000000000 |
| | Std. Dev. | 0.0000000139615552 | 0.3653844307786430 | 0.0000000000000030 | 0.0000000000000000 | 0.0001015919507724 | 0.0000000000000000 | 0.0000000000000000 | 0.0000000000000000 | 0.0000000000000000 |
| | Best | 0.0000000000000000 | 0.0000000000000000 | 0.0000000000000005 | 0.0000000000000000 | 0.0000000000000000 | 0.0000000000000000 | 0.0000000000000000 | 0.0082236059357692 | 0.0000000000000000 |
| | Runtime | 32.409 | 4.455 | 22.367 | 1.279 | 125.839 | 4.544 | 0.962 | 50.246 | 1.356 |
| | | | | | | | | | | |
| F7 | Mean | 0.0000000000000000 | 0.0622354533647150 | 0.0000000000000000 | 0.0000000000000000 | 0.0000000000000000 | 0.0000000000000000 | 0.0000000000000000 | 0.0000000000000000 | 0.0000000000000000 |
| | Std. Dev. | 0.0000000000000000 | 0.1345061339146580 | 0.0000000000000000 | 0.0000000000000000 | 0.0000000000000000 | 0.0000000000000000 | 0.0000000000000000 | 0.0000000000000000 | 0.0000000000000000 |
| | Best | 0.0000000000000000 | 0.0000000000000000 | 0.0000000000000000 | 0.0000000000000000 | 0.0000000000000000 | 0.0000000000000000 | 0.0000000000000000 | 0.0000000000000000 | 0.0000000000000000 |
| | Runtime | 16.956 | 6.845 | 1.832 | 1.141 | 2.926 | 4.409 | 0.825 | 38.506 | 1.434 |
| | | | | | | | | | | |
| F8 | Mean | 0.0000000000000000 | 0.0072771062590204 | 0.0000000000000000 | 0.0000000000000000 | 0.0000000000000000 | 0.0000000000000000 | 0.0000000000000000 | 0.0000000000000000 | 0.0000000000000000 |
| | Std. Dev. | 0.0000000000000000 | 0.0398583525142753 | 0.0000000000000000 | 0.0000000000000000 | 0.0000000000000000 | 0.0000000000000000 | 0.0000000000000000 | 0.0000000000000000 | 0.0000000000000000 |
| | Best | 0.0000000000000000 | 0.0000000000000000 | 0.0000000000000000 | 0.0000000000000000 | 0.0000000000000000 | 0.0000000000000000 | 0.0000000000000000 | 0.0000000000000000 | 0.0000000000000000 |
| | Runtime | 17.039 | 2.174 | 1.804 | 1.139 | 2.891 | 4.417 | 0.824 | 39.023 | 1.542 |
| | | | | | | | | | | |
| F9 | Mean | 0.0000000000000000 | 0.0001048363065820 | 0.0000000000000006 | 0.0000000000000000 | 0.0000193464326398 | 0.0000000000000000 | 0.0000000000000000 | 0.0000000000000000 | 0.0000000000000000 |
| | Std. Dev. | 0.0000000000000000 | 0.0005742120996051 | 0.0000000000000003 | 0.0000000000000000 | 0.0000846531630676 | 0.0000000000000000 | 0.0000000000000000 | 0.0000000000000000 | 0.0000000000000000 |
| | Best | 0.0000000000000000 | 0.0000000000000000 | 0.0000000000000001 | 0.0000000000000000 | 0.0000000000000000 | 0.0000000000000000 | 0.0000000000000000 | 0.0000000000000000 | 0.0000000000000000 |
| | Runtime | 17.136 | 2.127 | 21.713 | 1.129 | 33.307 | 4.303 | 0.829 | 40.896 | 1.433 |
| | | | | | | | | | | |
| F10 | Mean | 0.0000000000000000 | 0.0000000000000000 | 0.0000000000000000 | 0.0000000000000000 | 0.0006005122443674 | 0.0000000000000000 | 0.0000000000000000 | 0.8346587090000000 | 0.0000000000000000 |
| | Std. Dev. | 0.0000000000000000 | 0.0000000000000000 | 0.0000000000000000 | 0.0000000000000000 | 0.0029861918862801 | 0.0000000000000000 | 0.0000000000000000 | 0.0000000000000005 | 0.0000000000000000 |
| | Best | 0.0000000000000000 | 0.0000000000000000 | 0.0000000000000000 | 0.0000000000000000 | 0.0000000000000000 | 0.0000000000000000 | 0.0000000000000000 | 0.8346587086917530 | 0.0000000000000000 |
| | Runtime | 17.072 | 1.375 | 22.395 | 1.099 | 28.508 | 4.371 | 0.790 | 39.978 | 1.260 |
| | | | | | | | | | | |
| F11 | Mean | 0.3978873577297380 | 0.6372170283279430 | 0.3978873577297380 | 0.3978873577297380 | 0.3978873577297390 | 0.3978873577297380 | 0.3978873577297380 | 0.4156431270000000 | 0.3978873577297380 |
| | Std. Dev. | 0.0000000000000000 | 0.7302632173480510 | 0.0000000000000000 | 0.0000000000000000 | 0.0000000000000049 | 0.0000000000000000 | 0.0000000000000000 | 0.0406451050000000 | 0.0000000000000001 |
| | Best | 0.3978873577297380 | 0.3978873577297380 | 0.3978873577297380 | 0.3978873577297380 | 0.3978873577297380 | 0.3978873577297380 | 0.3978873577297380 | 0.4012748152492080 | 0.3978873577297380 |
| | Runtime | 17.049 | 24.643 | 10.941 | 6.814 | 17.283 | 27.981 | 5.450 | 40.099 | 0.603 |
| | | | | | | | | | | |
| F12 | Mean | 0.0000000000000000 | 0.0000000000000000 | 0.0715675060725970 | 0.0000000000000000 | 0.1593872502094070 | 0.0000000000000000 | 0.0000000000000000 | 0.0014898620000000 | 0.0000000000000000 |
| | Std. Dev. | 0.0000000000000000 | 0.0000000000000000 | 0.0579425013417103 | 0.0000000000000000 | 0.6678482786713720 | 0.0000000000000000 | 0.0000000000000000 | 0.0000000000000000 | 0.0000000000000000 |
| | Best | 0.0000000000000000 | 0.0000000000000000 | 0.0013425253994745 | 0.0000000000000000 | 0.0000094069599934 | 0.0000000000000000 | 0.0000000000000000 | 0.0082029783984983 | 0.0000000000000000 |
| | Runtime | 44.065 | 1.548 | 21.487 | 1.251 | 166.965 | 4.405 | 2.460 | 48.067 | 41.69 |
| | | | | | | | | | | |
| F13 | Mean | 0.6666666666666750 | 0.6666666666666670 | 0.0000000000000038 | 0.6666666666666670 | 0.0023282133668190 | 0.6666666666666670 | 0.6444444444444440 | 0.2528116640000000 | 0.6728903646849310 |
| | Std. Dev. | 0.0000000000000022 | 0.0000000000000000 | 0.0000000000000012 | 0.0000000000000002 | 0.0051792840882291 | 0.0000000000000000 | 0.1217161238900370 | 0.0000000006509080 | 0.2130263402454600 |
| | Best | 0.6666666666666720 | 0.6666666666666670 | 0.0000000000000021 | 0.6666666666666670 | 0.0000120708732167 | 0.6666666666666670 | 0.0000000000000000 | 0.2528116633611470 | 0.0000000000000020 |
| | Runtime | 167.094 | 3.719 | 37.604 | 18.689 | 216.261 | 47.833 | 21.192 | 67.463 | 11.104 |
| | | | | | | | | | | |
| F14 | Mean | -1.0000000000000000 | -0.1000000000000000 | -1.0000000000000000 | -1.0000000000000000 | -1.0000000000000000 | -1.0000000000000000 | -1.0000000000000000 | -0.9997989620000000 | 0.0000000000000000 |
| | Std. Dev. | 0.0000000000000000 | 0.3051285766293650 | 0.0000000000000000 | 0.0000000000000000 | 0.0000000000000000 | 0.0000000000000000 | 0.0000000000000000 | 0.0000000000167151 | 0.0000000000000000 |
| | Best | -1.0000000000000000 | -1.0000000000000000 | -1.0000000000000000 | -1.0000000000000000 | -1.0000000000000000 | -1.0000000000000000 | -1.0000000000000000 | -0.9997989624626810 | 0.0000000000000000 |
| | Runtime | 16.633 | 3.606 | 13.629 | 6.918 | 16.910 | 28.739 | 5.451 | 39.685 | 0.10 |
| | | | | | | | | | | |
| F15 | Mean | 0.0000000000000000 | 1028.3930784026900000 | 0.0000000000000000 | 0.0000000000000000 | 0.0000000000000000 | 0.0000000000000000 | 0.0000000000000000 | 0.0000000000000000 | 0.0000000000000000 |
| | Std. Dev. | 0.0000000000000000 | 1298.1521820113500000 | 0.0000000000000000 | 0.0000000000000000 | 0.0000000000000000 | 0.0000000000000000 | 0.0000000000000000 | 0.0000000000000000 | 0.0000000000000000 |
| | Best | 0.0000000000000000 | 0.0000000000000000 | 0.0000000000000000 | 0.0000000000000000 | 0.0000000000000000 | 0.0000000000000000 | 0.0000000000000000 | 0.0000000000000000 | 0.0000000000000000 |
| | Runtime | 27.859 | 15.541 | 40.030 | 2.852 | 4.030 | 6.020 | 2.067 | 38.867 | 1.860 |
| | | | | | | | | | | |
| F16 | Mean | 48.7465164446927000 | 1680.3460230073400000 | 0.0218688498331872 | 0.9443728655432830 | 81.7751618148164000 | 0.0000000000000000 | 0.0000000000000000 | 0.0000000000000000 | 0.0000000000000000 |
| | Std. Dev. | 88.8658510972991000 | 2447.7484859066000000 | 0.0418409568792831 | 2.8815514827061600 | 379.9241117377270000 | 0.0000000000000000 | 0.0000000000000000 | 0.0000000000000000 | 0.0000000000000000 |
| | Best | 0.0000000000000000 | 0.0000000000000000 | 0.0000000000000016 | 0.0000000000000000 | 0.0000000000000000 | 0.0000000000000000 | 0.0000000000000000 | 0.0000000000000000 | 0.0000000000000000 |
| | Runtime | 95.352 | 11.947 | 44.572 | 4.719 | 162.941 | 5.763 | 7.781 | 48.262 | 0.459 |
| | | | | | | | | | | |
| F17 | Mean | 918.9518492782850000 | 12340.2283326398000000 | 11.0681496253548000 | 713.7226974626920000 | 0.8530843976878610 | 0.0000000000000000 | 0.0000000000000000 | 0.0000000000000000 | 0.0000000000000000 |
| | Std. Dev. | 1652.4810858411400000 | 22367.1698875802000000 | 9.8810950146557100 | 1710.0713074301200000 | 2.9208253191698800 | 0.0000000000000000 | 0.0000000000000000 | 0.0000000000000000 | 0.0000000000000000 |
| | Best | 0.0000000000000000 | 0.0000000000000000 | 0.3274654777056860 | 0.0000000000000000 | 0.0016957837829822 | 0.0000000000000000 | 0.0000000000000000 | 0.0000000000000000 | 0.0000000000000000 |
| | Runtime | 271.222 | 7.631 | 43.329 | 16.105 | 268.894 | 168.310 | 33.044 | 69.060 | 1.860 |
| | | | | | | | | | | |
| F18 | Mean | 0.0068943694819713 | 0.0011498935321349 | 0.0000000000000000 | 0.0048193578543185 | 0.0000000000000000 | 0.0226359326967139 | 0.0004930693556077 | 0.0000000000000000 | 0.0000000000000000 |
| | Std. Dev. | 0.0080565201649587 | 0.0036449413521107 | 0.0000000000000001 | 0.0133238235582874 | 0.0000000000000000 | 0.0283874287215679 | 0.0018764355751644 | 0.0000000000000000 | 0.0000000000000000 |
| | Best | 0.0000000000000000 | 0.0000000000000000 | 0.0000000000000000 | 0.0000000000000000 | 0.0000000000000000 | 0.0000000000000000 | 0.0000000000000000 | 0.0000000000000000 | 0.0000000000000000 |
| | Runtime | 73.895 | 2.647 | 19.073 | 6.914 | 14.864 | 25.858 | 5.753 | 2.717 | 4.261 |
| | | | | | | | | | | |
| F19 | Mean | -3.8627821478207500 | -3.7243887744664700 | -3.8627821478207500 | -3.8627821478207500 | -3.8627821478207500 | -3.8627821478207500 | -3.8627821478207500 | -3.8596352620000000 | -3.8627819786235600 |
| | Std. Dev. | 0.0000000000000027 | 0.5407823545193820 | 0.0000000000000024 | 0.0000000000000027 | 0.0000000000000027 | 0.0000000000000027 | 0.0000000000000027 | 0.0033967610000000 | 0.0000001322237558 |
| | Best | -3.8627821478207600 | -3.8627821478207600 | -3.8627821478207600 | -3.8627821478207600 | -3.8627821478207600 | -3.8627821478207600 | -3.8627821478207600 | -3.8613076574052300 | -3.8627821093820800 |
| | Runtime | 19.280 | 21.881 | 12.613 | 7.509 | 17.504 | 24.804 | 6.009 | 46.167 | 1.285 |
| | | | | | | | | | | |
| F20 | Mean | -3.3180320675402500 | -3.2942534432762600 | -3.3219951715842400 | -3.2982165473202600 | -3.3219951715842400 | -3.3140689634962500 | -3.3219951715842400 | -2.5710247593206100 | -3.3223582775589500 |
| | Std. Dev. | 0.0217068148263721 | 0.0511458075926848 | 0.0000000000000014 | 0.0483702518391572 | 0.0000000000000013 | 0.0301641516823498 | 0.0000000000000013 | 0.0000000000000009 | 0.0099173853696568 |
| | Best | -3.3219951715842400 | -3.3219951715842400 | -3.3219951715842400 | -3.3219951715842400 | -3.3219951715842400 | -3.3219951715842400 | -3.3219951715842400 | -2.5710247593206100 | -3.3223651489966400 |
| | Runtime | 26.209 | 7.333 | 13.562 | 8.008 | 20.099 | 33.719 | 6.822 | 59.083 | 2.021 |
| | | | | | | | | | | |
| F21 | Mean | 0.0003074859878056 | 0.0064830287538208 | 0.0004414866359626 | 0.0003685318137604 | 0.0003100479704151 | 0.0003074859878056 | 0.0003074859878056 | 0.0016993410000000 | 0.0003516458357319 |
| | Std. Dev. | 0.0000000000000000 | 0.0148565973286009 | 0.0000568392289725 | 0.0002323173367683 | 0.0000059843325073 | 0.0000000000000000 | 0.0000000000000000 | 0.0000013058400000 | 0.0000539770693216 |
| | Best | 0.0003074859878056 | 0.0003074859878056 | 0.0003230956007045 | 0.0003074859878056 | 0.0003074859941292 | 0.0003074859878056 | 0.0003074859878056 | 0.0016989914552560 | 0.0003243793470953 |
| | Runtime | 84.471 | 13.864 | 20.255 | 7.806 | 156.095 | 45.443 | 11.722 | 48.920 | 1.800 |
| | | | | | | | | | | |
| F22 | Mean | -1.0809384421344400 | -0.7323679641701760 | -1.0809384421344400 | -1.0764280762657400 | -1.0202940450426400 | -1.0809384421344400 | -1.0809384421344400 | -1.4315374190000000 | -1.0820489785202800 |
| | Std. Dev. | 0.0000000000000006 | 0.4136688304155380 | 0.0000000000000008 | 0.0247042912888477 | 0.1190811583120530 | 0.0000000000000005 | 0.0000000000000005 | 0.0000000000000009 | 0.0000000000000000 |
| | Best | -1.0809384421344400 | -1.0809384421344400 | -1.0809384421344400 | -1.0809384421344400 | -1.0809384421344400 | -1.0809384421344400 | -1.0809384421344400 | -1.4315374193830000 | -1.0820489785202800 |
| | Runtime | 27.372 | 32.311 | 27.546 | 19.673 | 52.853 | 36.659 | 21.421 | 34.714 | 1.299 |
| | | | | | | | | | | |
| F23 | Mean | -1.3891992200744600 | -0.5235864386288060 | -1.4999990070800800 | -1.3431399432579700 | -1.4765972735526500 | -1.4999992233525000 | -1.4821658762555300 | -1.5000000000000000 | -1.4999999979385900 |
| | Std. Dev. | 0.2257194403158630 | 0.2585330714077300 | 0.0000008440502079 | 0.2680292304904580 | 0.1281777579497830 | 0.0000000000000009 | 0.0976772648082733 | 0.0000000000000000 | 0.0000000166613427 |
| | Best | -1.4999992233524900 | -0.7977041047646610 | -1.4999992233524900 | -1.4999992233524900 | -1.4999992233524900 | -1.4999992233524900 | -1.4999992233524900 | -1.5000000000000000 | -1.4999999997292700 |
| | Runtime | 33.809 | 17.940 | 37.986 | 20.333 | 42.488 | 36.037 | 18.930 | 41.848 | 0.510 |
| | | | | | | | | | | |
| F24 | Mean | -0.9166206788680230 | -0.3105071678265780 | -0.8406348096500680 | -0.8827152798835760 | -0.9431432797743700 | -1.2765515661973800 | -1.3127183561646500 | -1.5000000000000000 | -1.4999999478041200 |
| | Std. Dev. | 0.3917752367440500 | 0.2080317241440800 | 0.2000966365984320 | 0.3882445165494030 | 0.3184175870987750 | 0.3599594108130040 | 0.3158807699946290 | 0.0000000000000000 | 0.0000003682666585 |
| | Best | -1.5000000000003800 | -0.7976938356122860 | -1.4999926800631400 | -1.5000000000003800 | -1.5000000000003800 | -1.5000000000003800 | -1.5000000000003800 | -1.5000000000000000 | -1.4999999976445200 |
| | Runtime | 110.798 | 8.835 | 38.470 | 21.599 | 124.609 | 47.171 | 35.358 | 54.651 | 0.842 |
| | | | | | | | | | | |
| F25 | Mean | 0.0000000000000000 | 0.0000000000000000 | 0.0000000000000004 | 0.0000000000000000 | 0.0000041787372626 | 0.0000000000000000 | 0.0000000000000000 | 0.0000000000000000 | 0.0000000000000000 |
| | Std. Dev. | 0.0000000000000000 | 0.0000000000000000 | 0.0000000000000003 | 0.0000000000000000 | 0.0000161643637543 | 0.0000000000000000 | 0.0000000000000000 | 0.0000000000000000 | 0.0000000000000000 |
| | Best | 0.0000000000000000 | 0.0000000000000000 | 0.0000000000000001 | 0.0000000000000000 | 0.0000000000000000 | 0.0000000000000000 | 0.0000000000000000 | 0.0000000000000000 | 0.0000000000000000 |
| | Runtime | 25.358 | 1.340 | 19.689 | 1.142 | 31.632 | 4.090 | 0.813 | 35.662 | 2.890 |
| | | | | | | | | | | |
| F26 | Mean | -1.8210436836776800 | -1.7829268228561700 | -1.8210436836776800 | -1.8210436836776800 | -1.8210436836776800 | -1.8210436836776800 | -1.8210436836776800 | -1.8203821100000000 | -1.8210436836776800 |
| | Std. Dev. | 0.0000000000000009 | 0.1450583631808370 | 0.0000000000000009 | 0.0000000000000009 | 0.0000000000000009 | 0.0000000000000009 | 0.0000000000000009 | 0.0000000000000014 | 0.0000000000000005 |
| | Best | -1.8210436836776800 | -1.8210436836776800 | -1.8210436836776800 | -1.8210436836776800 | -1.8210436836776800 | -1.8210436836776800 | -1.8210436836776800 | -1.8203821095139300 | -1.8210436836776800 |
| | Runtime | 19.154 | 26.249 | 17.228 | 9.663 | 18.091 | 28.453 | 7.472 | 34.891 | 0.346 |
| | | | | | | | | | | |
| F27 | Mean | -4.6565646397053900 | -4.1008953007033700 | -4.6934684519571100 | -4.6893456932617100 | -4.6920941990586400 | -4.6884965299983800 | -4.6934684519571100 | -3.2820108350000000 | -4.5982757883767500 |
| | Std. Dev. | 0.0557021530063238 | 0.4951250481844850 | 0.0000000000000009 | 0.0125797149251589 | 0.0075270931220834 | 0.0272323381095561 | 0.0000000000000008 | 0.0000000000000023 | 0.1140777982812540 |
| | Best | -4.6934684519571100 | -4.6934684519571100 | -4.6934684519571100 | -4.6934684519571100 | -4.6934684519571100 | -4.6934684519571100 | -4.6934684519571100 | -3.2820108345268900 | -4.6934684286288800 |
| | Runtime | 38.651 | 10.956 | 17.663 | 14.915 | 25.843 | 38.446 | 11.971 | 45.085 | 0.530 |
| | | | | | | | | | | |
| F28 | Mean | -8.9717330307549300 | -7.6193507368464700 | -9.6601517156413500 | -9.6397230986132500 | -9.6400278592589600 | -9.6572038232921700 | -9.6601517156413500 | -6.2086254390000000 | -8.4871985036037100 |
| | Std. Dev. | 0.4927013165009220 | 0.7904830398850970 | 0.0000000000000008 | 0.0393668145094111 | 0.0437935551332868 | 0.0105890022905617 | 0.0000000000000007 | 0.0000000000000027 | 0.2867921564163950 |
| | Best | -9.5777818097208200 | -9.1383975057875100 | -9.6601517156413500 | -9.6601517156413500 | -9.6601517156413500 | -9.6601517156413500 | -9.6601517156413500 | -6.2086254392105500 | -8.9978275376597000 |
| | Runtime | 144.093 | 6.959 | 27.051 | 20.803 | 32.801 | 46.395 | 22.250 | 71.652 | 4.784 |
| | | | | | | | | | | |
| F29 | Mean | 0.0119687224560441 | 0.0788734736114700 | 0.0838440014038032 | 0.0154105130055856 | 0.0198686590210374 | 0.0140272066690658 | 0.0007283694780796 | 1.3116221610000000 | 0.0049933819581781 |
| | Std. Dev. | 0.0385628598040034 | 0.1426911799629180 | 0.0778327303965192 | 0.0308963906374663 | 0.0613698943155661 | 0.0328868042987376 | 0.0014793717464195 | 0.5590904820000000 | 0.0023147314691019 |
| | Best | 0.0000044608370213 | 0.0000000000000000 | 0.0129834451730589 | 0.0000000000000000 | 0.0000175219764526 | 0.0000000000000000 | 0.0000000000000000 | 1.0960146962658900 | 0.0007717562336873 |
| | Runtime | 359.039 | 17.056 | 60.216 | 35.044 | 316.817 | 92.412 | 191.881 | 34.697 | 0.875 |
| | | | | | | | | | | |
| F30 | Mean | 0.0000130718912008 | 0.0000000000000000 | 0.0002604330013462 | 0.0000000000000001 | 0.0458769685199585 | 0.0000002733806735 | 0.0000000028443186 | 0.0000000000000000 | 0.0000000000000000 |
| | Std. Dev. | 0.0000014288348929 | 0.0000000000000000 | 0.0000394921919294 | 0.0000000000000002 | 0.0620254411839524 | 0.0000001788830279 | 0.0000000033308990 | 0.0000000000000000 | 0.0000000000000000 |
| | Best | 0.0000095067504097 | 0.0000000000000000 | 0.0001682411286088 | 0.0000000000000000 | 0.0005277712020642 | 0.0000000944121661 | 0.0000000004769768 | 0.0000000000000000 | 0.0000000000000000 |
| | Runtime | 567.704 | 14.535 | 215.722 | 194.117 | 252.779 | 360.380 | 144.784 | 153.221 | 4.297 |
| | | | | | | | | | | |
| F31 | Mean | 0.0001254882834238 | 0.0000000000000000 | 0.0077905311094958 | 0.0020185116261490 | 0.0002674563703837 | 0.0000000000000000 | 0.0000000111676630 | 0.0071082040000000 | 0.0003936439985429 |
| | Std. Dev. | 0.0001503556280087 | 0.0000000000000000 | 0.0062425841086448 | 0.0077448684015362 | 0.0003044909265796 | 0.0000000000000000 | 0.0000000184322163 | 0.0000000000000000 | 0.0002001204487121 |
| | Best | 0.0000000156460198 | 0.0000000000000000 | 0.0003958766023752 | 0.0000000000000000 | 0.0000023064754605 | 0.0000000000000000 | 0.0000000000000000 | 0.0071082039505830 | 0.0000029388885444 |
| | Runtime | 250.248 | 12.062 | 34.665 | 48.692 | 227.817 | 220.886 | 149.882 | 43.098 | 8.902 |
| | | | | | | | | | | |
| F32 | Mean | 0.0003548345513179 | 0.0701619169853449 | 0.0250163252527030 | 0.0013010316180679 | 0.0019635752485802 | 0.0016730768406953 | 0.0019955316015528 | 0.0002254250000000 | 0.0000236661877907 |
| | Std. Dev. | 0.0001410817500914 | 0.0288760292572957 | 0.0077209314806873 | 0.0009952078711752 | 0.0043423828633839 | 0.0007330246909835 | 0.0009698942217908 | 0.0005270410000000 | 0.0000236731760330 |
| | Best | 0.0001014332605364 | 0.0299180701536354 | 0.0094647580732654 | 0.0001787238105452 | 0.0004206447422138 | 0.0005630852254632 | 0.0006084880639553 | 0.0000023800831017 | 0.0000050196149891 |
| | Runtime | 290.669 | 2.154 | 34.982 | 82.124 | 103.283 | 171.637 | 48.237 | 218.722 | 2.860 |
| | | | | | | | | | | |
| F33 | Mean | 25.6367602258676000 | 95.9799861204982000 | 0.0000000000000000 | 1.1276202647057400 | 0.6301407361590880 | 0.8622978494808570 | 0.0000000000000000 | 0.0000000000000000 | 0.0000000000000000 |
| | Std. Dev. | 8.2943512684216700 | 56.6919245985100000 | 0.0000000000000000 | 1.0688393637536800 | 0.8046401822326410 | 0.9323785263847000 | 0.0000000000000000 | 0.0000000000000000 | 0.0000000000000000 |
| | Best | 12.9344677422129000 | 29.8487565993415000 | 0.0000000000000000 | 0.0000000000000000 | 0.0000000000000000 | 0.0000000000000000 | 0.0000000000000000 | 0.0000000000000000 | 0.0000000000000000 |
| | Runtime | 76.083 | 2.740 | 4.090 | 7.635 | 18.429 | 23.594 | 5.401 | 2.266 | 3.516 |
| | | | | | | | | | | |
| F34 | Mean | 2.6757043114269700 | 0.3986623855035210 | 0.2856833465904130 | 1.0630996944802500 | 5.7631786582751800 | 1.2137377447007000 | 0.3986623854300930 | 0.0000154715000000 | 28.8334517794009000 |
| | Std. Dev. | 12.3490058210004000 | 1.2164328621946200 | 0.6247370987465170 | 1.7930895051734300 | 13.9484817304201000 | 1.8518519388285700 | 1.2164328622195200 | 0.0000022373400000 | 0.0144695690509943 |
| | Best | 0.0042535368984501 | 0.0000000000000000 | 0.0004266049929880 | 0.0000000000000000 | 0.0268003205820685 | 0.0001448955835246 | 0.0000000000000000 | 0.0000118803557196 | 28.8053841187578000 |
| | Runtime | 559.966 | 9.462 | 35.865 | 23.278 | 187.894 | 268.449 | 34.681 | 7.250 | 8.431 |
| | | | | | | | | | | |
| F35 | Mean | 0.0000000000000000 | 0.4651202457398910 | 0.0000000000000000 | 0.0038863639514140 | 0.0019431819755029 | 0.0006477273251676 | 0.0000000000000000 | 0.0000000000000000 | 0.0000000000000000 |
| | Std. Dev. | 0.0000000000000000 | 0.0933685176073728 | 0.0000000000000000 | 0.0048411743884718 | 0.0039528023354469 | 0.0024650053428137 | 0.0000000000000000 | 0.0000000000000000 | 0.0000000000000000 |
| | Best | 0.0000000000000000 | 0.0097159098775144 | 0.0000000000000000 | 0.0000000000000000 | 0.0000000000000000 | 0.0000000000000000 | 0.0000000000000000 | 0.0000000000000000 | 0.0000000000000000 |
| | Runtime | 18.163 | 24.021 | 7.861 | 4.216 | 8.304 | 5.902 | 1.779 | 33.155 | 2.943 |
| | | | | | | | | | | |
| F36 | Mean | -7684.6104757783800000 | -6835.1836730901400000 | -12569.4866181730000 | -12304.9743375341000 | -12210.8815698372000 | -12549.746895737300000 | -12569.486618173000000 | -12569.3622100000000000 | -12569.4866181730000000 |
| | Std. Dev. | 745.3954005014180000 | 750.7338055436110000 | 0.0000000000022659 | 221.4322514436480000 | 205.9313376284770000 | 44.8939348779747000 | 0.0000000000024122 | 0.0000000273871000 | 0.0000000000018828 |
| | Best | -8912.8855854978200000 | -8340.0386911070600000 | -12569.4866181730000 | -12569.4866181730000 | -12569.4866181730000 | -12569.486618173000000 | -12569.486618173000000 | -12569.3622054081000000 | -12569.4866181730000000 |
| | Runtime | 307.427 | 3.174 | 19.225 | 10.315 | 31.499 | 34.383 | 11.069 | 2.306 | 10.825 |
| | | | | | | | | | | |
| F37 | Mean | 0.0000000000000000 | 0.0000000000000000 | 14.5668734126948000 | 0.0000000000000000 | 6.4655746330439100 | 0.0000000000000000 | 0.0000000000000000 | 0.0000000000000000 | 0.0000000000000000 |
| | Std. Dev. | 0.0000000000000000 | 0.0000000000000000 | 8.7128443012950300 | 0.0000000000000000 | 8.2188901353055800 | 0.0000000000000000 | 0.0000000000000000 | 0.0000000000000000 | 0.0000000000000000 |
| | Best | 0.0000000000000000 | 0.0000000000000000 | 4.0427699323673400 | 0.0000000000000000 | 0.1816624029553790 | 0.0000000000000000 | 0.0000000000000000 | 0.0000000000000000 | 0.0000000000000000 |
| | Runtime | 543.180 | 3.370 | 111.841 | 19.307 | 179.083 | 109.551 | 57.294 | 100.947 | 5.112 |
| | | | | | | | | | | |
| F38 | Mean | 0.0000000000000000 | 0.0000000000000000 | 0.0000000000000005 | 0.0000000000000000 | 0.0000000000000000 | 0.0000000000000000 | 0.0000000000000000 | 0.0000000000000000 | 0.0000000000000000 |
| | Std. Dev. | 0.0000000000000000 | 0.0000000000000000 | 0.0000000000000001 | 0.0000000000000000 | 0.0000000000000000 | 0.0000000000000000 | 0.0000000000000000 | 0.0000000000000000 | 0.0000000000000000 |
| | Best | 0.0000000000000000 | 0.0000000000000000 | 0.0000000000000003 | 0.0000000000000000 | 0.0000000000000000 | 0.0000000000000000 | 0.0000000000000000 | 0.0000000000000000 | 0.0000000000000000 |
| | Runtime | 163.188 | 2.558 | 20.588 | 1.494 | 12.563 | 5.627 | 3.208 | 47.009 | 6.738 |
| | | | | | | | | | | |
| F39 | Mean | -10.1061873621653000 | -5.2607563471326400 | -10.5364098166920000 | -10.3130437162426000 | -10.3130437162026000 | -10.5364098166921000 | -10.5364098166921000 | -10.5063235800000000 | -10.5364098166920000 |
| | Std. Dev. | 1.6679113661236400 | 3.6145751818694000 | 0.0000000000000023 | 1.2234265179812200 | 1.2234265179736500 | 0.0000000000000016 | 0.0000000000000018 | 0.0000000025211900 | 0.0000000000000055 |
| | Best | -10.5364098166921000 | -10.5364098166921000 | -10.5364098166920000 | -10.5364098166921000 | -10.5364098166920000 | -10.5364098166921000 | -10.5364098166920000 | -10.5063235792920000 | -10.5364098166920000 |
| | Runtime | 31.018 | 11.024 | 16.015 | 8.345 | 37.275 | 28.031 | 7.045 | 55.666 | 0.892 |
| | | | | | | | | | | |
| F40 | Mean | -9.5373938082045500 | -5.7308569926624600 | -10.1531996790582000 | -9.5656135761215700 | -10.1531996790582000 | -9.9847854277673500 | -10.1531996790582000 | -10.1529842600000000 | -10.1531996790582000 |
| | Std. Dev. | 1.9062127067994200 | 3.5141202468383400 | 0.0000000000000055 | 1.8315977756329900 | 0.0000000000000076 | 0.9224428443735560 | 0.0000000000000072 | 0.0000000000542921 | 0.0000000000000000 |
| | Best | -10.1531996790582000 | -10.1531996790582000 | -10.1531996790582000 | -10.1531996790582000 | -10.1531996790582000 | -10.1531996790582000 | -10.1531996790582000 | -10.1529842649756000 | -10.1531996790582000 |
| | Runtime | 25.237 | 11.177 | 11.958 | 7.947 | 30.885 | 25.569 | 6.864 | 51.507 | 0.860 |
| | | | | | | | | | | |
| F41 | Mean | -10.4029405668187000 | -6.8674070870953700 | -10.4029405668187000 | -9.1615813354737300 | -10.4029405668187000 | -10.4029405668187000 | -10.4029405668187000 | -10.3988303400000000 | -10.4029405668187000 |
| | Std. Dev. | 0.0000000000000018 | 3.6437803702691000 | 0.0000000000000006 | 2.8277336448396200 | 0.0000000000000010 | 0.0000000000000018 | 0.0000000000000017 | 0.0000000001978980 | 0.0000000000000000 |
| | Best | -10.4029405668187000 | -10.4029405668187000 | -10.4029405668187000 | -10.4029405668187000 | -10.4029405668187000 | -10.4029405668187000 | -10.4029405668187000 | -10.3988303385534000 | -10.4029405668187000 |
| | Runtime | 21.237 | 11.482 | 14.911 | 8.547 | 31.207 | 27.064 | 8.208 | 53.190 | 0.395 |
| | | | | | | | | | | |
| F42 | Mean | -186.7309073569880000 | -81.5609772893002000 | -186.730908831024000 | -186.730908831024000 | -186.730908831024000 | -186.7309088310240000 | -186.7309088310240000 | -186.2926481000000000 | -186.7309088310240000 |
| | Std. Dev. | 0.0000046401472660 | 66.4508342743478000 | 0.0000000000000236 | 0.0000000000000388 | 0.0000000000000279 | 0.0000000000000377 | 0.0000000000000224 | 0.0000000000000578 | 0.0000000000000291 |
| | Best | -186.7309088310240000 | -186.7309088310240000 | -186.730908831024000 | -186.730908831024000 | -186.730908831024000 | -186.7309088310240000 | -186.7309088310240000 | -186.2926480689880000 | -186.7309088310240000 |
| | Runtime | 19.770 | 25.225 | 13.342 | 8.213 | 20.344 | 27.109 | 9.002 | 31.766 | 2.466 |
| | | | | | | | | | | |
| F43 | Mean | -1.0316284534898800 | -1.0044229658530100 | -1.0316284534898800 | -1.0316284534898800 | -1.0316284534898800 | -1.0316284534898800 | -1.0316284534898800 | -1.0304357800000000 | -1.0316284534898800 |
| | Std. Dev. | 0.0000000000000005 | 0.1490105926664260 | 0.0000000000000005 | 0.0000000000000005 | 0.0000000000000005 | 0.0000000000000005 | 0.0000000000000005 | 0.0014911900000000 | 0.0000000000000000 |
| | Best | -1.0316284534898800 | -1.0316284534898800 | -1.0316284534898800 | -1.0316284534898800 | -1.0316284534898800 | -1.0316284534898800 | -1.0316284534898800 | -1.0314500753985900 | -1.0316284534898800 |
| | Runtime | 16.754 | 24.798 | 11.309 | 7.147 | 18.564 | 27.650 | 5.691 | 39.897 | 0.391 |
| | | | | | | | | | | |
| F44 | Mean | 0.0000000000000000 | 0.0000000000000000 | 0.0000000000000004 | 0.0000000000000000 | 0.0000000000000000 | 0.0000000000000000 | 0.0000000000000000 | 0.0000000000000000 | 0.0000000000000000 |
| | Std. Dev. | 0.0000000000000000 | 0.0000000000000000 | 0.0000000000000001 | 0.0000000000000000 | 0.0000000000000000 | 0.0000000000000000 | 0.0000000000000000 | 0.0000000000000000 | 0.0000000000000000 |
| | Best | 0.0000000000000000 | 0.0000000000000000 | 0.0000000000000003 | 0.0000000000000000 | 0.0000000000000000 | 0.0000000000000000 | 0.0000000000000000 | 0.0000000000000000 | 0.0000000000000000 |
| | Runtime | 159.904 | 2.321 | 21.924 | 1.424 | 14.389 | 5.920 | 3.302 | 174.577 | 4.791 |
| | | | | | | | | | | |
| F45 | Mean | 2.3000000000000000 | 0.0666666666666667 | 0.0000000000000000 | 0.9000000000000000 | 0.0000000000000000 | 0.0000000000000000 | 0.0000000000000000 | 0.0000538870000000 | 0.0000000000000000 |
| | Std. Dev. | 1.8597367258983700 | 0.2537081317024630 | 0.0000000000000000 | 3.0211895350832500 | 0.0000000000000000 | 0.0000000000000000 | 0.0000000000000000 | 0.0000000005399890 | 0.0000000000000000 |
| | Best | 0.0000000000000000 | 0.0000000000000000 | 0.0000000000000000 | 0.0000000000000000 | 0.0000000000000000 | 0.0000000000000000 | 0.0000000000000000 | 0.0000538860819891 | 0.0000000000000000 |
| | Runtime | 57.276 | 1.477 | 1.782 | 2.919 | 3.042 | 4.307 | 0.883 | 2.215 | 14.850 |
| | | | | | | | | | | |
| F46 | Mean | 0.1333333333333330 | 0.2666666666666670 | 0.0000000000000000 | 0.0000000000000000 | 0.2000000000000000 | 0.0000000000000000 | 0.0000000000000000 | -0.0153463301609662 | 0.1276607794072780 |
| | Std. Dev. | 0.3457459036417600 | 0.9444331755018490 | 0.0000000000000000 | 0.0000000000000000 | 0.4068381021724860 | 0.0000000000000000 | 0.0000000000000000 | 0.0000000000000000 | 0.0737285129670879 |
| | Best | 0.0000000000000000 | 0.0000000000000000 | 0.0000000000000000 | 0.0000000000000000 | 0.0000000000000000 | 0.0000000000000000 | 0.0000000000000000 | -0.0153463301609662 | 0.0148704298643416 |
| | Runtime | 20.381 | 2.442 | 1.700 | 1.074 | 6.142 | 4.319 | 0.764 | 31.068 | 2.890 |
| | | | | | | | | | | |
| F47 | Mean | 0.0000000000000000 | 0.0000000000000000 | 0.0000000000000005 | 0.0000000000000000 | 0.0000000000000000 | 0.0000000000000000 | 0.0000000000000000 | 0.0000000000000000 | 0.0000000000000000 |
| | Std. Dev. | 0.0000000000000000 | 0.0000000000000000 | 0.0000000000000000 | 0.0000000000000000 | 0.0000000000000000 | 0.0000000000000000 | 0.0000000000000000 | 0.0000000000000000 | 0.0000000000000000 |
| | Best | 0.0000000000000000 | 0.0000000000000000 | 0.0000000000000003 | 0.0000000000000000 | 0.0000000000000000 | 0.0000000000000000 | 0.0000000000000000 | 0.0000000000000000 | 0.0000000000000000 |
| | Runtime | 564.178 | 2.565 | 24.172 | 1.870 | 15.948 | 6.383 | 4.309 | 31.296 | 4.685 |
| | | | | | | | | | | |
| F48 | Mean | -50.0000000000002000 | -50.0000000000002000 | -49.9999999999997000 | -50.0000000000002000 | -49.4789234062579000 | -50.0000000000002000 | -50.0000000000002000 | -44.7416748700000000 | -50.0000000000000000 |
| | Std. Dev. | 0.0000000000000361 | 0.0000000000000268 | 0.0000000000001408 | 0.0000000000000354 | 1.3150773145311700 | 0.0000000000000268 | 0.0000000000000361 | 0.0000000000000217 | 0.0000000000000000 |
| | Best | -50.0000000000002000 | -50.0000000000002000 | -50.0000000000001000 | -50.0000000000002000 | -49.9999994167392000 | -50.0000000000002000 | -50.0000000000002000 | -44.7416748706606000 | -50.0000000000000000 |
| | Runtime | 24.627 | 8.337 | 22.480 | 8.623 | 142.106 | 36.804 | 7.747 | 52.486 | 0.806 |
| | | | | | | | | | | |
| F49 | Mean | -210.0000000000010000 | - 210.0000000000030000 | -209.999999999947000 | -210.000000000003000 | -199.592588547503000 | -210.0000000000030000 | -210.0000000000030000 | -150.5540859185450000 | -210.0000000000000000 |
| | Std. Dev. | 0.0000000000009434 | 0.0000000000003702 | 0.0000000000138503 | 0.0000000000008251 | 9.6415263953591700 | 0.0000000000004625 | 0.0000000000003950 | 0.0000000000000000 | 0.0000000000000000 |
| | Best | -210.0000000000030000 | - 210.0000000000030000 | -209.999999999969000 | -210.000000000004000 | -209.985867409029000 | -210.0000000000040000 | -210.0000000000040000 | -150.5540859185450000 | -210.0000000000000000 |
| | Runtime | 48.580 | 5.988 | 36.639 | 11.319 | 187.787 | 54.421 | 11.158 | 70.887 | 10.962 |
| | | | | | | | | | | |
| F50 | Mean | 0.0000000000000000 | 0.0000000000000000 | 0.0000000402380424 | 0.0000000000000000 | 0.0000000001597805 | 0.0000000000000000 | 0.0000000000000000 | 0.0000000000000000 | 0.0000000000000000 |
| | Std. Dev. | 0.0000000000000000 | 0.0000000000000000 | 0.0000002203520334 | 0.0000000000000000 | 0.0000000006266641 | 0.0000000000000000 | 0.0000000000000000 | 0.0000000000000000 | 0.0000000000000000 |
| | Best | 0.0000000000000000 | 0.0000000000000000 | 0.0000000000000210 | 0.0000000000000000 | 0.0000000000000000 | 0.0000000000000000 | 0.0000000000000000 | 0.0000000000000000 | 0.0000000000000000 |
| | Runtime | 86.369 | 1.868 | 86.449 | 1.412 | 157.838 | 4.930 | 5.702 | 33.573 | 2.15 |

Table 4: Statistical solutions to Test 2 Problems using PSO, CMAES, ABC, CLPSO, SADE, BSA, IA and Multi-CI
(Mean = Mean solution; Std. Dev. = Standard-deviation of mean solution; Best = Best solution; Runtime = Mean runtime in seconds)

| Problem | Statistics | PSO2011 | CMAES | ABC | JDE | CLPSO | SADE | BSA | IA | Multi CI |
|---|---|---|---|---|---|---|---|---|---|---|
| F51 | Mean | -450.0000000000000000 | -450.0000000000000000 | -450.0000000000000000 | -450.0000000000000000 | -450.0000000000000000 | -450.0000000000000000 | -450.0000000000000000 | -447.6018854297170000 | -450.0000000000000000 |
| | Std. Dev. | 0.0000000000000000 | 0.0000000000000000 | 0.0000000000000000 | 0.0000000000000000 | 0.0000000000000000 | 0.0000000000000000 | 0.0000000000000000 | 89.3142986500000000 | 0.0000000000000000 |
| | Best | -450.0000000000000000 | -450.0000000000000000 | -450.0000000000000000 | -450.0000000000000000 | -450.0000000000000000 | -450.0000000000000000 | -450.0000000000000000 | -450.0000000000000000 | -450.0000000000000000 |
| | Runtime | 212.862 | 23.146 | 113.623 | 118.477 | 167.675 | 154.232 | 140.736 | 30.282 | 28.930 |
| | | | | | | | | | | |
| F52 | Mean | -450.0000000000000000 | -450.0000000000000000 | -449.9999999999220000 | -450.0000000000000000 | -418.8551838547760000 | -450.0000000000000000 | -450.0000000000000000 | -449.9967727000000000 | -450.0000000000000000 |
| | Std. Dev. | 0.0000000000000350 | 0.0000000000000000 | 0.0000000002052730 | 0.0000000000000615 | 51.0880511039985000 | 0.0000000000000000 | 0.0000000000000259 | 0.0176705780000000 | 0.0000000000000000 |
| | Best | -450.0000000000000000 | -450.0000000000000000 | -449.9999999999970000 | -450.0000000000000000 | -449.4789299923810000 | -450.0000000000000000 | -450.0000000000000000 | -450.0000000000000000 | -450.0000000000000000 |
| | Runtime | 230.003 | 23.385 | 648.784 | 139.144 | 1462.706 | 185.965 | 243.657 | 48.003 | 74.497 |
| | | | | | | | | | | |
| F53 | Mean | -44.5873911956554000 | -450.0000000000000000 | 387131.24412139700000 | -197.9999999999850000 | 62142.8213760465000000 | 245.0483283713550000 | -449.9999567867430000 | -449.7873452000000000 | 67397.8550894185000000 |
| | Std. Dev. | 458.5794120016290000 | 0.0000000000000000 | 166951.73365926400000 | 391.5169437474990000 | 34796.1785167236000000 | 790.6056596723160000 | 0.0001175386756044 | 0.0000000000001734 | 102453.8929479460000000 |
| | Best | -443.9511286079800000 | -450.0000000000000000 | 165173.18530956000000 | -449.9999999999990000 | 17306.9066792474000000 | -421.4054944641620000 | -450.0000000000000000 | -450.0000000000000000 | 11750.9831357565000000 |
| | Runtime | 2658.937 | 35.464 | 240.094 | 1017.557 | 1789.643 | 1808.954 | 1883.713 | 52.463 | 50.447 |
| | | | | | | | | | | |
| F54 | Mean | -450.0000000000000000 | 77982.4567046980000000 | 140.4509447125110000 | -414.0000000000000000 | -178.8320689185280000 | -450.0000000000000000 | -450.0000000000000000 | -388.7807630000000000 | -450.0000000000000000 |
| | Std. Dev. | 0.0000000000000460 | 131376.7365456010000000 | 217.2646715063190000 | 55.9309919639279000 | 394.8667499339530000 | 0.0000000000000000 | 0.0000000000000259 | 1.1928333530000000 | 0.0000000000000000 |
| | Best | -450.0000000000000000 | -450.0000000000000000 | -324.3395691109350000 | -450.0000000000000000 | -447.9901256558030000 | -450.0000000000000000 | -450.0000000000000000 | -389.7573633109500000 | -450.0000000000000000 |
| | Runtime | 247.256 | 32.726 | 209.188 | 143.767 | 1248.616 | 185.438 | 347.167 | 46.072 | 83.596 |
| | | | | | | | | | | |
| F55 | Mean | -310.0000000000000000 | -310.0000000000000000 | -291.5327549384120000 | -271.0000000000000000 | 333.4108259915760000 | -309.9999999999960000 | -309.9999999999980000 | -310.8207993000000000 | -310.0000000000000000 |
| | Std. Dev. | 0.0000000000000000 | 0.0000000000000000 | 17.6942171217937000 | 60.5919079609218000 | 512.6920837704510000 | 0.0000000000133965 | 0.0000000000023443 | 0.0208030240000000 | 0.0000000000000000 |
| | Best | -310.0000000000000000 | -310.0000000000000000 | -307.7611364354020000 | -310.0000000000000000 | -309.9740055344430000 | -310.0000000000000000 | -310.0000000000000000 | -310.8367924750510000 | -310.0000000000000000 |
| | Runtime | 241.517 | 39.293 | 205.568 | 134.078 | 1481.686 | 210.684 | 386.633 | 44.84710031 | 120.725 |
| | | | | | | | | | | |
| F56 | Mean | 393.4959999056240000 | 390.5315438816460000 | 391.2531452421960000 | 231.3986579112350000 | 405.5233436479650000 | 390.2657719408230000 | 390.1328859704120000 | 390.8036739982730000 | 392.3754700583880000 |
| | Std. Dev. | 16.0224965900462000 | 1.3783433976378300 | 3.7254660805238600 | 247.2968415284400000 | 10.7480096852869000 | 1.0114275384776600 | 0.7278464357038200 | 0.0000000000000000 | 0.6527183145462900 |
| | Best | 390.0000000000150000 | 390.0000000000000000 | 390.0101471658490000 | -140.0000000000000000 | 390.5776683413440000 | 390.0000000000000000 | 390.0000000000000000 | 390.8036739982730000 | 391.2787609196740000 |
| | Runtime | 1178.079 | 27.894 | 159.762 | 153.715 | 1441.859 | 1214.303 | 290.236 | 45.632 | 88.645 |
| | | | | | | | | | | |
| F57 | Mean | 1091.0644335162500000 | 1087.2645466786700000 | 1087.0459486286000000 | 1141.0459486286000000 | 1087.0459486286000000 | 1087.0459486286000000 | 1087.0459486286000000 | 1087.2265890000000000 | 1087.2402022380600000 |
| | Std. Dev. | 3.4976948942723200 | 0.5365230018001780 | 0.0000000000005585 | 83.8964879458918000 | 0.0000000000004264 | 0.0000000000004814 | 0.0000000000004428 | 0.0019192200000000 | 0.2217515717485750 |
| | Best | 1087.0696772583000000 | 1087.0459486286000000 | 1087.0459486286000000 | 1087.0459486286000000 | 1087.0459486286000000 | 1087.0459486286000000 | 1087.0459486286000000 | 1087.2262037455100000 | 1087.0546352983500000 |
| | Runtime | 334.064 | 37.047 | 180.472 | 159.922 | 267.342 | 259.760 | 332.132 | 52.621 | 145.678 |
| | | | | | | | | | | |
| F58 | Mean | -119.8190232990920000 | -119.9261073509850000 | -119.7446063439080000 | -119.4450938018030000 | -119.9300269839980000 | -119.7727713703720000 | -119.8356122057440000 | -119.6006412865410000 | -119.9717811854090000 |
| | Std. Dev. | 0.0720107560874199 | 0.1554021446157740 | 0.0623866434489108 | 0.0927418223065644 | 0.0417913553101429 | 0.1248514853682450 | 0.0704515460477787 | 0.0000000000000434 | 0.0536174052618470 |
| | Best | -119.9302772694110000 | -120.0000000000000000 | -119.8779554779730000 | -119.6575717927190000 | -119.9756745390830000 | -119.9999999999980000 | -119.9802847896350000 | -119.6006412865410000 | -119.9905751250590000 |
| | Runtime | 602.507 | 49.209 | 265.319 | 160.806 | 1586.286 | 648.489 | 717.375 | 52.56165118 | 117.872 |
| | | | | | | | | | | |
| F59 | Mean | -324.6046006320200000 | -306.5782069681560000 | -330.0000000000000000 | -329.8673387923880000 | -329.4361898676470000 | -329.9668346980970000 | -330.0000000000000000 | -327.1635938000000000 | -323.0352916375860000 |
| | Std. Dev. | 2.5082306041521000 | 21.9475396048756000 | 0.0000000000000000 | 0.3440030182812760 | 0.6229063711904190 | 0.1816538397880230 | 0.0000000000000000 | 0.0000000000001156 | 2.3797197171188800 |
| | Best | -329.0050409429070000 | -327.0151228287200000 | -330.0000000000000000 | -330.0000000000000000 | -330.0000000000000000 | -330.0000000000000000 | -330.0000000000000000 | -327.163593801473 | -328.0100818858130000 |
| | Runtime | 982.449 | 22.237 | 111.629 | 128.494 | 162.873 | 155.645 | 176.994 | 45.867 | 70.902 |
| | | | | | | | | | | |
| F60 | Mean | -324.3311322538170000 | -314.7871102989330000 | -306.7949047862760000 | -319.6763749798700000 | -321.7278926895280000 | -322.9689591871600000 | -319.2544515903510000 | -335.0171647000000000 | -322.5378095907100000 |
| | Std. Dev. | 3.0072222933667300 | 8.3115989308305500 | 5.1787864195870400 | 4.9173541245304800 | 1.8971778613701300 | 2.8254645254663600 | 3.3091959975390800 | 10.6369134000000000 | 2.4434060592713200 |
| | Best | -327.1650513120000000 | -327.0151228287200000 | -318.9403196374510000 | -326.0201637716270000 | -326.1788303102740000 | -328.0100818858130000 | -325.0252097523530000 | -347.2509173436740000 | -326.0032742097850000 |
| | Runtime | 1146.013 | 29.860 | 259.258 | 179.039 | 1594.096 | 210.534 | 420.851 | 54.661 | 96.444 |
| | | | | | | | | | | |
| F61 | Mean | 92.5640111212146000 | 90.7642785704506000 | 94.8428485804138000 | 93.2972315784963000 | 94.6109567642977000 | 91.6859083842723000 | 92.3519494286347000 | 92.0170440500000000 | 90.0000910864369000 |
| | Std. Dev. | 1.5827416781636900 | 26.4613831425879000 | 0.6869412813090850 | 1.8766951726453600 | 0.6689129174038950 | 0.9033073777915270 | 1.0901581870340800 | 0.000000000000014453 | 0.3262402308775350 |
| | Best | 90.1142082473923000 | -45.0054133586912000 | 93.1500794016147000 | 91.0295373630387000 | 92.9690673344598000 | 90.1363685040678000 | 90.2628852415150000 | 92.0170440535006000 | 90.0000665813171000 |
| | Runtime | 1310.457 | 44.217 | 308.501 | 282.150 | 1421.545 | 506.829 | 1771.860 | 60.350 | 182.115 |
| | | | | | | | | | | |
| F62 | Mean | 18611.314225480900000 | -70.0486708747625000 | -337.3273080760500000 | 400.3240208136310000 | -447.8870804905020000 | -394.5206365378250000 | -437.1125728026770000 | -410.1361631000000000 | -450.8165121271770000 |
| | Std. Dev. | 12508.786612631600000 | 637.4585182420270000 | 56.5730759032367000 | 688.3344299264300000 | 11.8934815947019000 | 128.6353424718180000 | 20.3541618366546000 | 34.8795385900000000 | 4.6244223495662800 |
| | Best | 4568.3350537809200000 | -460.0000000000000000 | -449.1707421778360000 | -434.8788220982740000 | -459.6890294276810000 | -460.0000000000000000 | -459.1772521346520000 | -421.5672584975600000 | -459.9993498733460000 |
| | Runtime | 2381.974 | 34.857 | 232.916 | 202.941 | 1636.440 | 1277.975 | 1466.985 | 48.480 | 143.358 |
| | | | | | | | | | | |
| F63 | Mean | -129.2373581503910000 | -128.7850616923410000 | -129.8343428775830000 | -129.6294851450880000 | -129.8382867796110000 | -129.7129164862680000 | -129.8981409848090000 | -122.2126680000000000 | -129.3840699141840000 |
| | Std. Dev. | 0.5986210944493790 | 0.6157633658946230 | 0.0408016481905455 | 0.1054759371085400 | 0.0372256921835666 | 0.0875456568200232 | 0.0682328484314248 | 0.0000000000000434 | 0.0756137409573255 |
| | Best | -129.6861385930680000 | -129.5105509483130000 | -129.9098920058450000 | -129.8125711770830000 | -129.9098505660780000 | -129.8717592632560000 | -129.9901230990300000 | -122.2126679617240000 | -129.5432045245170000 |
| | Runtime | 2183.218 | 25.496 | 205.194 | 186.347 | 1526.365 | 660.986 | 1064.114 | 46.260 | 170.218 |
| | | | | | | | | | | |
| F64 | Mean | -298.2835926212850000 | -295.1290938304830000 | -296.9323391084610000 | -296.8839733969750000 | -297.5119726691150000 | -297.8403738182600000 | -297.5359077431460000 | -295.4721554000000000 | -297.3117952188210000 |
| | Std. Dev. | 0.5587676271753680 | 0.1634039984609270 | 0.2251930667702880 | 0.4330673614598290 | 0.3440115280624180 | 0.4536801689800720 | 0.4085859316264990 | 0.1118191570000000 | 0.3262896571402280 |
| | Best | -299.6022022972560000 | -295.7382222729600000 | -297.4659619544820000 | -297.8411886637500000 | -298.3030560759620000 | -299.2417795907860000 | -298.3869295150680000 | -295.6307146941910000 | -297.7411404965860000 |
| | Runtime | 2517.138 | 32.084 | 262.533 | 334.888 | 1615.452 | 1289.814 | 1953.289 | 55.118 | 85.817 |
| | | | | | | | | | | |
| F65 | Mean | 417.4613663019860000 | 492.5045364088000000 | 120.0000000000000000 | 326.6601114362900000 | 131.3550392249760000 | 234.2689845349590000 | 120.0000000000000000 | 120.0000000000000000 | 211.7934679874670000 |
| | Std. Dev. | 153.9215808771580000 | 181.5709657779580000 | 0.0000000000000188 | 174.6877238188330000 | 26.1407360548431000 | 150.7595974059750000 | 0.0000000000000000 | 0.0000000000000000 | 161.3764717430970000 |
| | Best | 120.0000000000000000 | 262.7619554120320000 | 120.0000000000000000 | 120.0000000000000000 | 120.0000000000000000 | 120.0000000000000000 | 120.0000000000000000 | 120.0000000000000000 | 204.1733303990720000 |
| | Runtime | 3156.336 | 239.823 | 2285.787 | 1834.967 | 3210.655 | 1932.016 | 2351.478 | 69.052 | 662.766 |
| | NFE | | | | | | | | | |
| F66 | Mean | 221.4232628350220000 | 455.1151684594550000 | 258.8582688922670000 | 231.1806131539990000 | 231.5547154800990000 | 222.0256674919140000 | 234.4843380488580000 | 276.3946208000000000 | 223.0150462881420000 |
| | Std. Dev. | 12.2450207482898000 | 254.3583511786970000 | 11.8823213189685000 | 13.5473380962764000 | 11.5441451076421000 | 6.1841489800660300 | 8.9091119100451100 | 19.2196655800000000 | 5.5838205188784200 |
| | Best | 181.5746616282570000 | 120.0000000000000000 | 235.6600739998890000 | 210.3582705649860000 | 214.7661703584830000 | 206.4520786020840000 | 219.6244910167680000 | 259.8700033222460000 | 215.3853171009670000 |
| | Runtime | 4242.280 | 202.808 | 2237.308 | 1824.388 | 8649.998 | 2970.950 | 8270.920 | 252.234 | 334.256 |
| | | | | | | | | | | |
| F67 | Mean | 217.3338617866620000 | 681.0349114021570000 | 265.0370119084380000 | 228.7309024901770000 | 240.3635189964930000 | 221.1801916743850000 | 228.3769828342800000 | 201.0516618000000000 | 222.0150462881420000 |
| | Std. Dev. | 20.6685850658838000 | 488.0618274343640000 | 12.4033917090208000 | 12.3682716268631000 | 14.8435137485293000 | 5.7037006844690500 | 8.7086794471239900 | 2.4309010810000000 | 4.5838205188784200 |
| | Best | 120.0000000000000000 | 223.0782617790520000 | 241.9810089596350000 | 181.6799927773160000 | 221.3817133141830000 | 209.2509748304710000 | 204.6479138174220000 | 197.8966349103590000 | 215.3853171009670000 |
| | Runtime | 8208.697 | 197.497 | 2159.392 | 5873.112 | 4599.027 | 5938.879 | 8189.243 | 254.253 | 294.256 |
| | | | | | | | | | | |
| F68 | Mean | 668.9850326105730000 | 926.9488078829420000 | 513.8925774904480000 | 743.9859973770210000 | 892.4391527217660000 | 845.4504613493740000 | 587.5732354221340000 | 310.0161021000000000 | 366.0263626038670000 |
| | Std. Dev. | 275.8071370273340000 | 174.1027182659660000 | 31.0124861524005000 | 175.6497294240330000 | 79.1422224454971000 | 120.8505129523180000 | 250.0556329707140000 | 0.0370586450000000 | 231.2038781149790000 |
| | Best | 310.0000000000000000 | 310.0000000000000000 | 444.4692044973030000 | 310.0000000000000000 | 738.3764781625320000 | 310.0000000000000000 | 310.0000000000000000 | 310.0014955442130000 | 310.0622831487350000 |
| | Runtime | 3687.235 | 251.155 | 2445.259 | 1777.638 | 8398.690 | 3073.274 | 4554.102 | 253.064 | 310.288 |
| | | | | | | | | | | |
| F69 | Mean | 708.2979222913040000 | 831.2324139697050000 | 500.5478931040730000 | 776.5150806087790000 | 863.8926908090610000 | 809.7183195902260000 | 587.6511686191670000 | 310.0029796000000000 | 810.0062247333440000 |
| | Std. Dev. | 256.2419561521300000 | 250.1848775931620000 | 31.2240894705539000 | 160.7307526692470000 | 96.5618989087194000 | 147.3158109824600000 | 236.1141037692630000 | 0.0082796490000000 | 223.3247693967360000 |
| | Best | 310.0000000000000000 | 310.0000000000000000 | 407.3155842366960000 | 363.8314566805740000 | 493.0042540796450000 | 310.0000000000000000 | 310.0000000000000000 | 310.0000285440690000 | 310.0000000000000000 |
| | Runtime | 5258.509 | 222.015 | 2341.791 | 1849.670 | 9909.479 | 3213.601 | 4764.968 | 291.084 | 281.124 |
| | | | | | | | | | | |
| F70 | Mean | 711.2970397614200000 | 876.9306188768990000 | 483.2984167460740000 | 761.2954767038960000 | 844.6391674419360000 | 810.5227124472170000 | 612.0906184834040000 | 310.0041570000000000 | 660.0000000106290000 |
| | Std. Dev. | 258.9317052508320000 | 289.7296413284470000 | 99.3976740616107000 | 163.4084080635650000 | 113.6848457105400000 | 104.7139423525340000 | 249.5599278421970000 | 0.0128812140000000 | 202.7587641256140000 |
| | Best | 310.0000000000000000 | 310.0000000000000000 | 155.5049931377980000 | 363.8314568648180000 | 489.0742585970560000 | 310.0000000000000000 | 310.0000000000000000 | 310.0002219576930000 | 310.0000000051680000 |
| | Runtime | 4346.055 | 228.619 | 2250.917 | 1900.279 | 9988.261 | 2818.575 | 4945.132 | 268.701 | 440.520 |
| | | | | | | | | | | |
| F71 | Mean | 1117.8857079625100000 | 1258.1065536572400000 | 659.5351969346130000 | 959.3735119754180000 | 911.4640642691360000 | 990.8546718748010000 | 836.1411004458200000 | 577.7786170000000000 | 760.0000063070120000 |
| | Std. Dev. | 311.0011859260640000 | 359.7382897536570000 | 98.5410511961986000 | 240.5568407069990000 | 238.3180009803040000 | 235.1014092849970000 | 128.9346234954740000 | 1.8288684190000000 | 105.4092619871220000 |
| | Best | 560.0000000000000000 | 660.0000000000000000 | 560.0001912324020000 | 660.0000000000000000 | 560.0000121795840000 | 660.0000000000000000 | 560.0000000000000000 | 574.8590032551840000 | 660.0000000000000000 |
| | Runtime | 3012.883 | 241.541 | 2728.060 | 1573.484 | 10891.124 | 1769.459 | 2972.618 | 279.0646913 | 530.172 |
| | NFE | | | | | | | | | |
| F72 | Mean | 1094.8305116977000000 | -7.159E + 49 | 915.4958100611630000 | 1133.7536009808600000 | 1075.5292326436900000 | 1094.6823697304900000 | 984.5106541514410000 | 694.3706620000000000 | 1088.6626563226300000 |
| | Std. Dev. | 121.3539576317800000 | 4.387E + 50 | 242.1993331983530000 | 42.1171260000361000 | 166.9355145236330000 | 87.9884000140656000 | 199.1563947691970000 | 20.9754439100000000 | 136.0666138798300000 |
| | Best | 660.0000000000000000 | -133.9585340104890000 | 660.0006867770510000 | 1088.9543269392600000 | 660.0000000000020000 | 660.0000000000000000 | 660.0000000000000000 | 644.2542524502140000 | 660.0000000000000000 |
| | Runtime | 6363.267 | 290.334 | 2326.112 | 1730.723 | 9601.880 | 3854.148 | 10458.467 | 273.922 | 997.068 |
| | | | | | | | | | | |
| F73 | Mean | 1304.3661550124000000 | 1159.9280867973000000 | 830.2290165794410000 | 1167.9040488743800000 | 1070.4327462836400000 | 1105.2511774948600000 | 976.2273885425320000 | 559.6581705000000000 | 919.4683268438060000 |
| | Std. Dev. | 262.1065863453340000 | 742.1215416320490000 | 60.2286903507069000 | 236.7325108248320000 | 203.0676662707430000 | 190.6172874229610000 | 160.1543461970300000 | 16.1193896300000000 | 136.2630721125810000 |
| | Best | 919.4683107913200000 | -460.7504508023100000 | 785.1725102979490000 | 785.1725102979490000 | 785.1725102979480000 | 919.4683107913240000 | 785.1725102979480000 | 546.1130231359180000 | 919.4683114379170000 |
| | Runtime | 2165.640 | 238.261 | 2045.582 | 1580.067 | 7459.005 | 1901.540 | 4209.110 | 287.271 | 118.771 |
| | | | | | | | | | | |
| F74 | Mean | 500.0000000000000000 | 653.3355378428050000 | 460.0000000000020000 | 510.0000000000000000 | 493.3333333333340000 | 490.0000000000000000 | 460.0000000000000000 | 463.2262530000000000 | 460.0000000000000000 |
| | Std. Dev. | 103.7237710925280000 | 302.5312999719650000 | 0.0000000000016493 | 113.7147065368360000 | 137.2973951415090000 | 91.5385729888094000 | 0.0000000000000000 | 4.9321910760000000 | 0.0000000000000000 |
| | Best | 460.0000000000000000 | 460.0000000000000000 | 460.0000000000000000 | 460.0000000000000000 | 460.0000000000000000 | 460.0000000000000000 | 460.0000000000000000 | 458.5444354721460000 | 460.0000000000000000 |
| | Runtime | 1811.980 | 165.962 | 1698.121 | 1366.710 | 3016.959 | 1410.399 | 1795.637 | 257.960 | 1278.572 |
| | | | | | | | | | | |
| F75 | Mean | 1107.9038127876700000 | 1401.6553278264300000 | 930.4565414149210000 | 1072.9924659809200000 | 1258.5157766524700000 | 1074.3695435628600000 | 1063.7363787709700000 | 471.2797518000000000 | 1084.7073068225200000 |
| | Std. Dev. | 127.9566489362040000 | 253.2428066220210000 | 87.9959072391079000 | 2.2606058314671500 | 241.4024507676890000 | 2.8314182838917800 | 55.8479313799755000 | 2.2346287190000000 | 4.9504851126832800 |
| | Best | 1069.5511765775700000 | 1072.4973401423200000 | 862.4476004191700000 | 1068.5560012648600000 | 871.8607884176050000 | 1069.8723890709000000 | 856.8214538442850000 | 469.3372925643150000 | 1078.2231646698500000 |
| | Runtime | 4060.091 | 214.580 | 2113.339 | 2951.018 | 5262.210 | 3410.902 | 4280.901 | 263.829 | 711.530 |
| | | | | | | | | | | |

Table 5: Statistical results for Test 1 Problems using two-sided Wilcoxon Signed-Rank Test ($\alpha = 0.05$)

| **Problem** | **PSO2011 vs Multi CI** | | | | **CMAES vs Multi CI** | | | | **ABC vs Multi CI** | | | | **JDE vs Multi CI** | | | |
|---|---|---|---|---|---|---|---|---|---|---|---|---|---|---|---|---|
| | p-value | T+ | T- | winner | p-value | T+ | T- | winner | p-value | T+ | T- | winner | p-value | T+ | T- | winner |
| F1 | 4.32E-08 | 0 | 465 | + | 4.32E-08 | 0 | 465 | + | 4.32E-08 | 0 | 465 | + | 4.32E-08 | 0 | 465 | + |
| F2 | 4.32E-08 | 0 | 465 | + | 4.32E-08 | 0 | 465 | + | 4.32E-08 | 0 | 465 | + | 4.32E-08 | 0 | 465 | + |
| F3 | 4.32E-08 | 0 | 465 | + | 4.32E-08 | 0 | 465 | + | 4.32E-08 | 0 | 465 | + | 1.34E-06 | 465 | 0 | - |
| F4 | 4.32E-08 | 0 | 465 | + | 6.80E-08 | 0 | 465 | + | 3.35E-07 | 0 | 465 | + | 4.32E-08 | 0 | 465 | + |
| F5 | 1.73E-06 | 0 | 465 | + | 1.73E-06 | 0 | 465 | + | 1.73E-06 | 465 | 0 | - | 1.73E-06 | 0 | 465 | + |
| F6 | 1.73E-06 | 465 | 0 | - | 1.73E-06 | 0 | 465 | + | 1.73E-06 | 465 | 0 | - | 1.73E-06 | 465 | 0 | - |
| F7 | 1.73E-06 | 0 | 465 | + | 1.73E-06 | 0 | 465 | + | 1.73E-06 | 465 | 0 | - | 1.73E-06 | 465 | 0 | - |
| F8 | 1.73E-06 | 0 | 465 | + | 1.73E-06 | 0 | 465 | + | 1.73E-06 | 0 | 465 | + | 1.73E-06 | 0 | 465 | + |
| F9 | 4.32E-08 | 0 | 465 | + | 4.32E-08 | 0 | 465 | + | 4.32E-08 | 0 | 465 | + | 4.32E-08 | 0 | 465 | + |
| F10 | 4.32E-08 | 0 | 465 | + | 4.32E-08 | 0 | 465 | + | 4.32E-08 | 0 | 465 | + | 4.32E-08 | 0 | 465 | + |
| F11 | 4.32E-08 | 0 | 465 | + | 4.32E-08 | 0 | 465 | + | 4.32E-08 | 0 | 465 | + | 4.32E-08 | 0 | 465 | + |
| F12 | 4.32E-08 | 0 | 465 | + | 4.32E-08 | 0 | 465 | + | 4.32E-08 | 0 | 465 | + | 4.32E-08 | 0 | 465 | + |
| F13 | 1.66E-06 | 0 | 465 | + | 1.66E-06 | 465 | 0 | - | 1.66E-06 | 465 | 0 | - | 5.66E-02 | 140 | 325 | + |
| F14 | 1.67E-06 | 465 | 0 | - | 1.67E-06 | 465 | 0 | - | 1.55E-06 | 465 | 0 | - | 1.55E-06 | 465 | 0 | - |
| F15 | 1.01E-07 | 0 | 465 | + | 1.01E-07 | 0 | 465 | + | 1.01E-07 | 465 | 0 | - | 1.01E-07 | 465 | 0 | - |
| F16 | 6.87E-07 | 0 | 465 | + | 6.87E-07 | 0 | 465 | + | 6.87E-07 | 0 | 465 | + | 6.87E-07 | 0 | 465 | + |
| F17 | 1.10E-06 | 0 | 465 | + | 1.10E-06 | 0 | 465 | + | 1.10E-06 | 0 | 465 | + | 1.10E-06 | 0 | 465 | + |
| F18 | 1.01E-07 | 465 | 0 | - | 1.01E-07 | 465 | 0 | - | 1.01E-07 | 465 | 0 | - | 1.01E-07 | 465 | 0 | - |
| F19 | 1.20E-06 | 0 | 465 | + | 1.20E-06 | 0 | 465 | + | 1.20E-06 | 465 | 0 | - | 1.20E-06 | 465 | 0 | - |
| F20 | 4.32E-08 | 465 | 0 | - | 4.32E-08 | 465 | 0 | - | 4.32E-08 | 465 | 0 | - | 4.32E-08 | 465 | 0 | - |
| F21 | 1.73E-06 | 0 | 465 | + | 1.73E-06 | 0 | 465 | + | 1.73E-06 | 0 | 465 | + | 1.73E-06 | 0 | 465 | + |
| F22 | 1.66E-06 | 0 | 465 | + | 1.66E-06 | 0 | 465 | + | 1.66E-06 | 0 | 465 | + | 1.66E-06 | 0 | 465 | + |
| F23 | 1.73E-06 | 0 | 465 | + | 1.73E-06 | 0 | 465 | + | 1.73E-06 | 0 | 465 | + | 1.73E-06 | 0 | 465 | + |
| F24 | 4.32E-08 | 465 | 0 | - | 4.32E-08 | 465 | 0 | - | 4.32E-08 | 465 | 0 | - | 4.32E-08 | 465 | 0 | - |
| F25 | 4.32E-08 | 0 | 465 | + | 4.32E-08 | 0 | 465 | + | 4.32E-08 | 0 | 465 | + | 4.32E-08 | 0 | 465 | + |
| F26 | 4.32E-08 | 465 | 0 | - | 4.32E-08 | 465 | 0 | - | 4.32E-08 | 465 | 0 | - | 4.32E-08 | 465 | 0 | - |
| F27 | 4.32E-08 | 465 | 0 | - | 4.32E-08 | 465 | 0 | - | 4.32E-08 | 465 | 0 | - | 4.32E-08 | 465 | 0 | - |
| F28 | 4.32E-08 | 465 | 0 | - | 4.32E-08 | 465 | 0 | - | 4.32E-08 | 465 | 0 | - | 4.32E-08 | 465 | 0 | - |
| F29 | 5.99E-07 | 0 | 465 | + | 5.99E-07 | 0 | 465 | + | 5.99E-07 | 0 | 465 | + | 5.99E-07 | 0 | 465 | + |
| F30 | 1.70E-06 | 465 | 0 | - | 1.70E-06 | 465 | 0 | - | 1.70E-06 | 465 | 0 | - | 1.70E-06 | 465 | 0 | - |
| F31 | 1.08E-06 | 0 | 465 | + | 1.08E-06 | 465 | 0 | - | 1.08E-06 | 0 | 465 | + | 1.08E-06 | 0 | 465 | + |
| F32 | 4.32E-08 | 0 | 465 | + | 4.32E-08 | 0 | 465 | + | 4.32E-08 | 0 | 465 | + | 4.32E-08 | 0 | 465 | + |
| F33 | 4.32E-08 | 0 | 465 | + | 4.32E-08 | 0 | 465 | + | 4.32E-08 | 0 | 465 | + | 4.32E-08 | 0 | 465 | + |
| F34 | 4.32E-08 | 0 | 465 | + | 4.32E-08 | 0 | 465 | + | 4.32E-08 | 0 | 465 | + | 4.32E-08 | 0 | 465 | + |
| F35 | 4.32E-08 | 0 | 465 | + | 4.32E-08 | 0 | 465 | + | 4.32E-08 | 0 | 465 | + | 4.32E-08 | 0 | 465 | + |
| F36 | 4.32E-08 | 465 | 0 | - | 4.32E-08 | 465 | 0 | - | 4.32E-08 | 465 | 0 | - | 4.32E-08 | 465 | 0 | - |
| F37 | 4.32E-08 | 0 | 465 | + | 4.32E-08 | 0 | 465 | + | 4.32E-08 | 0 | 465 | + | 4.32E-08 | 0 | 465 | + |
| F38 | 3.32E-07 | 465 | 0 | - | 3.32E-07 | 465 | 0 | - | 4.32E-08 | 0 | 465 | + | 3.32E-07 | 465 | 0 | - |
| F39 | 4.32E-08 | 465 | 0 | - | 4.32E-08 | 465 | 0 | - | 4.32E-08 | 465 | 0 | - | 4.32E-08 | 465 | 0 | - |
| F40 | 4.32E-08 | 465 | 0 | - | 4.32E-08 | 465 | 0 | - | 4.32E-08 | 465 | 0 | - | 4.32E-08 | 465 | 0 | - |
| F41 | 4.32E-08 | 465 | 0 | - | 4.32E-08 | 465 | 0 | - | 4.32E-08 | 465 | 0 | - | 4.32E-08 | 465 | 0 | - |
| F42 | 1.96E-07 | 465 | 0 | - | 1.96E-07 | 465 | 0 | - | 1.96E-07 | 465 | 0 | - | 1.96E-07 | 465 | 0 | - |
| F43 | 4.32E-08 | 465 | 0 | - | 4.32E-08 | 465 | 0 | - | 4.32E-08 | 465 | 0 | - | 4.32E-08 | 465 | 0 | - |
| F44 | 4.32E-08 | 465 | 0 | - | 4.32E-08 | 465 | 0 | - | 4.32E-08 | 465 | 0 | - | 4.32E-08 | 465 | 0 | - |
| F45 | 4.32E-08 | 0 | 465 | + | 4.32E-08 | 0 | 465 | + | 4.32E-08 | 0 | 465 | + | 4.32E-08 | 0 | 465 | + |
| F46 | 4.32E-08 | 465 | 0 | - | 4.32E-08 | 465 | 0 | - | 4.32E-08 | 465 | 0 | - | 4.32E-08 | 465 | 0 | - |
| F47 | 4.32E-08 | 0 | 465 | + | 4.32E-08 | 0 | 465 | + | 4.32E-08 | 0 | 465 | + | 4.32E-08 | 0 | 465 | + |
| F48 | 4.32E-08 | 465 | 0 | - | 4.32E-08 | 465 | 0 | - | 4.32E-08 | 465 | 0 | - | 4.32E-08 | 465 | 0 | - |
| F49 | 4.32E-08 | 465 | 0 | - | 4.32E-08 | 465 | 0 | - | 4.32E-08 | 465 | 0 | - | 4.32E-08 | 465 | 0 | - |
| F50 | 4.32E-08 | 0 | 465 | + | 4.32E-08 | 0 | 465 | + | 4.32E-08 | 0 | 465 | + | 4.32E-08 | 0 | 465 | + |
| +/=/- | 30/0/20 | | | | 29/0/21 | | | | 26/0/24 | | | | 26/0/24 | | | |

| Problem | CLPSO vs Multi CI | | | | SADE vs Multi CI | | | | BSA vs Multi CI | | | | IA vs Multi CI | | | |
|---|---|---|---|---|---|---|---|---|---|---|---|---|---|---|---|---|
| | p-value | T+ | T- | winner | p-value | T+ | T- | winner | p-value | T+ | T- | winner | p-value | T+ | T- | winner |
| F1 | 4.32E-08 | 0 | 465 | + | 4.32E-08 | 0 | 465 | + | 1 | 0 | 0 | = | 1.67E-06 | 0 | 465 | + |
| F2 | 4.32E-08 | 0 | 465 | + | 4.32E-08 | 0 | 465 | + | 1 | 0 | 0 | = | 1.01E-06 | 0 | 465 | + |
| F3 | 1.34E-06 | 465 | 0 | - | 4.32E-08 | 0 | 465 | + | 1 | 0 | 0 | = | 4.32E-08 | 0 | 465 | + |
| F4 | 3.35E-07 | 465 | 0 | - | 4.32E-08 | 0 | 465 | + | 1 | 0 | 0 | = | 1.66E-06 | 0 | 465 | + |
| F5 | 1.73E-06 | 0 | 465 | + | 1.73E-06 | 0 | 465 | + | 4.32E-08 | 0 | 465 | + | 1 | 0 | 0 | = |
| F6 | 1.73E-06 | 465 | 0 | - | 1.73E-06 | 465 | 0 | - | 1 | 0 | 0 | = | 4.32E-08 | 0 | 465 | + |
| F7 | 1.73E-06 | 465 | 0 | - | 1.73E-06 | 465 | 0 | - | 1 | 0 | 0 | = | 1 | 0 | 0 | = |
| F8 | 1.73E-06 | 0 | 465 | + | 1.73E-06 | 0 | 465 | + | 1 | 0 | 0 | = | 1 | 0 | 0 | = |
| F9 | 4.32E-08 | 0 | 465 | + | 1 | 0 | 0 | = | 1 | 0 | 0 | = | 1 | 0 | 0 | = |
| F10 | 4.32E-08 | 0 | 465 | + | 4.32E-08 | 465 | 0 | - | 1 | 0 | 0 | = | 4.32E-08 | 0 | 465 | + |
| F11 | 4.32E-08 | 0 | 465 | + | 4.32E-08 | 0 | 465 | + | 1 | 0 | 0 | = | 1.72E+00 | 0 | 465 | + |
| F12 | 4.32E-08 | 0 | 465 | + | 4.32E-08 | 465 | 0 | - | 1 | 0 | 0 | = | 5.99E-07 | 0 | 465 | + |
| F13 | 0.0566 | 140 | 325 | + | 1.66E-06 | 0 | 465 | + | 0.0027 | 378 | 87 | - | 3.19E-06 | 459 | 6 | - |
| F14 | 1.55E-06 | 465 | 0 | - | 1.55E-06 | 465 | 0 | - | 4.88E-04 | 78 | 0 | - | 3.19E-06 | 459 | 6 | - |
| F15 | 1.01E-07 | 465 | 0 | - | 1.01E-07 | 465 | 0 | - | 1 | 0 | 0 | = | 1 | 0 | 0 | = |
| F16 | 6.87E-07 | 0 | 465 | + | 6.87E-07 | 0 | 465 | + | 1 | 0 | 0 | = | 1 | 0 | 0 | = |
| F17 | 1.10E-06 | 0 | 465 | + | 1.10E-06 | 465 | 0 | - | 1 | 0 | 0 | = | 1 | 0 | 0 | = |
| F18 | 1.01E-07 | 465 | 0 | - | 1.01E-07 | 465 | 0 | - | 4.32E-08 | 0 | 465 | + | 1 | 0 | 0 | = |
| F19 | 1.20E-06 | 465 | 0 | - | 1.20E-06 | 465 | 0 | - | 1.69E-06 | 465 | 0 | - | 1.73E-06 | 0 | 465 | + |
| F20 | 4.32E-08 | 465 | 0 | - | 4.32E-08 | 465 | 0 | - | 3.11E-06 | 459 | 6 | - | 1.69E-06 | 0 | 465 | + |
| F21 | 1.73E-06 | 0 | 465 | + | 1.73E-06 | 0 | 465 | + | 1.69E-06 | 465 | 0 | - | 1.70E-06 | 0 | 465 | + |
| F22 | 1.66E-06 | 0 | 465 | + | 1.66E-06 | 0 | 465 | + | 4.32E-08 | 0 | 465 | + | 1.69E-06 | 465 | 0 | - |
| F23 | 1.73E-06 | 0 | 465 | + | 1.73E-06 | 0 | 465 | + | 1.69E-06 | 0 | 465 | + | 1.69E-06 | 465 | 0 | - |
| F24 | 4.32E-08 | 465 | 0 | - | 4.32E-08 | 465 | 0 | - | 1.69E-06 | 0 | 465 | + | 1.69E-06 | 465 | 0 | - |
| F25 | 4.32E-08 | 0 | 465 | + | 1 | 0 | 0 | = | 1 | 0 | 0 | = | 1 | 0 | 0 | = |
| F26 | 4.32E-08 | 465 | 0 | - | 4.32E-08 | 465 | 0 | - | 1.69E-06 | 0 | 465 | + | 1.69E-06 | 0 | 465 | + |
| F27 | 4.32E-08 | 465 | 0 | - | 4.32E-08 | 465 | 0 | - | 1.66E-06 | 465 | 0 | - | 1.69E-06 | 0 | 465 | + |
| F28 | 4.32E-08 | 465 | 0 | - | 4.32E-08 | 465 | 0 | - | 1.66E-06 | 465 | 0 | - | 1.69E-06 | 0 | 465 | + |
| F29 | 5.99E-07 | 0 | 465 | + | 5.99E-07 | 0 | 465 | + | 1.40E-06 | 465 | 0 | - | 1.43E-06 | 0 | 465 | + |
| F30 | 1.70E-06 | 465 | 0 | - | 1.70E-06 | 465 | 0 | - | 4.32E-08 | 0 | 465 | + | 1 | 0 | 0 | = |
| F31 | 1.08E-06 | 0 | 465 | + | 1.08E-06 | 0 | 465 | + | 1.66E-06 | 465 | 0 | - | 1.69E-06 | 0 | 465 | + |
| F32 | 4.32E-08 | 0 | 465 | + | 4.32E-08 | 0 | 465 | + | 1.66E-06 | 0 | 465 | + | 8.94E-04 | 71 | 394 | + |
| F33 | 4.32E-08 | 0 | 465 | + | 4.32E-08 | 0 | 465 | + | 1 | 0 | 0 | = | 1 | 0 | 0 | = |
| F34 | 4.32E-08 | 0 | 465 | + | 4.32E-08 | 0 | 465 | + | 1 | 0 | 0 | = | 1.69E-06 | 465 | 0 | - |
| F35 | 4.32E-08 | 0 | 465 | + | 4.32E-08 | 0 | 465 | + | 1 | 0 | 0 | = | 1 | 0 | 0 | = |
| F36 | 4.32E-08 | 465 | 0 | - | 4.32E-08 | 465 | 0 | - | 1 | 0 | 0 | = | 1.69E-06 | 465 | 0 | - |
| F37 | 4.32E-08 | 0 | 465 | + | 4.32E-08 | 0 | 465 | + | 1 | 0 | 0 | = | 1 | 0 | 0 | = |
| F38 | 3.32E-07 | 0 | 465 | + | 3.32E-07 | 465 | 0 | - | 1 | 0 | 0 | = | 1 | 0 | 0 | = |
| F39 | 4.32E-08 | 465 | 0 | - | 4.32E-08 | 465 | 0 | - | 1.69E-06 | 465 | 0 | - | 1.73E-06 | 0 | 465 | + |
| F40 | 4.32E-08 | 465 | 0 | - | 4.32E-08 | 465 | 0 | - | 1 | 0 | 0 | = | 1.73E-06 | 0 | 465 | + |
| F41 | 4.32E-08 | 465 | 0 | - | 4.32E-08 | 465 | 0 | - | 1 | 0 | 0 | = | 1.73E-06 | 0 | 465 | + |
| F42 | 1.96E-07 | 465 | 0 | - | 1.96E-08 | 465 | 0 | - | 1 | 0 | 0 | = | 1.69E-06 | 0 | 465 | + |
| F43 | 4.32E-08 | 465 | 0 | - | 4.32E-08 | 465 | 0 | - | 1 | 0 | 0 | = | 2.10E-03 | 382 | 83 | - |
| F44 | 4.32E-08 | 465 | 0 | - | 4.32E-08 | 465 | 0 | - | 1 | 0 | 0 | = | 1 | 0 | 0 | = |
| F45 | 1 | 0 | 0 | = | 1 | 0 | 0 | = | 1 | 0 | 0 | = | 1.69E-06 | 0 | 465 | + |
| F46 | 4.32E-08 | 465 | 0 | - | 4.32E-08 | 465 | 0 | - | 4.32E-08 | 465 | 0 | - | 1.69E-06 | 0 | 465 | - |
| F47 | 4.32E-08 | 0 | 465 | + | 4.32E-08 | 0 | 465 | + | 1 | 0 | 0 | = | 1 | 0 | 0 | = |
| F48 | 4.32E-08 | 465 | 0 | - | 4.32E-08 | 0 | 465 | + | 1 | 0 | 0 | = | 1.69E-06 | 0 | 465 | + |
| F49 | 4.32E-08 | 465 | 0 | - | 4.32E-08 | 465 | 0 | - | 1 | 0 | 0 | = | 1.69E-06 | 0 | 465 | + |
| F50 | 4.32E-08 | 0 | 465 | + | 4.32E-08 | 0 | 465 | + | 4.32E-08 | 0 | 465 | - | 1 | 0 | 0 | = |
| +/=/- | 25/1/24 | | | | 22/3/25 | | | | 8/30/12 | | | | 24/17/9 | | | |

Table 6: Statistical results for Test 2 Problems using two-sided Wilcoxon Signed-Rank Test ($\alpha = 0.05$)

| **Problem** | **PSO vs Multi CI** | | | | **CMAES vs Multi CI** | | | | **ABC vs Multi CI** | | | | **JDE vs Multi CI** | | | |
|---|---|---|---|---|---|---|---|---|---|---|---|---|---|---|---|---|
| | p-value | T+ | T- | winner | p-value | T+ | T- | p-value | T+ | T- | winner | winner | p-value | T+ | T- | winner |
| F51 | 6.91E-07 | 465 | 0 | - | 6.91E-07 | 465 | 0 | - | 6.91E-07 | 465 | 0 | - | 6.91E-07 | 465 | 0 | - |
| F52 | 1.18E-06 | 0 | 465 | + | 1.18E-06 | 465 | 0 | - | 1.18E-06 | 465 | 0 | - | 1.18E-06 | 0 | 465 | + |
| F53 | 4.32E-08 | 0 | 465 | + | 4.32E-08 | 465 | 0 | - | 4.32E-08 | 0 | 465 | + | 4.32E-08 | 0 | 465 | + |
| F54 | 1.73E-06 | 0 | 465 | + | 1.73E-06 | 0 | 465 | + | 1.73E-06 | 0 | 465 | + | 1.73E-06 | 465 | 0 | - |
| F55 | 4.32E-08 | 0 | 465 | + | 4.32E-08 | 0 | 465 | + | 4.32E-08 | 0 | 465 | + | 4.32E-08 | 0 | 465 | + |
| F56 | 4.32E-08 | 0 | 465 | + | 4.32E-08 | 465 | 0 | - | 4.32E-08 | 0 | 465 | + | 4.32E-08 | 465 | 0 | - |
| F57 | 6.80E-08 | 0 | 465 | + | 6.80E-08 | 0 | 465 | + | 6.80E-08 | 0 | 465 | + | 6.80E-08 | 0 | 465 | + |
| F58 | 4.32E-08 | 465 | 0 | - | 4.32E-08 | 465 | 0 | - | 4.32E-08 | 465 | 0 | - | 4.32E-08 | 0 | 465 | + |
| F59 | 4.32E-08 | 0 | 465 | + | 4.32E-08 | 0 | 465 | + | 4.32E-08 | 465 | 0 | - | 4.32E-08 | 465 | 0 | - |
| F60 | 3.96E-05 | 36 | 429 | + | 1.16E-06 | 0 | 465 | + | 1.16E-06 | 0 | 465 | + | 1.16E-06 | 0 | 465 | + |
| F61 | 4.32E-08 | 0 | 465 | + | 4.32E-08 | 465 | 0 | - | 4.32E-08 | 0 | 465 | + | 4.32E-08 | 0 | 465 | + |
| F62 | 1.44E-07 | 0 | 465 | + | 1.44E-07 | 0 | 465 | + | 2.99E-07 | 6 | 459 | + | 1.44E-07 | 0 | 465 | + |
| F63 | 4.32E-08 | 465 | 0 | - | 4.32E-08 | 465 | 0 | - | 4.32E-08 | 465 | 0 | - | 4.32E-08 | 465 | 0 | - |
| F64 | 1.51E-06 | 465 | 0 | - | 1.51E-06 | 0 | 465 | + | 1.51E-06 | 465 | 0 | - | 1.51E-06 | 465 | 0 | - |
| F65 | 4.32E-08 | 465 | 0 | - | 4.32E-08 | 465 | 0 | - | 4.32E-08 | 465 | 0 | - | 4.32E-08 | 465 | 0 | - |
| F66 | 7.86E-07 | 465 | 0 | - | 7.86E-07 | 0 | 465 | + | 7.86E-07 | 465 | 0 | - | 7.86E-07 | 465 | 0 | - |
| F67 | 1.73E-06 | 0 | 465 | + | 1.73E-06 | 0 | 465 | + | 1.73E-06 | 0 | 465 | + | 1.73E-06 | 0 | 465 | + |
| F68 | 6.98E-07 | 0 | 465 | + | 6.98E-07 | 0 | 465 | + | 6.98E-07 | 0 | 465 | + | 6.98E-07 | 0 | 465 | + |
| F69 | 1.19E-06 | 0 | 465 | + | 1.19E-06 | 0 | 465 | + | 1.19E-06 | 0 | 465 | + | 1.19E-06 | 0 | 465 | + |
| F70 | 1.20E-06 | 0 | 465 | + | 1.20E-06 | 0 | 465 | + | 1.20E-06 | 0 | 465 | + | 1.20E-06 | 0 | 465 | + |
| F71 | 9.27E-07 | 0 | 465 | + | 9.27E-07 | 0 | 465 | + | 9.27E-07 | 0 | 465 | + | 9.27E-07 | 0 | 465 | + |
| F72 | 1.97E-07 | 465 | 0 | - | 4.32E-08 | 465 | 0 | - | 1.97E-07 | 465 | 0 | - | 4.32E-08 | 465 | 0 | - |
| F73 | 8.89E-07 | 0 | 465 | + | 8.89E-07 | 0 | 465 | + | 8.89E-07 | 0 | 465 | + | 8.89E-07 | 0 | 465 | + |
| F74 | 1.73E-06 | 0 | 465 | + | 1.73E-06 | 0 | 465 | + | 1.73E-06 | 0 | 465 | + | 1.73E-06 | 0 | 465 | + |
| F75 | 1.73E-06 | 0 | 465 | + | 1.73E-06 | 0 | 465 | + | 1.73E-06 | 0 | 465 | + | 1.73E-06 | 0 | 465 | + |
| +/=/- | 18/0/7 | | | | 16/0/9 | | | | 16/0/9 | | | | 16/0/9 | | | |

| Problem | CLPSO vs Multi CI | | | | SADE vs Multi CI | | | | BSA vs Multi CI | | | | IA vs Multi CI | | | |
|---|---|---|---|---|---|---|---|---|---|---|---|---|---|---|---|---|
| | p-value | T+ | T- | winner | p-value | T+ | T- | winner | p-value | T+ | T- | winner | p-value | T+ | T- | winner |
| F51 | 1.18E-06 | 0 | 465 | + | 6.91E-07 | 465 | 0 | - | 1 | 0 | 0 | = | 6.91E+00 | 0 | 465 | + |
| F52 | 4.32E-08 | 0 | 465 | + | 1.18E-06 | 465 | 0 | - | 1 | 0 | 0 | = | 1.18E-06 | 0 | 465 | + |
| F53 | 1.73E-06 | 0 | 465 | + | 4.32E-08 | 0 | 465 | + | 1.69E-06 | 465 | 0 | - | 1.58E-06 | 465 | 0 | - |
| F54 | 1.73E-06 | 0 | 465 | + | 1.73E-06 | 465 | 0 | - | 1 | 0 | 0 | = | 1.73E-06 | 0 | 465 | + |
| F55 | 4.32E-08 | 0 | 465 | + | 4.32E-08 | 0 | 465 | + | 4.32E-08 | 0 | 465 | + | 4.32E-08 | 465 | 0 | - |
| F56 | 4.32E-08 | 0 | 465 | + | 4.32E-08 | 465 | 0 | - | 1.58E-06 | 465 | 0 | - | 1.58E-06 | 465 | 0 | - |
| F57 | 6.80E-08 | 465 | 0 | - | 6.80E-08 | 465 | 0 | - | 1.58E-06 | 465 | 0 | - | 1 | 0 | 0 | = |
| F58 | 4.32E-08 | 465 | 0 | - | 4.32E-08 | 465 | 0 | - | 3.11E-06 | 6 | 459 | + | 1.69E-06 | 0 | 465 | + |
| F59 | 4.32E-08 | 465 | 0 | - | 4.32E-08 | 465 | 0 | - | 1.69E-06 | 465 | 0 | - | 3.13E-06 | 459 | 6 | - |
| F60 | 1.16E-06 | 0 | 465 | + | 1.16E-06 | 0 | 465 | + | 1.28E-05 | 21 | 444 | + | 3.13E-06 | 459 | 6 | - |
| F61 | 4.32E-08 | 0 | 465 | + | 4.32E-08 | 0 | 465 | + | 1.66E-06 | 0 | 465 | + | 1.69E-06 | 0 | 465 | + |
| F62 | 1.44E-07 | 465 | 0 | - | 9.96E-04 | 87 | 378 | + | 1.69E-06 | 0 | 465 | + | 1.69E-06 | 0 | 465 | + |
| F63 | 4.32E-08 | 465 | 0 | - | 4.32E-08 | 465 | 0 | - | 1.66E-06 | 465 | 0 | - | 1.69E-06 | 0 | 465 | + |
| F64 | 1.51E-06 | 465 | 0 | - | 1.51E-06 | 465 | 0 | - | 8.66E-05 | 423 | 42 | - | 1.73E-06 | 0 | 465 | + |
| F65 | 4.32E-08 | 465 | 0 | - | 4.32E-08 | 465 | 0 | - | 8.66E-05 | 465 | 0 | - | 4.32E-08 | 465 | 0 | - |
| F66 | 7.86E-07 | 465 | 0 | - | 7.86E-07 | 465 | 0 | - | 2.96E-06 | 6 | 459 | + | 1.69E-06 | 0 | 465 | + |
| F67 | 1.73E-06 | 0 | 465 | + | 1.73E-06 | 0 | 465 | + | 3.72E-04 | 60 | 405 | + | 1.73E-06 | 465 | 0 | - |
| F68 | 6.98E-07 | 0 | 465 | + | 6.98E-07 | 0 | 465 | + | 0.0266 | 126 | 339 | + | 1.73E-06 | 465 | 0 | - |
| F69 | 1.19E-06 | 0 | 465 | + | 1.19E-06 | 0 | 465 | + | 6.07E-04 | 399 | 66 | - | 1.73E-06 | 465 | 0 | - |
| F70 | 1.20E-06 | 0 | 465 | + | 1.20E-06 | 0 | 465 | + | 0.1646 | 300 | 165 | - | 1.73E-06 | 465 | 0 | - |
| F71 | 9.27E-07 | 0 | 465 | + | 9.27E-07 | 0 | 465 | + | 0.001 | 78 | 387 | + | 1.73E-06 | 465 | 0 | - |
| F72 | 4.32E-08 | 465 | 0 | - | 1.97E-07 | 465 | 0 | - | 1.97E-07 | 465 | 0 | - | 4.32E-08 | 465 | 0 | - |
| F73 | 8.89E-07 | 0 | 465 | + | 8.89E-07 | 0 | 465 | + | 1.13E-04 | 420 | 45 | - | 1.73E-06 | 465 | 0 | - |
| F74 | 1.73E-06 | 0 | 465 | + | 1.73E-06 | 0 | 465 | + | 1 | 0 | 0 | = | 1.69E-06 | 0 | 465 | + |
| F75 | 1.73E-06 | 0 | 465 | + | 1.73E-06 | 0 | 465 | + | 1.69E-06 | 465 | 0 | - | 1.73E-06 | 465 | 0 | - |
| +/=/- | 16/0/9 | | | | 13/0/12 | | | | 9/4/12 | | | | 10/1/14 | | | |

Table 7: Multi-problem based statistical pairwise comparison of PSO, CMAES, ABC, JDE, CLPSO, SADE, BSA, IA and Multi-CI

| Other Algorithm vs IA | p-Value | T+ | T- | Winner |
|---|---|---|---|---|
| PSO vs Multi CI | 0.0035 | 368 | 1010 | Multi CI |
| CMAES vs Multi CI | 3.3367e-07 | 235 | 1656 | Multi CI |
| ABC vs Multi CI | 0.1355 | 615 | 981 | Multi CI |
| JDE vs Multi CI | 4.6305e-04 | 320 | 1111 | Multi CI |
| CLPSO vs Multi CI | 1.507e-04 | 366 | 1345 | Multi CI |
| SADE vs Multi CI | 0.2031 | 424 | 657 | Multi CI |
| BSA vs Multi CI | 0.9144 | 402 | 418 | Multi CI |
| IA vs Multi CI | 0.7003 | 834 | 936 | Multi CI |

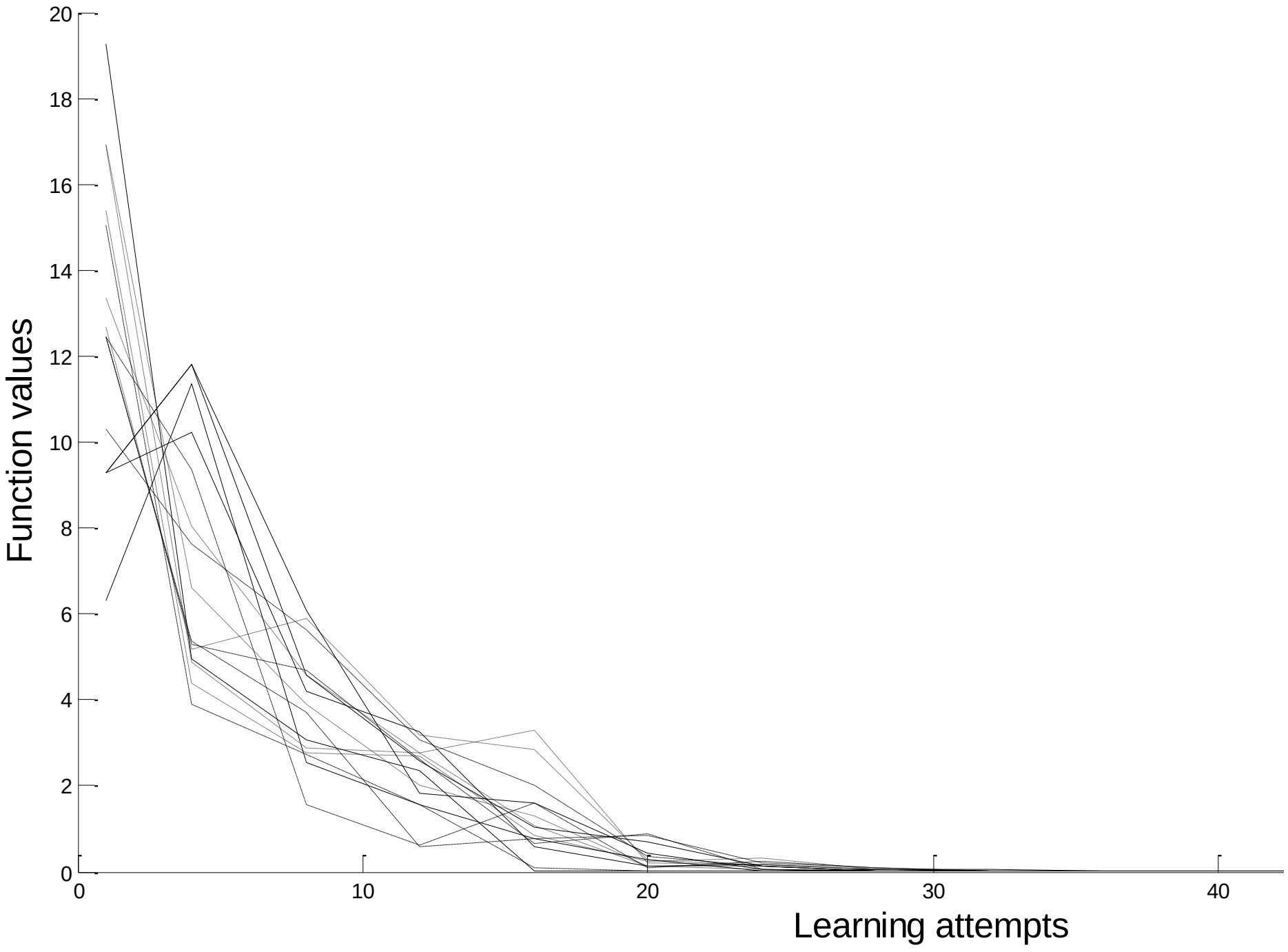

(a) Convergence of all candidates

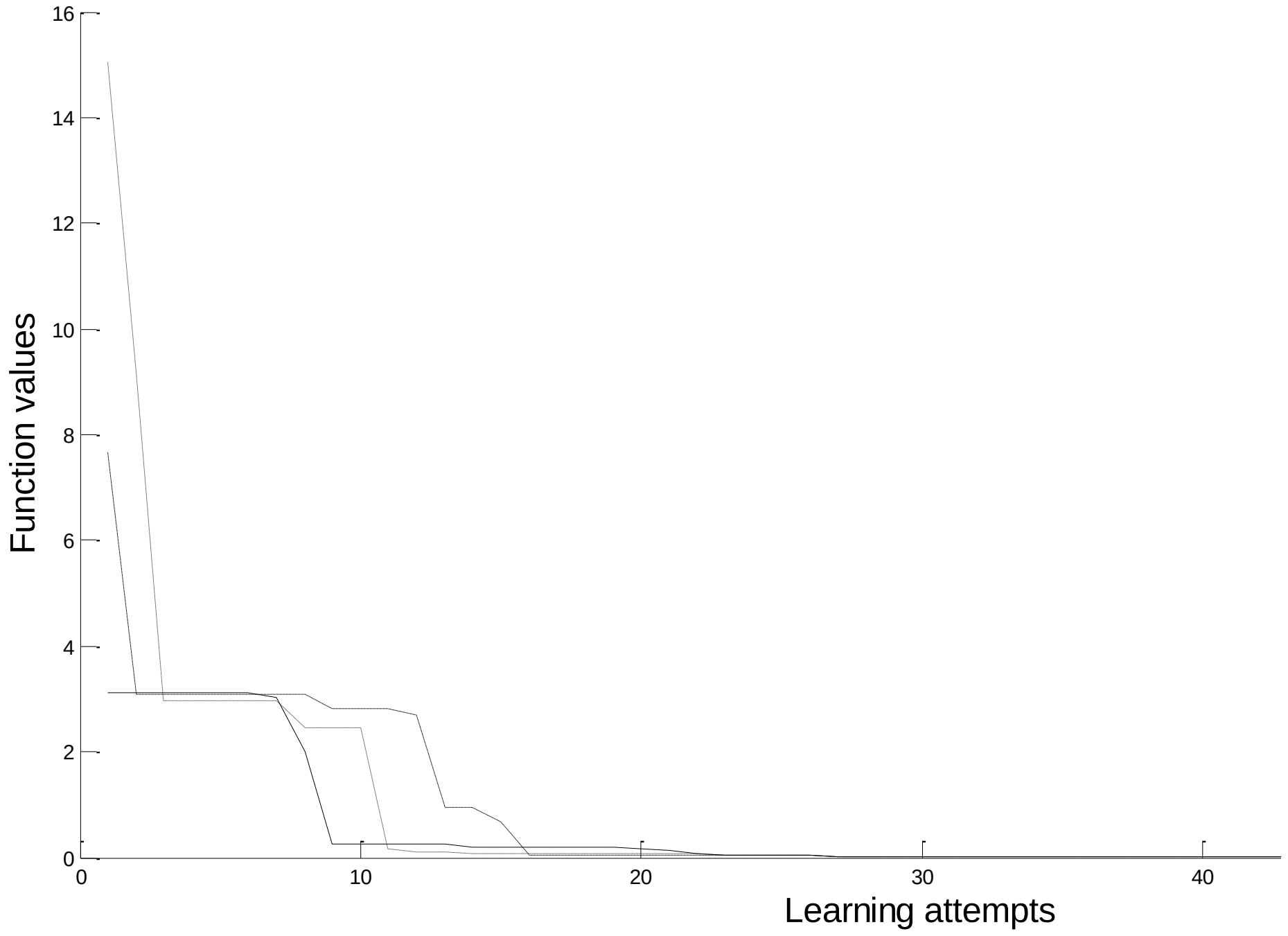

(b) Convergence of best candidates

——— Cohort 1 — — Cohort 2 ------- Cohort 3

Figure 2 Convergence for Ackley Function (F5)

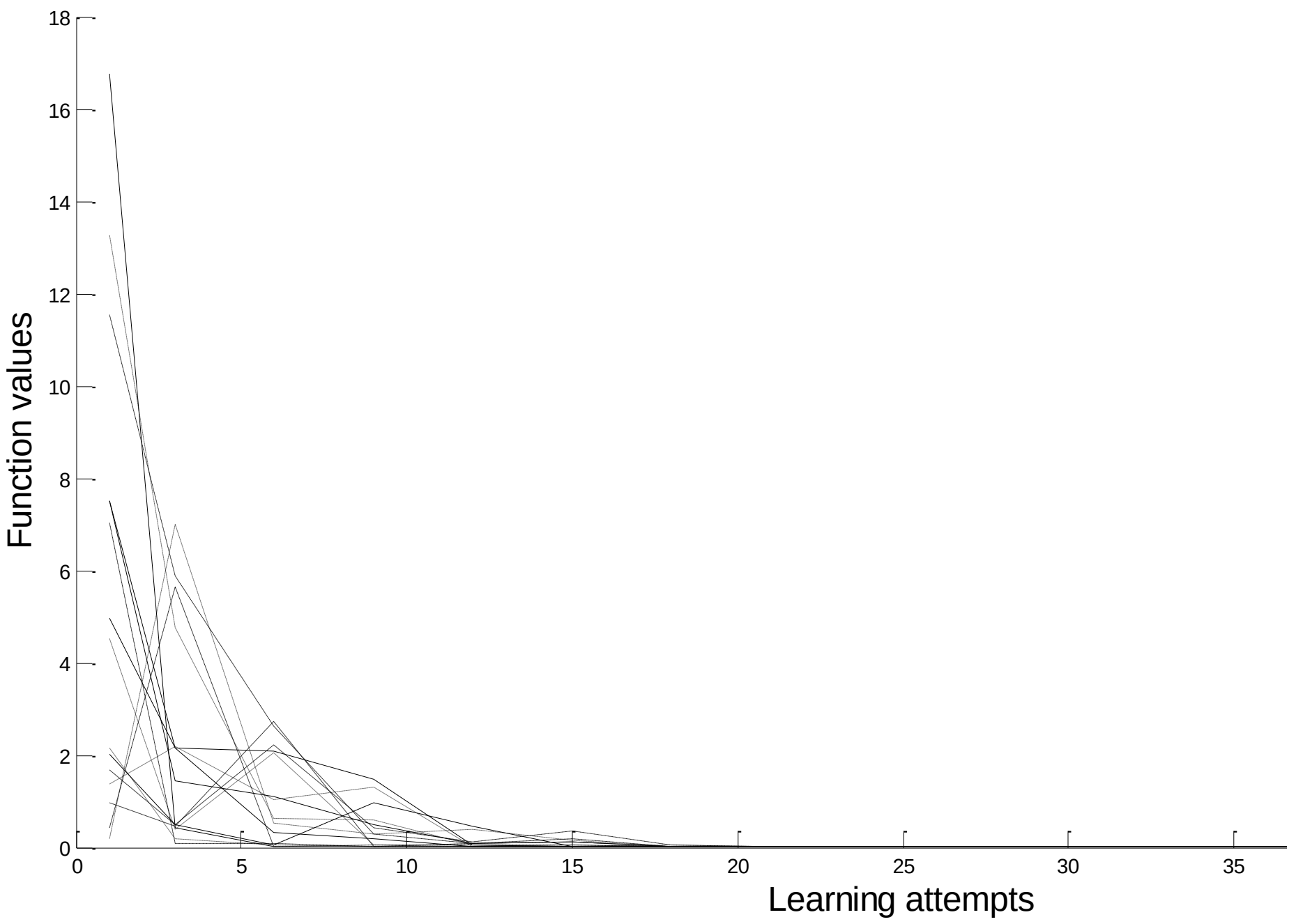

(a) Convergence of all candidates

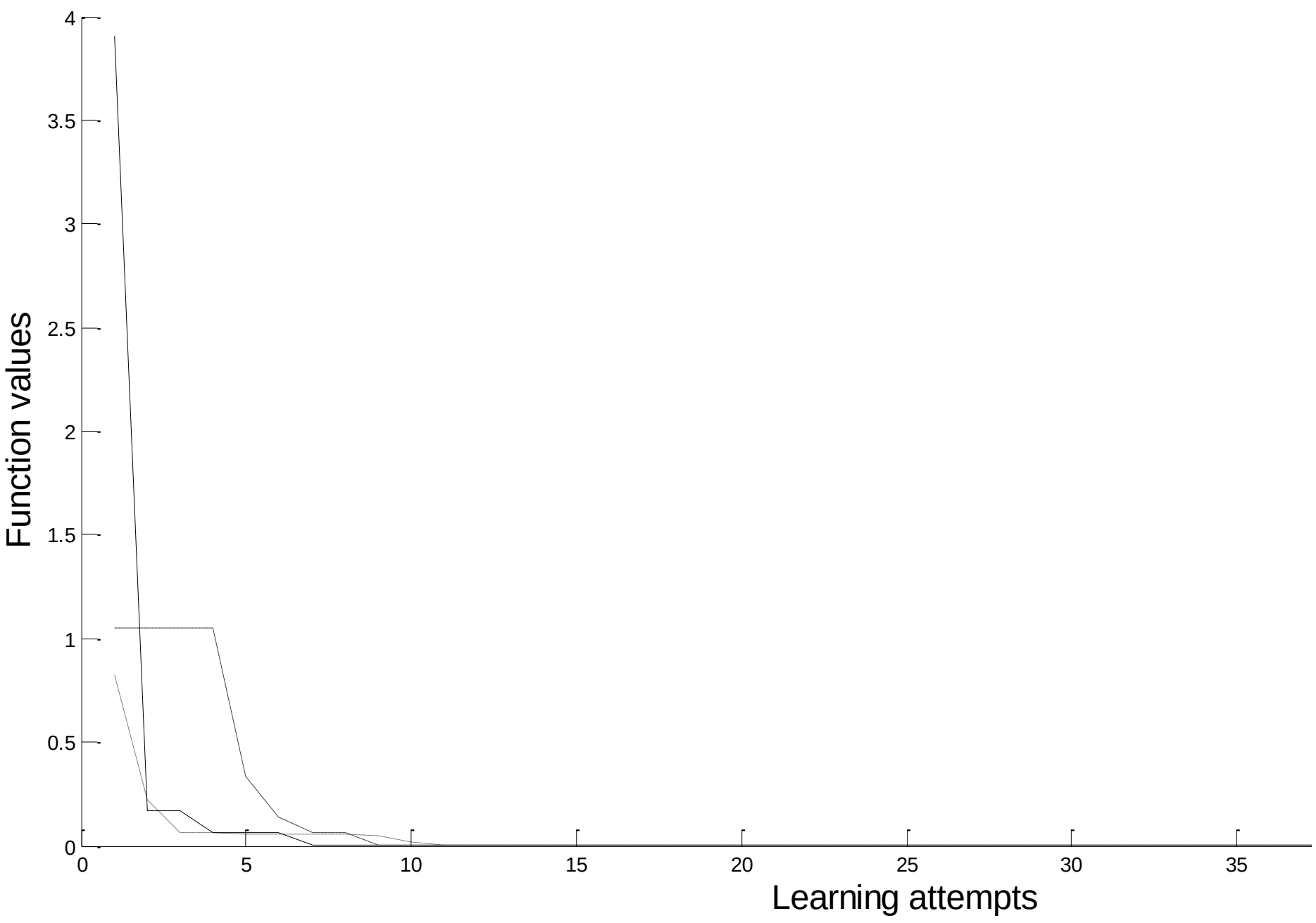

(c) Convergence of best solutions

——— Cohort 1 — — Cohort 2 ------- Cohort 3

Figure 3 Convergence for Beale Function (F6)

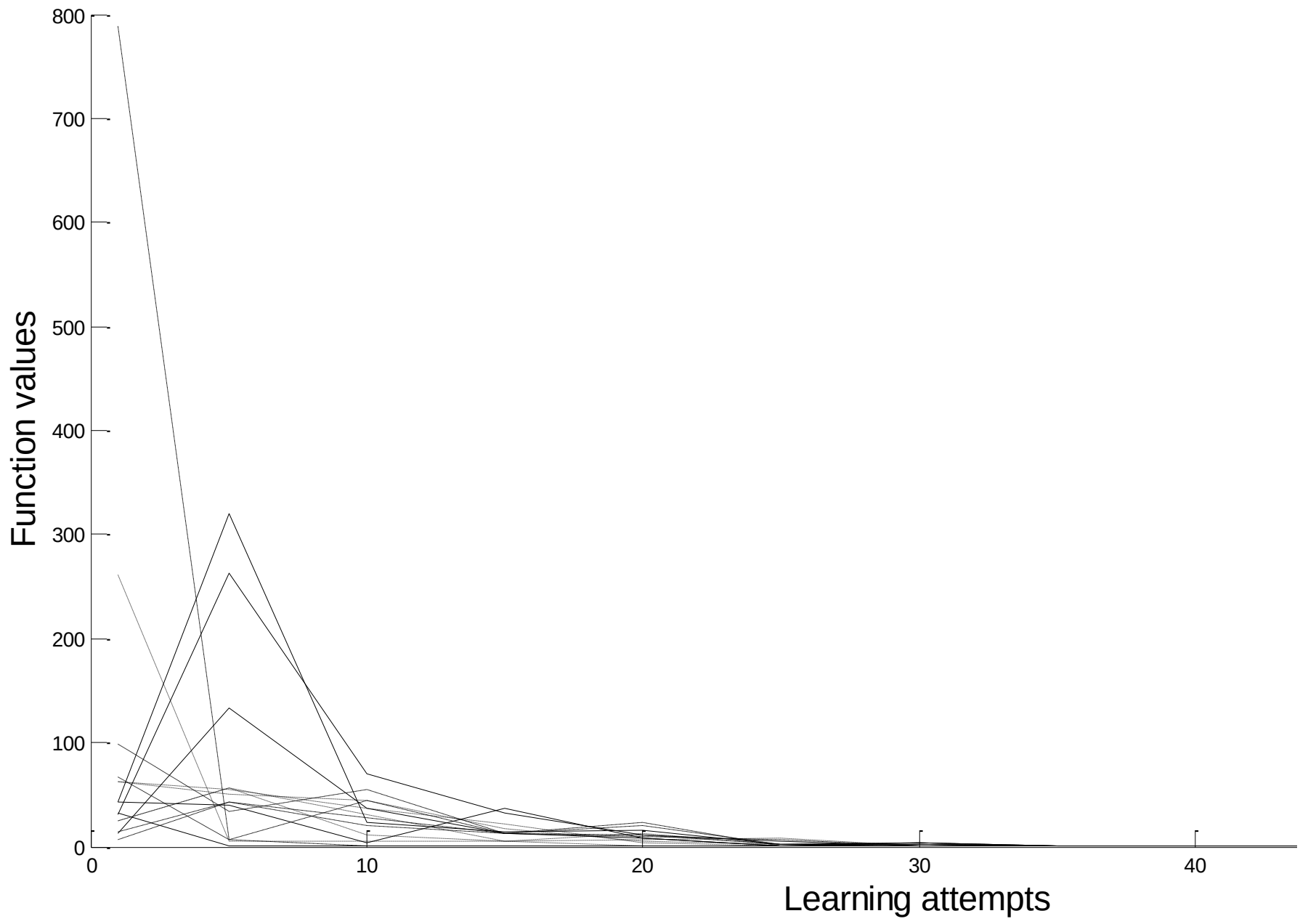

(a) Convergence of all candidates

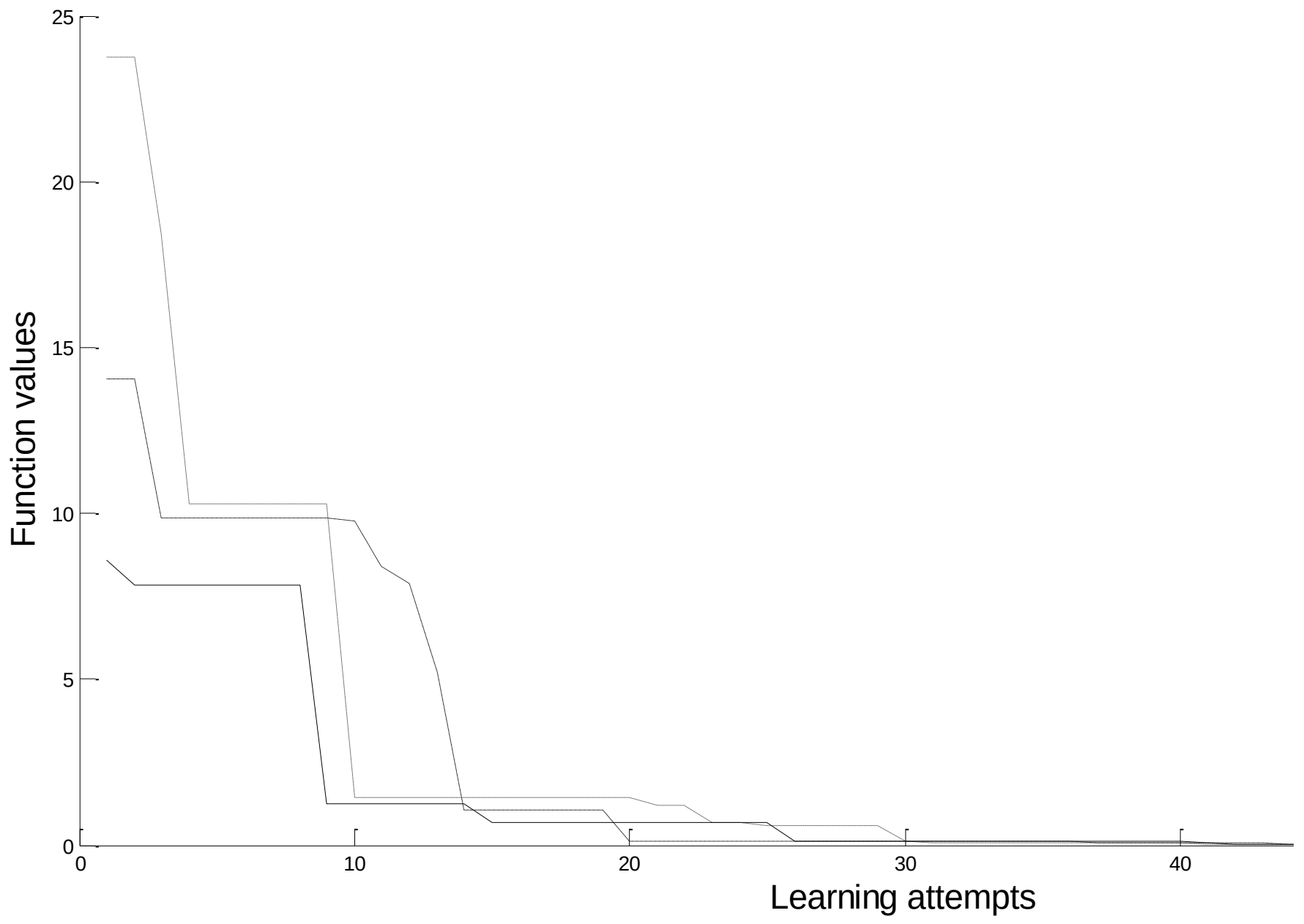

(b) Convergence of best candidates

——— Cohort 1 — — Cohort 2 ------- Cohort 3

Figure 4 Convergence for Fletcher Function (F16)

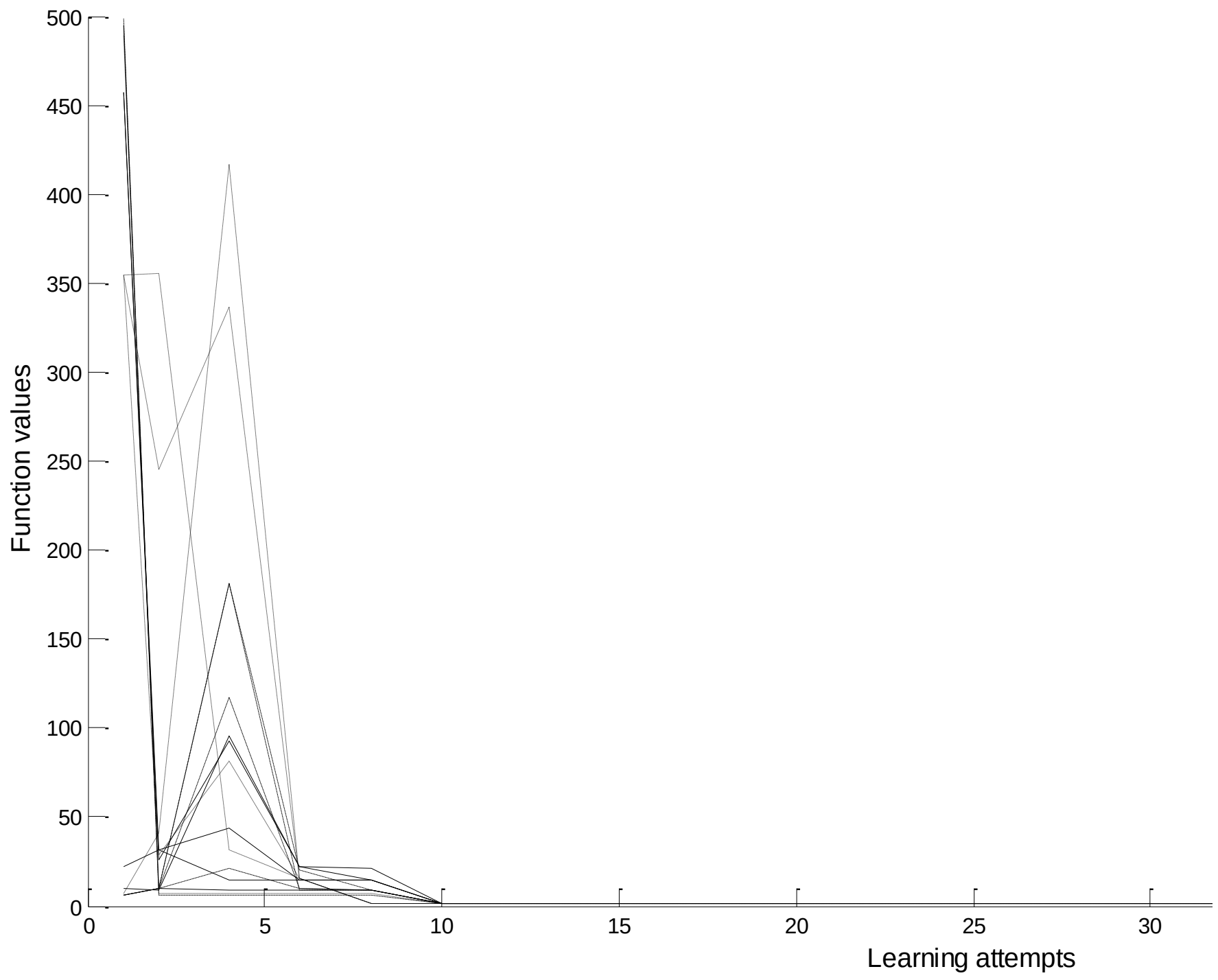

(a) Convergence of all candidates

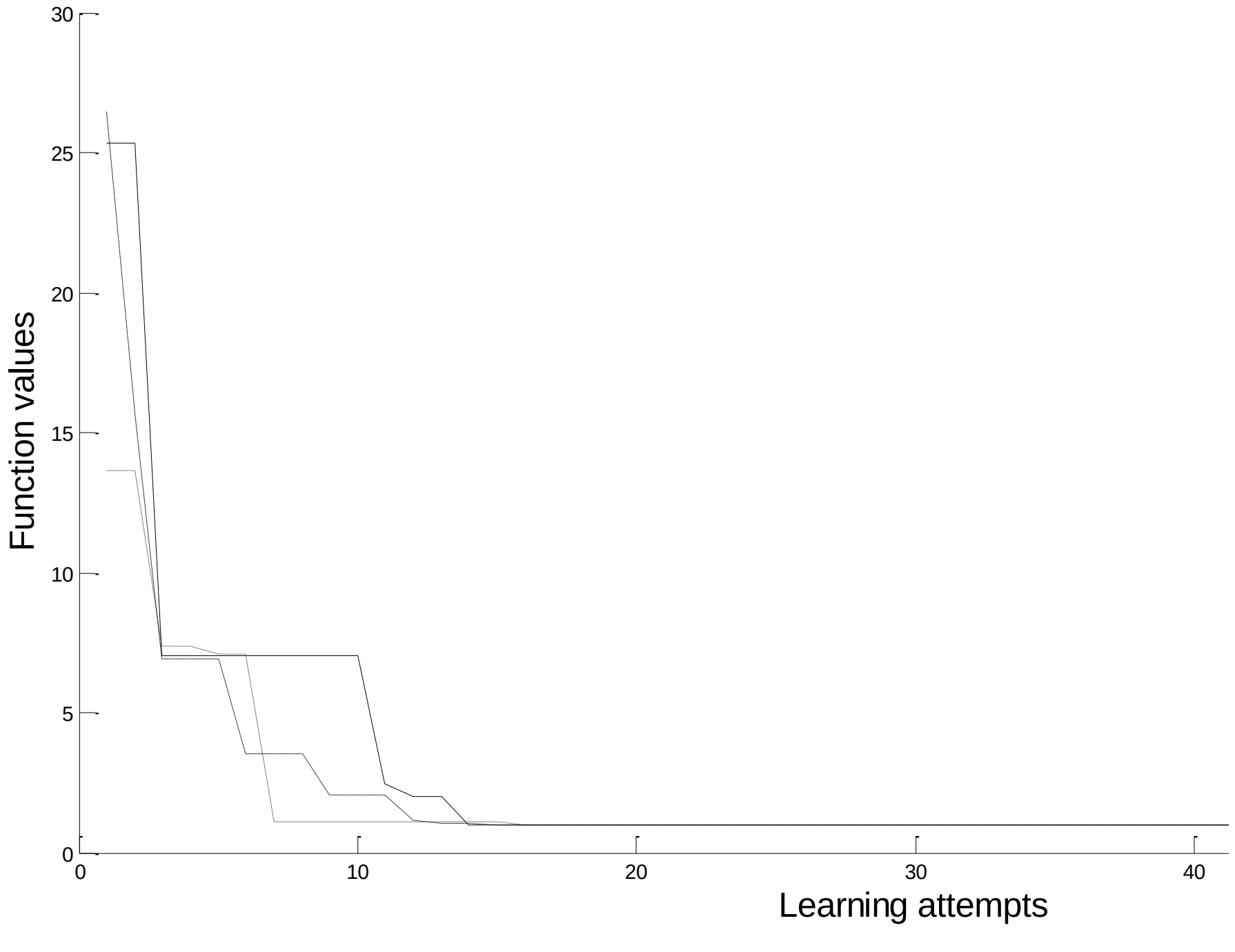

(d)  Convergence of best candidates

Cohort 1 Cohort 2 Cohort 3

Figure 5 Convergence for Foxholes Function (F1)

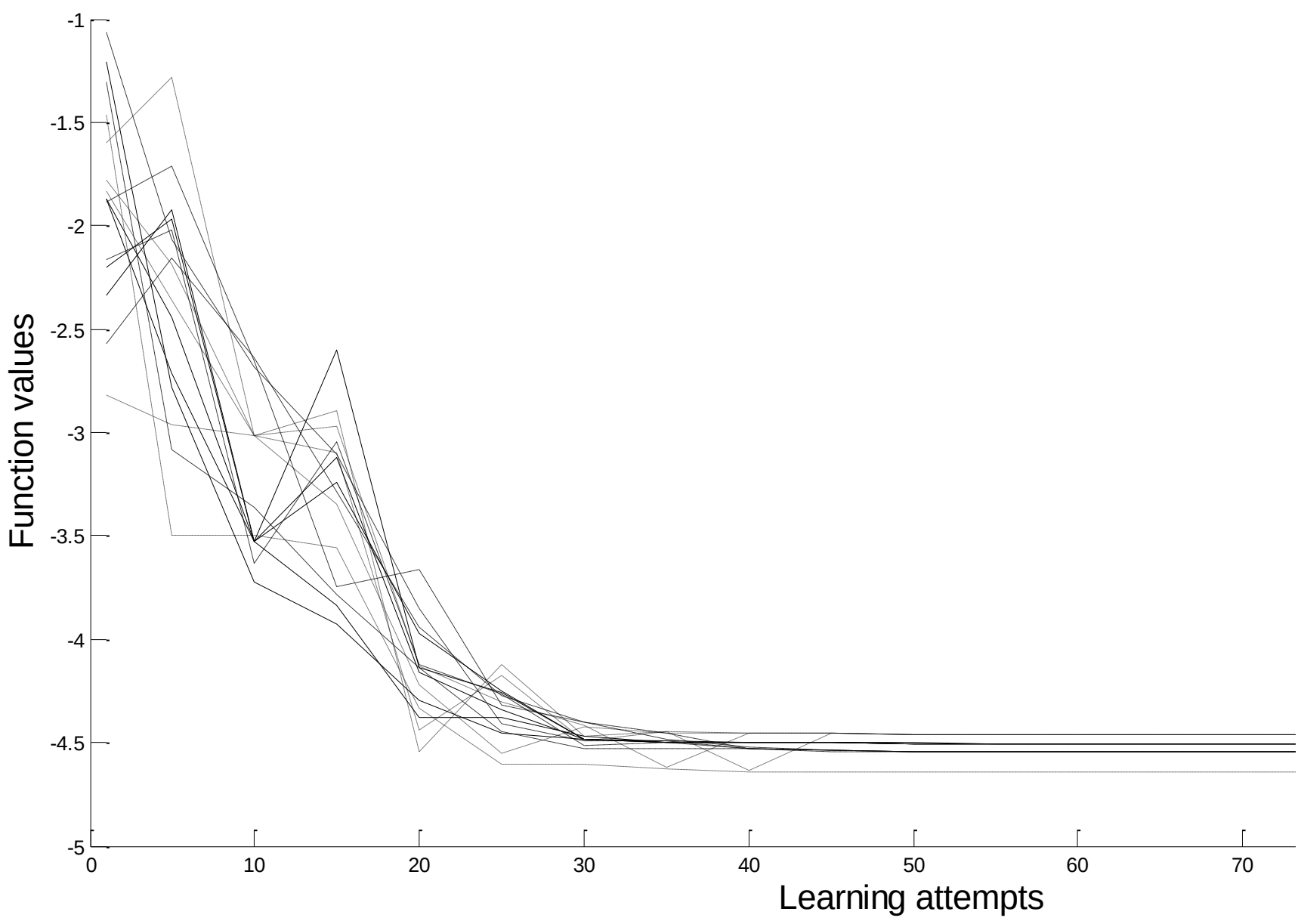

(a) Convergence of all candidates

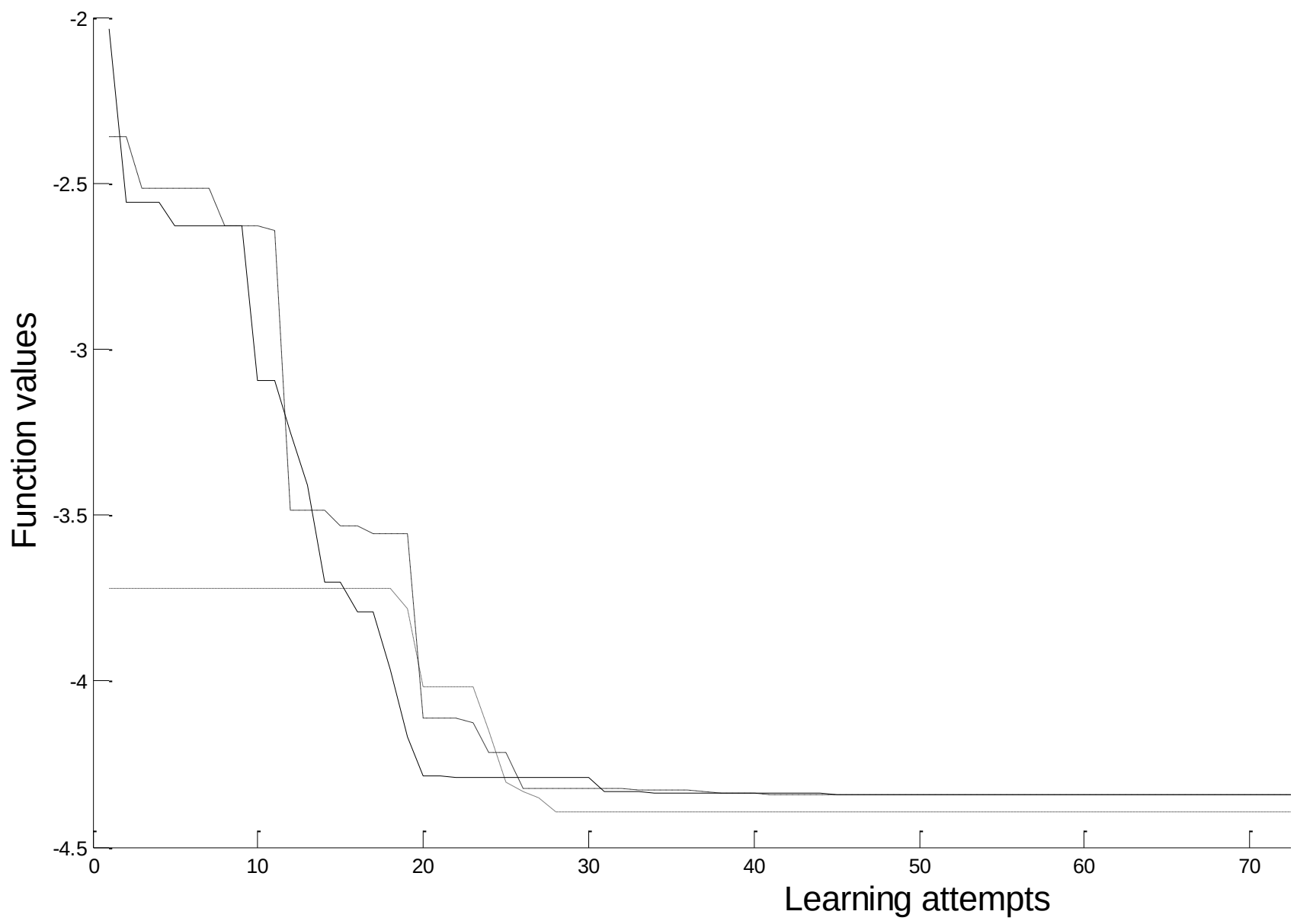

(e) Convergence of best candidates

Cohort 1 Cohort 2 Cohort 3

Figure 6 Convergence for Michalewics Function (F28)

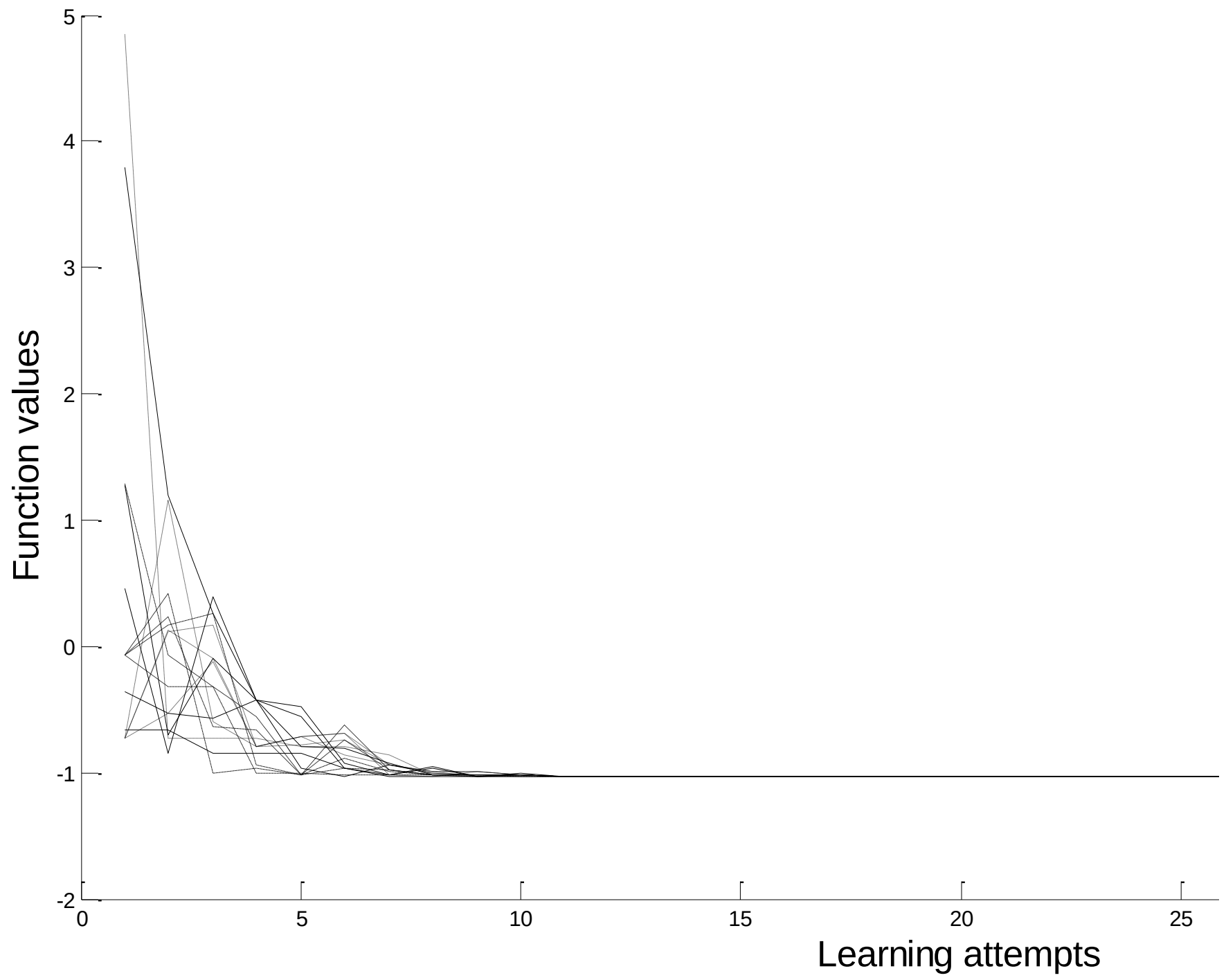

(a) Convergence of all candidates

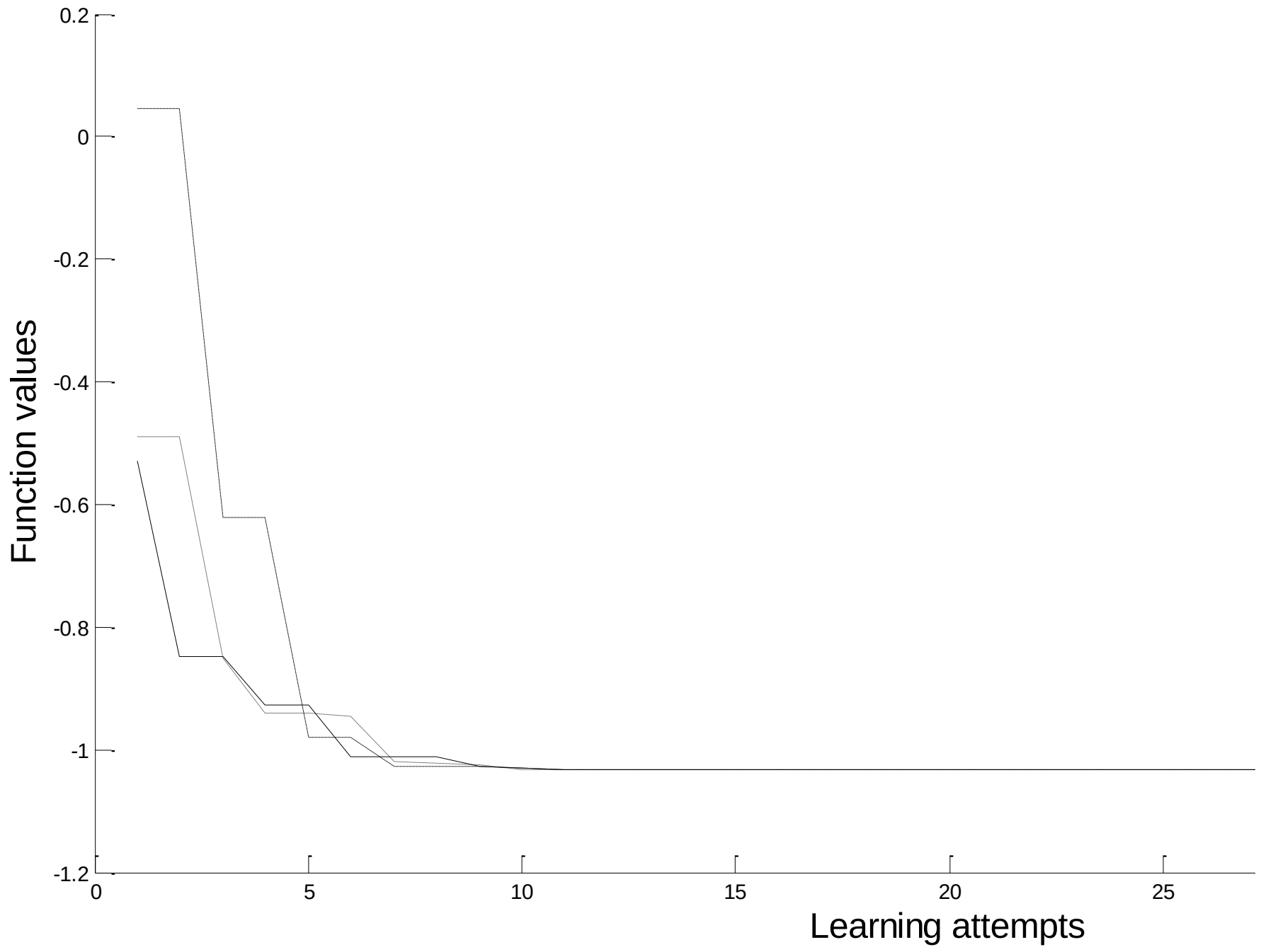

(b) Convergence of best candidates

Figure 7 Convergence for Six-hump camelback Function (F43)

This section provides theoretical comparison of the algorithms being compared with Multi-CI. The method of PSO a swarm of solutions modify their positions in the search space. Every particle of the swarm represents a solution which moves with certain velocity in the search space based on the best solution in the entire swarm as well as the best solution in certain close neighborhood. It imparts exploration as well as exploitation abilities to entire swarm. According to Teo et al. (2016), Li and Yao (2012) and Selvi and Umrani (2010) the PSO may not be efficient solving the problems with discrete search space as well as non-coordinate systems and may need supporting techniques to solve such problems. In this paper Multi-CI is compared with the advanced versions of the PSO referred to as Comprehensive Learning PSO (CLPSO) (Liang et al., 2016) and PSO2011 (Omran and Clerc2011).The technique of CMAES (Igel et al. 2007) is a mathematical-based algorithm which exploits adaptive mutation parameters through computing a covariance matrix. The computational cost of the covariance matrix calculation, sampling using multivariate normal distribution and factorization of covariance matrix may increase exponentially with increase in problem dimension (Selvi and Umrani, 2010). The algorithm of ABC (Karaboga and Akay, 2009) carries out exploration using random search by scout bees and exploitation using employed bees. Some studies highlightedthat the algorithm of ABC can perform well with exploration; however, it is not efficient in local search and exploitation (Murugan and Mohan, 2012). This may make the algorithm trap into local minima.The BSA (Civicioglu, 2013) is a populations based technique which deploys genetic operators to generate initial solutions. Then therandomly chooses individuals to find the new solutions in the search space. The non-uniform crossover makes BSA unique and powerful technique. Similar to the BSA, DE (Storn and Price 1997, Qin and Suganthan 2005) is also a population based technique which exploits genetic operators. The search process is mainly driven by the mutation and selection operation. The crossover operator is further deployed for effectively sorting the trial vectors which helps to choose and retain better solutions. Teo et al. (2016) recently proposed IA. It is inspired from competitive behavior of political party individuals. The local party leaders exploit the concepts such as introspection, local competition and global competition improving the solution quality through exploitation and exploration. In addition, the ordinary party membersmay follow the own party leader or other party leader. This changes the priority of the search lead by certain party. The algorithm performed better as compared to most of the contemporary algorithms.

The Multi-CI algorithm proposed here exhibited certain prominent characteristics and limitations. These are discussed below.

1. In Multi-CI the best individual from within every cohort are moved to a separate pool $\boldsymbol{Z}$. One best candidate is chosen from within each cohort. Then every candidate chooses the best behavior/objective function value from within the $T + T_P$ choices. Thus every candidate competes with its own local best behavior as well as the best behavior chosen from the other cohorts. This gives more exploitation power to the algorithm due to which chance of avoiding the local minima increases with faster convergence.
2. The elite candidate behaviours from Pool $\boldsymbol{Z}$ is carry forwarded to the subsequent learning attempt. This helps not to lose the best behaviour (solution) so far and also the influence of such solutions does not diminish if not followed by any candidate.
3. Initial random walks of individuals around the cohorts ensure exploration of the search space around the candidates.
4. The Multi-CI parameters could be easily tuned which may make it a flexible algorithm for handling variety of problems with different dimensions and complexity.

5. The results highlighted that the algorithm is sufficiently robust with reasonable computational cost and is successful at exploring multi-modal search spaces.
6. The computational performance was essentially governed by sampling interval reduction factor $r$. Its value was chosen based on the preliminary trials of the algorithm.

# 4. Conclusions and Future Directions

A modified version of the Cohort Intelligence (CI) algorithm referred to as Multi-Cohort Intelligence (Multi-CI) was proposed. In the proposed Multi-CI approach intra-group learning and inter-group learning mechanisms were implemented. It is more realistic representation of the learning through interaction and competition of the cohort candidates. It imparted the exploitation and exploration capabilities to the algorithm. The approach was validated by solving two sets of test problems from CEC 2005. Wilcoxon statistical tests were conducted for comparing the performance of the algorithm with the existing algorithms. The performance of Multi-CI was exceedingly better as compared to PSO2011, CMAES, ABC, JDE, CLPSO and SADE in terms of objective function value (best and mean), robustness, as well as computational time. The performance of the Multi-CI was marginally better as compared to BSA and IA. The solution quality highlighted that the Multi-CI is a robust approach with reasonable computational cost and could quickly reach in the close neighborhood of the global optimum solution.

A generalized constraint handling mechanism needs to be developed and incorporated into the algorithm. This can help Multi-CI to solve real world problems which are generally constrained in nature. The Multi-CI algorithm could be further modified for solving constrained test problems as well as real world problems. The constrained Multi-CI version could be further extended to solve complex structural optimization problems (Azad 2017, 2018). This work is currently underway. A self-adaptive mechanism needs to be developed for the selection of the sampling interval reduction factor $r$.

# Appendix A: Illustration of Multi-CI Algorithm

An illustrative example (Sphere function with 2 variables: $mize\ \sum_{i=1}^{2} x_i^2$ , $Subject\ to - 5.12 \le x_i \le 5.12\ , i = 1, 2$) of the Multi-CI procedure discussed in Section 2 is detailed below. It includes every details of first learning attempt followed by evaluation of every step (1 to 8) is listed in Table A.1 till convergence along with the convergence plot in Figure A.1.The Multi-CI parameters chosen were as follows: number of cohorts $K = 3$, number of candidates $C^k = 3$, reduction factor value $r = 0.98$, quality variation parameters $T = 2$ and $T_p = 4$, the algorithm stopped when the objective function value is less than $10^{-16}$.

**Learning Attempt** $l = 1$

$$\boldsymbol{X} = \begin{bmatrix} 0.4426 & -2.7631 & -4.4698 & 1.2841 & -4.4839 & 4.8435 \\ 1.7060 & 4.5039\ , & -4.4155 & -0.7989, & -4.0203 & -1.1923 \\ -2.2525 & -1.3291 & -2.4503 & 3.8907 & 0.1308 & -1.8813 \end{bmatrix}$$

(**Step 1, Eq 2**):

$$\boldsymbol{F} = \begin{bmatrix} 7.8304 & 21.6280 & 43.5648 \\ 23.1957 & 20.1344 & 17.5841 \\ 6.8402 & 21.1409 & 3.5564 \end{bmatrix}$$

(**Step 2, Eq 3**): $\boldsymbol{F^Z} = [6.8402 \quad 20.1344 \quad 3.5564]$

(**Step 3, Eq 4**):

$$p_1^1 = 0.7476$$

$$p_1^2 = 0.2524$$

$$p_2^1 = 0.4943$$

$$p_2^3 = 0.5057$$

$$p_3^1 = 0.2876$$

$$p_3^2 = 0.7124$$

(**Step 4, Eq 5**):

Now every cohort $k$ $(k = 1, \dots, K)$ is left with $C^k - 1$ candidates.

Consider candidate $C_1$ in cohort 1 and the associated qualities $\boldsymbol{X}_1^{C_1} = [x_{1,1}^1, x_{2,1}^1] = [0.4426, -2.7681]$.

Sampling interval for $x_{1,1}^1$ is given by

$$[\psi_1^{1,lower}, \ \psi_1^{1,upper}] = \left[0.4426 - \left(\left\|\frac{5.12 - (-5.12)}{2}\right\|\right) \times 0.98, \quad 0.4426 + \left(\left\|\frac{5.12 - (-5.12)}{2}\right\|\right) \times 0.98\right]$$

$$[\psi_1^{1,lower}, \ \psi_1^{1,upper}] = [-4.575, 5.4602]$$

$$= [-4.575, 5.12] (\because 5.4602 \text{ is out of the interval})$$

Sampling interval for $x_{2,1}^1$ is given by

$$[\psi_2^{1,lower}, \ \psi_2^{1,upper}] = \left[-2.7681 - \left(\left\|\frac{5.12 - (-5.12)}{2}\right\|\right) \times 0.98, \quad -2.7681 + \left(\left\|\frac{5.12 - (-5.12)}{2}\right\|\right) \times 0.98\right]$$

$$[\psi_2^{1,lower}, \ \psi_2^{1,upper}] = [-7.7807, 2.2545]$$

$$= [-5.12, 2.2545] (\because -7.7807 \text{ is out of the interval})$$

Using roulette wheel selection, candidate $C_2$ in cohort 1 chooses to follow candidate $C_1$. So sampling intervals for candidate $C_2$ will be the same as that of candidate $C_1$. The $T = 2$ sampling intervals of every candidate $C^k - 1, \ k \ (k = 1, \dots, K)$ are as follows:

Cohort 1      Cohort 2      Cohort 3

$$\begin{bmatrix} \begin{bmatrix} [-4.575, 5.12] & [-5.12, 2.2545] \\ [-4.575, 5.12] & [-5.12, 2.2545] \end{bmatrix} & \begin{bmatrix} [-5.12, 2.5673] & [-1.1269, 5.12] \\ [-5.12, 2.5673] & [-1.1269, 5.12] \end{bmatrix} & \begin{bmatrix} [-5.12, 0.9973] & [-5.12, 3.8253] \\ [-5.12, 0.9973] & [-5.12, 3.8253] \end{bmatrix} \\ \begin{bmatrix} [-4.575, 5.12] & [-5.12, 2.2545] \\ [-4.575, 5.12] & [-5.12, 2.2545] \end{bmatrix} & \begin{bmatrix} [-5.12, 2.5673] & [-1.1269, 5.12] \\ [-5.12, 2.5673] & [-1.1269, 5.12] \end{bmatrix} & \begin{bmatrix} [-5.12, 0.9973] & [-5.12, 3.8253] \\ [-5.12, 0.9973] & [-5.12, 3.8253] \end{bmatrix} \end{bmatrix}$$

(**Step 4, Eq 6**): Every candidate samples the qualities from within these sampling intervals and forms quality matrix $\boldsymbol{Z}^T$ as follows:

$$\boldsymbol{Z}^T = \begin{bmatrix} \begin{bmatrix} 3.7775 & -4.9854 \\ -0.0271 & 2.1268 \end{bmatrix} & \begin{bmatrix} -2.4902 & 4.1556 \\ -2.8815 & 4.4581 \end{bmatrix} & \begin{bmatrix} -1.6575 & -0.5544 \\ -1.3713 & -4.2098 \end{bmatrix} \\ \begin{bmatrix} -1.5062 & -0.00348 \\ -2.2765 & -2.2535 \end{bmatrix} & \begin{bmatrix} -4.7380 & 4.1285 \\ -1.7774 & 2.8955 \end{bmatrix} & \begin{bmatrix} -4.4906 & -2.1964 \\ -4.65257 & -1.0430 \end{bmatrix} \end{bmatrix}$$

(**Step 4, Eq 7**):

$$\boldsymbol{F}^T = \begin{bmatrix} \begin{bmatrix} 39.1242 \\ 4.5240 \end{bmatrix} & \begin{bmatrix} 23.4702 \\ 28.1783 \end{bmatrix} & \begin{bmatrix} 3.0562 \\ 19.6036 \end{bmatrix} \\ \begin{bmatrix} 2.2700 \\ 11.6100 \end{bmatrix} & \begin{bmatrix} 39.4941 \\ 11.5436 \end{bmatrix} & \begin{bmatrix} 24.9901 \\ 22.4855 \end{bmatrix} \end{bmatrix}$$

(**Step 5, Eq 8**):

$$p_1^1 = 0.3065$$

$$p_2^2 = 0.1041$$

$$p_3^3 = 0.5894$$

(**Step 6, Eq 9**):

$$\boldsymbol{Z}^{T_p} = \begin{bmatrix} \begin{bmatrix} -0.9714 & 0.1627 \\ -4.5412 & -2.4676 \\ -4.0294 & -0.5608 \\ -4.2622 & -1.8622 \end{bmatrix} & \begin{bmatrix} 0.0057 & 0.6931 \\ -4.8913 & 2.3370 \\ -4.4565 & -1.9554 \\ -1.5795 & 1.0522 \end{bmatrix} & \begin{bmatrix} -2.9102 & -3.4787 \\ -2.9835 & -0.8647 \\ -2.1960 & -4.7693 \\ -2.2042 & -3.3672 \end{bmatrix} \\ \begin{bmatrix} 1.1820 & -2.2922 \\ -0.9737 & -3.0238 \\ -1.7371 & -1.50645 \\ 0.0012 & 2.3825 \end{bmatrix} & \begin{bmatrix} 1.2259 & -3.5126 \\ -0.6327 & -0.3592 \\ -0.2333 & -2.3488 \\ -2.5639 & -1.2459 \end{bmatrix} & \begin{bmatrix} 0.7412 & -2.7817 \\ -3.0381 & -0.8921 \\ -0.3807 & 2.1570 \\ -2.7075 & -0.9127 \end{bmatrix} \end{bmatrix}$$

(**Step 6, Eq 10**):

$$\boldsymbol{F}^{T_p} = \begin{bmatrix} \begin{bmatrix} 0.9703 \\ 26.7122 \\ 16.5508 \\ 21.6350 \end{bmatrix} & \begin{bmatrix} 0.4805 \\ 29.387 \\ 23.6842 \\ 3.6022 \end{bmatrix} & \begin{bmatrix} 20.5714 \\ 9.6495 \\ 27.5692 \\ 18.3430 \end{bmatrix} \\ \begin{bmatrix} 9.9358 \\ 10.0916 \\ 5.2871 \\ 5.6763 \end{bmatrix} & \begin{bmatrix} 13.8416 \\ 0.5294 \\ 5.5713 \\ 8.1260 \end{bmatrix} & \begin{bmatrix} 8.2876 \\ 10.0264 \\ 4.7976 \\ 8.1636 \end{bmatrix} \end{bmatrix}$$

(**Step 8, Eq 12**):

$$\boldsymbol{F} = \begin{bmatrix} 0.9703 & 0.4805 & 3.0562 \\ 2.2700 & 0.5294 & 4.7976 \\ 6.8402 & 20.1344 & 3.5564 \end{bmatrix}$$

$$Minimum = 0.4805$$

**End of Learning Attempt**

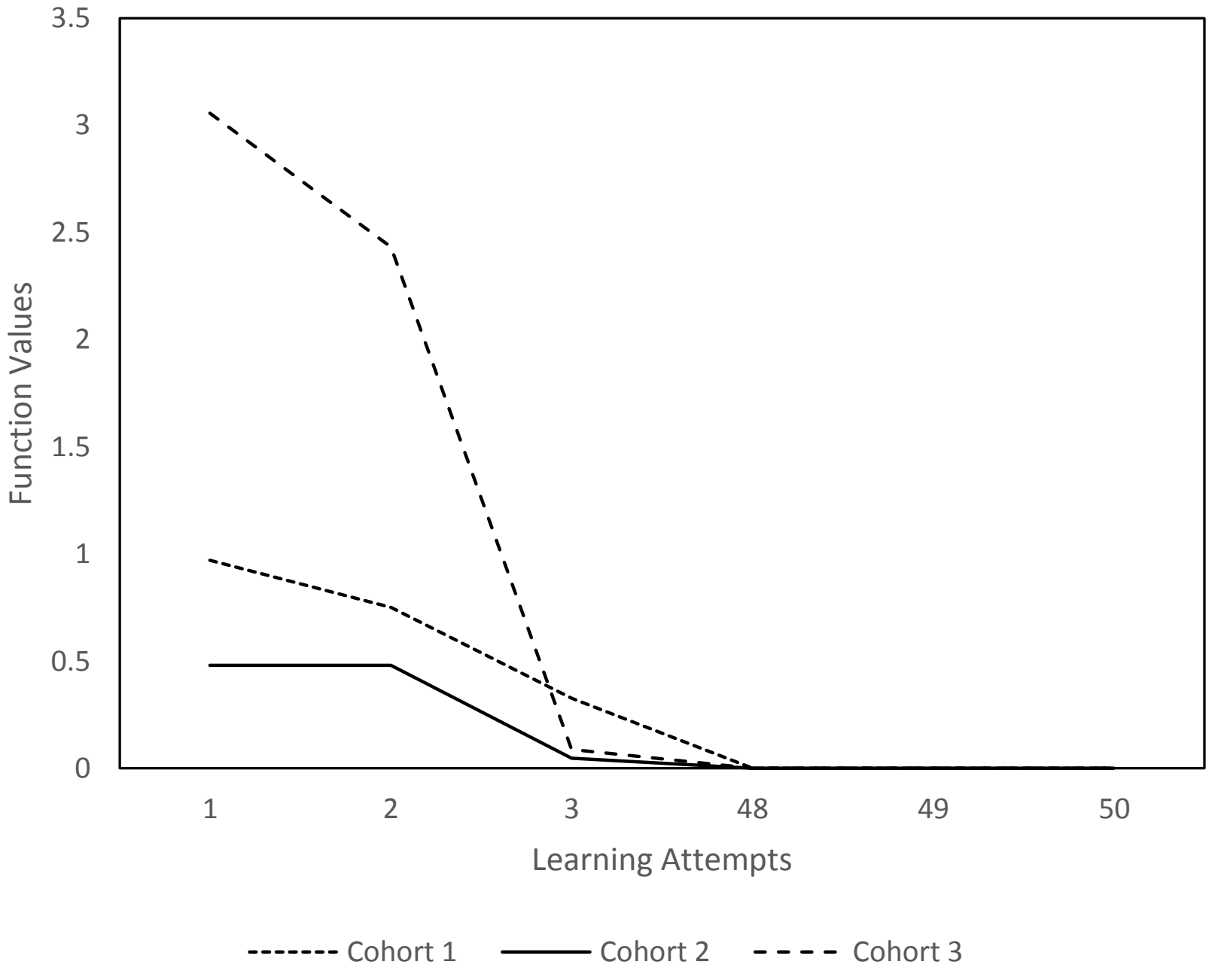

Figure A.1 Convergence for Sphere Function

**Table A.1 Illustration of Multi-CI Algorithm solving Sphere Function**

| **Learning Attempt ($\boldsymbol{l}$)** | $\boldsymbol{X}$ | $\boldsymbol{F}$ (Step 1, Eq 2) | $\boldsymbol{F^Z}$ (Step 2, Eq 3) | |
|---|---|---|---|---|
| $l=2$ | $\begin{bmatrix} -0.9715 & -0.1627 \\ -0.9715 & -0.1627 \\ -2.2525 & -1.3291 \end{bmatrix}, \begin{bmatrix} 0.0057 & 0.6932 \\ 0.0057 & 0.6932 \\ -4.4155 & -0.7989 \end{bmatrix}, \begin{bmatrix} -1.6579 & -0.5544 \\ -2.9836 & -0.8647 \\ 0.1308 & -1.8813 \end{bmatrix}$ | $\begin{bmatrix} 0.9703 & 0.4805 & 3.0562 \\ 0.9703 & 0.4805 & 9.6465 \\ 6.8402 & 20.1344 & 3.5564 \end{bmatrix}$ | $\begin{bmatrix} 0.9703 & 0.4805 & 3.0562 \end{bmatrix}$ | |
| | $\boldsymbol{F^T}$ (Step 4, Eq 7) | $\boldsymbol{F^{T_Z}}$ (Step 6, Eq 10) | $\boldsymbol{F}$ (Step 8, Eq 12) | Minimum |
| | $\begin{bmatrix} \begin{bmatrix} 6.6966 \\ 20.3143 \end{bmatrix} & \begin{bmatrix} 3.2039 \\ 13.4206 \end{bmatrix} & \begin{bmatrix} 17.4818 \\ 20.2246 \end{bmatrix} \\ \begin{bmatrix} 4.6984 \\ 4.1166 \end{bmatrix} & \begin{bmatrix} 3.5017 \\ 3.7385 \end{bmatrix} & \begin{bmatrix} 3.5744 \\ 4.4140 \end{bmatrix} \end{bmatrix}$ | $\begin{bmatrix} \begin{bmatrix} 6.6475 \\ 17.5217 \\ 2.9930 \\ 0.7515 \end{bmatrix} & \begin{bmatrix} 7.7984 \\ 2.9576 \\ 1.1647 \\ 7.7003 \end{bmatrix} & \begin{bmatrix} 8.7400 \\ 3.3670 \\ 6.1257 \\ 2.4325 \end{bmatrix} \\ \begin{bmatrix} 0.9872 \\ 1.0732 \\ 3.0378 \\ 0.3062 \end{bmatrix} & \begin{bmatrix} 3.9120 \\ 0.4951 \\ 4.5588 \\ 1.2267 \end{bmatrix} & \begin{bmatrix} 10.2398 \\ 0.6033 \\ 2.4180 \\ 0.6774 \end{bmatrix} \end{bmatrix}$ | $\begin{bmatrix} 0.7515 & 1.1647 & 2.4325 \\ 0.3062 & 0.4951 & 0.6033 \\ 0.9703 & 0.4805 & 3.0562 \end{bmatrix}$ | 0.3062 |
| $l=3$ | $\boldsymbol{X}$ | $\boldsymbol{F}$(Step 1, Eq 2) | $\boldsymbol{F^Z}$ (Step 2, Eq 3) | |
| | $\begin{bmatrix} -0.4357 & -0.7494 \\ -0.4357 & -0.7494 \\ -0.9715 & 0.1627 \end{bmatrix}, \begin{bmatrix} -0.3565 & -1.0186 \\ -0.3565 & -1.0186 \\ 0.0057 & 0.6932 \end{bmatrix}, \begin{bmatrix} 1.0491 & 1.1541 \\ 1.0491 & 1.1541 \\ -1.6579 & -0.5544 \end{bmatrix}$ | $\begin{bmatrix} 0.7515 & 1.1647 & 2.4325 \\ 0.7515 & 1.1647 & 2.4325 \\ 0.9703 & 0.4805 & 3.0562 \end{bmatrix}$ | $\begin{bmatrix} 0.7515 & 0.4805 & 2.4325 \end{bmatrix}$ | |
| | $\boldsymbol{F^T}$ (Step 4, Eq 7) | $\boldsymbol{F^{T_Z}}$ (Step 6, Eq 10) | $\boldsymbol{F}$ (Step 8, Eq 12) | Minimum |
| | $\begin{bmatrix} \begin{bmatrix} 0.3276 \\ 0.6218 \end{bmatrix} & \begin{bmatrix} 1.7695 \\ 1.0992 \end{bmatrix} & \begin{bmatrix} 0.0881 \\ 8.6904 \end{bmatrix} \\ \begin{bmatrix} 1.2772 \\ 6.1061 \end{bmatrix} & \begin{bmatrix} 1.6018 \\ 2.0007 \end{bmatrix} & \begin{bmatrix} 2.5886 \\ 11.1223 \end{bmatrix} \end{bmatrix}$ | $\begin{bmatrix} \begin{bmatrix} 10.4735 \\ 6.1670 \\ 1.6276 \\ 0.7038 \end{bmatrix} & \begin{bmatrix} 3.1505 \\ 0.6858 \\ 1.5501 \\ 0.0470 \end{bmatrix} & \begin{bmatrix} 0.9388 \\ 2.2083 \\ 6.8189 \\ 5.8672 \end{bmatrix} \\ \begin{bmatrix} 4.3181 \\ 2.0717 \\ 5.4549 \\ 1.4087 \end{bmatrix} & \begin{bmatrix} 4.6674 \\ 0.4319 \\ 6.6115 \\ 1.2808 \end{bmatrix} & \begin{bmatrix} 0.1611 \\ 7.0118 \\ 2.3274 \\ 0.3228 \end{bmatrix} \end{bmatrix}$ | $\begin{bmatrix} 0.3276 & 0.0470 & 0.0881 \\ 1.2772 & 0.4319 & 0.1611 \\ 0.7515 & 0.4805 & 2.4325 \end{bmatrix}$ | 0.0470 |
| $\vdots$ | $\vdots$ | $\vdots$ | $\vdots$ | $\vdots$ |
| $\vdots$ | $\vdots$ | $\vdots$ | $\vdots$ | $\vdots$ |
| | | | | |

| ⋮ | ⋮ | ⋮ | ⋮ | ⋮ |
|---|---|---|---|---|
| $l = 48$ | $\boldsymbol{X}$ | $\boldsymbol{F}$ (Step 1, Eq 2) | $\boldsymbol{F^Z}$ (Step 2, Eq 3) | |
| | | | | |
| | $\begin{bmatrix} 4.98\text{E}-8 & 1.03\text{E}-7 \\ -4.33\text{E}-8 & 4.71\text{E}-8 \\ -2.59\text{E}-8 & 6.17\text{E}-8 \end{bmatrix}, \begin{bmatrix} -2.95\text{E}-9 & -7.71\text{E}-8 \\ -2.95\text{E}-9 & -7.71\text{E}-8 \\ 6.5\text{E}-8 & 8.76\text{E}-8 \end{bmatrix}, \begin{bmatrix} -8.29\text{E}-8 & -1.11\text{E}-7 \\ -8.29\text{E}-8 & -1.11\text{E}-7 \\ 1.03\text{E}-7 & -5.9\text{E}-9 \end{bmatrix}$ | $\begin{bmatrix} 1.31E-14 & 5.96E-15 & 1.90E-14 \\ 4.09E-15 & 5.96E-15 & 1.90E-14 \\ 4.48E-15 & 1.12E-14 & 1.06E-14 \end{bmatrix}$ | $[4.09\text{E}-15 \quad 5.96\text{E}-15 \quad 1.06\text{E}-14]$ | |
| | $\boldsymbol{F}^T$ (Step 4, Eq 7) | $\boldsymbol{F}^{Tz}$ (Step 6, Eq 10) | $\boldsymbol{F}$ (Step 8, Eq 12) | Minimum |
| | $\begin{bmatrix} \begin{bmatrix} 2.05\text{E}-14 \\ 5.64\text{E}-14 \end{bmatrix} & \begin{bmatrix} 3.59\text{E}-14 \\ 5.99\text{E}-14 \end{bmatrix} & \begin{bmatrix} 9.92\text{E}-15 \\ 5.32\text{E}-14 \end{bmatrix} \\ \begin{bmatrix} 4.13\text{E}-14 \\ 2.92\text{E}-15 \end{bmatrix} & \begin{bmatrix} 3.54\text{E}-14 \\ 9.02\text{E}-15 \end{bmatrix} & \begin{bmatrix} 2.12\text{E}-14 \\ 3.58\text{E}-14 \end{bmatrix} \end{bmatrix}$ | $\begin{bmatrix} \begin{bmatrix} 1.80\text{E}-14 \\ 5.62\text{E}-15 \\ 3.76\text{E}-14 \\ 1.57\text{E}-14 \end{bmatrix} & \begin{bmatrix} 2.54\text{E}-14 \\ 2.00\text{E}-14 \\ 1.95\text{E}-15 \\ 7.82\text{E}-14 \end{bmatrix} & \begin{bmatrix} 1.05\text{E}-13 \\ 4.06\text{E}-14 \\ 2.19\text{E}-14 \\ 1.70\text{E}-14 \end{bmatrix} \\ \begin{bmatrix} 2.49\text{E}-15 \\ 2.38\text{E}-14 \\ 1.67\text{E}-15 \\ 2.30\text{E}-14 \end{bmatrix} & \begin{bmatrix} 3.56\text{E}-15 \\ 1.35\text{E}-14 \\ 7.74\text{E}-15 \\ 2.36\text{E}-14 \end{bmatrix} & \begin{bmatrix} 3.86\text{E}-14 \\ 6.24\text{E}-15 \\ 5.02\text{E}-14 \\ 3.03\text{E}-14 \end{bmatrix} \end{bmatrix}$ | $\begin{bmatrix} 5.62\text{E}-15 & 1.95\text{E}-15 & 9.92\text{E}-15 \\ 1.67\text{E}-15 & 3.56\text{E}-15 & 6.24\text{E}-15 \\ 4.09\text{E}-15 & 5.96\text{E}-15 & 1.06\text{E}-14 \end{bmatrix}$ | 1.67E-15 |
| $l = 49$ | $\boldsymbol{X}$ | $\boldsymbol{F}$ (Step 1, Eq 2) | $\boldsymbol{F^Z}$ (Step 2, Eq 3) | |
| | $\begin{bmatrix} 5.05\text{E}-8 & -5.54\text{E}-8 \\ 1.28\text{E}-9 & -5.4\text{E}-8 \\ -4.33\text{E}-8 & 4.71\text{E}-8 \end{bmatrix}, \begin{bmatrix} 1.92\text{E}-8 & -3.98\text{E}-8 \\ -1.92\text{E}-8 & -3.98\text{E}-8 \\ 2.95\text{E}-9 & -7.77\text{E}-8 \end{bmatrix}, \begin{bmatrix} -9.49\text{E}-8 & 3\text{E}-8 \\ -1.28\text{E}-7 & 2.46\text{E}-8 \\ 1.03\text{E}-7 & -5.9\text{E}-9 \end{bmatrix}$ | $\begin{bmatrix} 5.62E-15 & 1.95E-15 & 9.92E-15 \\ 2.92E-15 & 1.95E-15 & 1.72E-15 \\ 4.09E-15 & 5.96E-14 & 1.06E-14 \end{bmatrix}$ | $[2.92\text{E}-15 \quad 1.95\text{E}-15 \quad 1.72\text{E}-15]$ | |
| | $\boldsymbol{F}^T$ (Step 4, Eq 7) | $\boldsymbol{F}^{Tz}$ (Step 6, Eq 10) | $\boldsymbol{F}$ (Step 8, Eq 12) | Minimum |
| | $\begin{bmatrix} \begin{bmatrix} 3.34\text{E}-14 \\ 4.73\text{E}-15 \end{bmatrix} & \begin{bmatrix} 7.96\text{E}-15 \\ 2.11\text{E}-14 \end{bmatrix} & \begin{bmatrix} 1.78E-14 \\ 4.17E-14 \end{bmatrix} \\ \begin{bmatrix} 2.47E-14 \\ 7.68E-15 \end{bmatrix} & \begin{bmatrix} 1.19\text{E}-14 \\ 8.36\text{E}-16 \end{bmatrix} & \begin{bmatrix} 6.41\text{E}-15 \\ 1.33\text{E}-14 \end{bmatrix} \end{bmatrix}$ | $\begin{bmatrix} \begin{bmatrix} 2.38\text{E}-15 \\ 2.25\text{E}-14 \\ 1.86\text{E}-14 \\ 2.87\text{E}-14 \end{bmatrix} & \begin{bmatrix} 4.28\text{E}-14 \\ 1.15\text{E}-14 \\ 1.32\text{E}-14 \\ 4.69\text{E}-14 \end{bmatrix} & \begin{bmatrix} 2.79\text{E}-14 \\ 3.41\text{E}-14 \\ 1.09\text{E}-14 \\ 8.24\text{E}-16 \end{bmatrix} \\ \begin{bmatrix} 2.65\text{E}-14 \\ 3.08\text{E}-15 \\ 2.02\text{E}-14 \\ 3.53\text{E}-14 \end{bmatrix} & \begin{bmatrix} 4.21\text{E}-14 \\ 3.54\text{E}-14 \\ 7.60\text{E}-15 \\ 3.37\text{E}-14 \end{bmatrix} & \begin{bmatrix} 1.24\text{E}-14 \\ 3.10\text{E}-15 \\ 3.37\text{E}-14 \\ 6.32\text{E}-16 \end{bmatrix} \end{bmatrix}$ | $\begin{bmatrix} 2.38E-15 & 7.96E-15 & 8.24E-16 \\ 3.08E-15 & 8.36E-16 & 3.10E-15 \\ 2.92E-15 & 1.95E-15 & 1.72E-15 \end{bmatrix}$ | 8.24E-16 |
| $l = 50$ | $\boldsymbol{X}$ | $\boldsymbol{F}$ (Step 1, Eq 2) | $\boldsymbol{F^Z}$ (Step 2, Eq 3) | |
| | $\begin{bmatrix} 3\text{E}-8 & -3.84\text{E}-8 \\ 3\text{E}-8 & -3.84\text{E}-8 \\ 1.28\text{E}-9 & 5.4\text{E}-8 \end{bmatrix}, \begin{bmatrix} 8.31\text{E}-8 & 3.18\text{E}-8 \\ -2.78\text{E}-8 & 8\text{E}-9 \\ 1.92\text{E}-8 & -3.98\text{E}-8 \end{bmatrix}, \begin{bmatrix} -8.6\text{E}-9 & 2.9\text{E}-8 \\ -8.6\text{E}-9 & 2.9\text{E}-8 \\ 9.4\text{E}-8 & 3\text{E}-8 \end{bmatrix}$ | $\begin{bmatrix} 2.4-15 & 7.96E-15 & 8.24E-16 \\ 2.4E-15 & 8.36E-16 & 8.24E-15 \\ 2.9E-15 & 1.95E-15 & 9.92E-15 \end{bmatrix}$ | $[2.4\text{E}-15 \quad 8.36\text{E}-16 \quad 8.24\text{E}-16]$ | |

| | $\boldsymbol{F}^T$ (Step 4, Eq 7) | $\boldsymbol{F}^{T_Z}$ (Step 6, Eq 10) | $\boldsymbol{F}$ (Step 8, Eq 12) | **Final Solution** $f^*$ |
|---|---|---|---|---|
| | $\begin{bmatrix} \begin{bmatrix} 2.67E-15 \\ 9.29E-15 \end{bmatrix} & \begin{bmatrix} 2.77E-15 \\ 1.15E-14 \end{bmatrix} & \begin{bmatrix} 3.82E-16 \\ 1.29E-14 \end{bmatrix} \\ \begin{bmatrix} 4.01E-15 \\ 8.98E-15 \end{bmatrix} & \begin{bmatrix} 1.24E-14 \\ 1.15E-14 \end{bmatrix} & \begin{bmatrix} 8.41E-15 \\ 8.02E-15 \end{bmatrix} \end{bmatrix}$ | $\begin{bmatrix} \begin{bmatrix} 7.15E-15 \\ 6.48E-15 \\ 6.32E-15 \\ 1.33E-14 \end{bmatrix} & \begin{bmatrix} 1.11E-15 \\ 5.92E-15 \\ 7.86E-15 \\ 1.61E-14 \end{bmatrix} & \begin{bmatrix} 9.89E-15 \\ 9.98E-15 \\ 1.32E-14 \\ 3.60E-14 \end{bmatrix} \\ \begin{bmatrix} 2.93E-15 \\ 1.65E-14 \\ 1.64E-15 \\ 5.38E-15 \end{bmatrix} & \begin{bmatrix} 2.05E-15 \\ 4.46E-15 \\ 1.69E-15 \\ 1.11E-14 \end{bmatrix} & \begin{bmatrix} 2.44E-15 \\ 1.51E-14 \\ 2.53E-15 \\ 4.77E-15 \end{bmatrix} \end{bmatrix}$ | $\begin{bmatrix} 2.67E-15 & 1.11E-15 & 3.82E-16 \\ 1.64E-15 & 1.69E-15 & 2.44E-15 \\ 2.4E-15 & 8.36E-15 & 8.24E-16 \end{bmatrix}$ | **3.82E-16** |